\documentclass[lettersize,journal]{IEEEtran}

\usepackage{amsmath,amsfonts,bm}

\def\eqref#1{equation~\ref{#1}}

\def\1{\bm{1}}

\DeclareMathAlphabet{\mathsfit}{\encodingdefault}{\sfdefault}{m}{sl}
\SetMathAlphabet{\mathsfit}{bold}{\encodingdefault}{\sfdefault}{bx}{n}

\usepackage{url}            
\usepackage{booktabs}       
\usepackage{amsfonts}       
\usepackage{nicefrac}       
\usepackage{microtype}      
\usepackage{xcolor}         
\usepackage{graphicx}
\usepackage{tabularx}
\usepackage{subcaption} 
\usepackage{multirow}
\usepackage{amsmath,amsfonts,amssymb,amsthm,epsfig,epstopdf,url,array}
\usepackage{bm}
\usepackage{enumitem}
\usepackage[hypcap=false]{caption}
\usepackage{xspace}
\usepackage[most]{tcolorbox}
\usepackage{wrapfig}
\usepackage{colortbl}
\usepackage{pgf}
\usepackage{pifont} 
\usepackage{cite}

\newcommand{\heatmaplevel}[1]{%
  \ifnum#1=1 \cellcolor{blue!0}\fi
  \ifnum#1=2 \cellcolor{blue!5}\fi
  \ifnum#1=3 \cellcolor{blue!10}\fi
  \ifnum#1=4 \cellcolor{blue!15}\fi
  \ifnum#1=5 \cellcolor{blue!20}\fi
  \ifnum#1=6 \cellcolor{blue!25}\fi
  \ifnum#1=7 \cellcolor{blue!40}\fi
}

\newcommand{\cmark}{\ding{51}} 
\newcommand{\xmark}{\ding{55}} 

\newcommand{\PARSE}{\texttt{PARSE}\xspace} 

\def\BibTeX{{\rm B\kern-.05em{\sc i\kern-.025em b}\kern-.08em
    T\kern-.1667em\lower.7ex\hbox{E}\kern-.125emX}}
\usepackage{balance}
\begin{document}
\title{PID-Guided Partial Alignment and Overlay Design for Multimodal Decentralized Federated Learning among  Heterogeneous  Agents}
\author{
Yanhang Shi, Xiaoyu Wang, Houwei Cao,~\IEEEmembership{Member,~IEEE}, \\Jian Li,~\IEEEmembership{Senior Member,~IEEE}, and Yong Liu,~\IEEEmembership{Fellow,~IEEE}
\thanks{
Yanhang Shi is with the Department of Electrical and Computer Engineering,
Stony Brook University, Stony Brook, NY, USA.
Email: \texttt{yanhang.shi@stonybrook.edu}.}
\thanks{Jian Li is with the Department of Applied Mathematics and Statistics and the Department of Computer Science,
Stony Brook University, Stony Brook, NY, USA.
Email: \texttt{jian.li.3@stonybrook.edu}.
}
\thanks{
Xiaoyu Wang and Yong Liu are with the Department of Electrical and Computer Engineering,
New York University, Brooklyn, NY, USA.
Emails: \texttt{xw2597@nyu.edu}, \texttt{yongliu@nyu.edu}.
}
\thanks{
Houwei Cao is with the Department of Computer Science,
New York Institute of Technology, New York, NY, USA.
Email: \texttt{hcao02@nyit.edu}.
}
}

\markboth{Journal of \LaTeX\ Class Files,~Vol.~18, No.~9, September~2020}%
{How to Use the IEEEtran \LaTeX \ Templates}

\maketitle

\begin{abstract}
Multimodal decentralized federated learning (DFL) must support collaboration among agents that hold different modality subsets and often different model components, while operating over peer-to-peer (P2P) overlays without a coordinating server or a global network view. A key obstacle is that conventional multimodal training often relies on a single shared representation, which implicitly assumes that heterogeneous peers can exchange and aggregate the same model components over the same communication links. In multimodal DFL, this assumption breaks down: uni- and multimodal agents may push incompatible updates through shared overlays, weakening both inter-agent transfer and cross-modal interaction.
We present \PARSE, a server-free framework that brings partial information decomposition (PID) into multimodal DFL. Each agent splits its latent features into redundant, unique, and synergistic slices (``feature fission''), and performs slice-aware communication over modality-conditioned P2P overlays. During training, agents exchange only the slices that are semantically alignable with their neighbors, according to the modalities and model components they share (``partial alignment''). This design avoids centralized orchestration and gradient-surgery style conflict handling, while remaining compatible with standard DFL constraints and a range of P2P overlay topologies.
Across multiple benchmarks and heterogeneous peer mixes, \PARSE consistently outperforms task-, modality-, and hybrid-sharing multimodal DFL baselines while keeping per-link payloads bounded. Ablations on fusion choices and split ratios, together with qualitative feature analyses and overlay-topology studies, demonstrate the robustness and communication efficiency of the proposed slice-aware design.
\end{abstract}

\section{Introduction}\label{sec:intro}

Decentralized federated learning (DFL)~\cite{yuan2024decentralized} enables a collection of agents to train collaboratively over peer-to-peer (P2P) communication graphs, eliminating reliance on a centralized parameter server and improving robustness, scalability, and privacy relative to server-based FL~\cite{mcmahan2017communication}. In many distributed deployments such as autonomous systems~\cite{caesar2020nuscenes,zheng2023autofed,cui2024survey}, healthcare~\cite{zhang2020pdlens,bertsimas2024m3h}, and human--computer interaction~\cite{gao2020listen,liu2021rfid,lv2022deep,moin2023emotion}, agents rarely share identical sensing pipelines or model architectures. Instead, \emph{modality heterogeneity} is the norm: each agent may observe only a subset of modalities and implement partially different encoders or fusion operators. This makes \textit{multimodal DFL} fundamentally challenging: agents must (i) collaborate despite modality/architecture mismatch, (ii) reach consensus through sparse and potentially time-varying overlays without a server to arbitrate updates, and (iii) contend with conflicting training signals between unimodal and multimodal agents.

A common design pattern in multimodal learning is to train a single shared representation into which all modalities are mapped. While effective in centralized settings, such \emph{monolithic} embeddings become brittle in DFL because they force heterogeneous peers to route a uniform model payload over links whose endpoints may not share the same modalities or objectives. As a result,
(i) peers with different modality subsets update shared parameters using different local objectives, so a message that is useful to one neighbor may be partially misaligned with another;
(ii) these per-link inconsistencies simultaneously degrade cross-modal interaction within multimodal peers and inter-peer transfer along the P2P graph~\cite{ouyang2023harmony}; and
(iii) without a coordinator and with only local neighborhood views, such misalignment is difficult to detect and correct as updates propagate hop by hop across the overlay.
Section~\ref{sec:motivation} illustrates how existing multimodal DFL strategies expose these tensions:
task-based sharing~\cite{xiong2022unified} routes updates only among peers with identical modality sets, leaving partially overlapping peers disconnected from one another;
modality-based sharing~\cite{yuan2024communication} preserves per-modality learning but largely forfeits cross-modal interaction across the graph;
and hybrid schemes~\cite{chen2022towards} recover some interaction benefits yet can amplify per-link gradient conflicts under topology and modality heterogeneity.
Motivated by these challenges, the primary question we seek to address in this paper is:

\begin{tcolorbox}[colback=white!5!white,colframe=white!75!white]
\textit{How can we design {multimodal knowledge representations and DFL overlays} that (a) maximize knowledge transfer among heterogeneous peers with diverse modalities, (b) exploit cross-modal interactions when available, and (c) avoid destructive gradient conflicts, all under standard DFL constraints and without centralized coordination?}
\end{tcolorbox}

\begin{table}[t]
\centering
\small
\caption{Comparison with representative multimodal centralized and decentralized FL methods.}
\label{tab:rw_summary}
\scalebox{0.7}{
\begin{tabular}{lccccc}
\toprule
\textbf{Method} & \textbf{Server-free} & \textbf{Topology-agnostic} & \textbf{Gradient surgery-free} & \textbf{Synergistic}  \\
\midrule
FedMSplit~\cite{chen2022fedmsplit}         & \xmark & \xmark & \xmark  & \xmark \\ 
Harmony~\cite{ouyang2023harmony}     & \xmark & \xmark & \cmark  & \xmark \\
DMML-KD~\cite{kd-dfl}          & \xmark & \xmark & \cmark  &  \xmark \\
MCARN~\cite{yang2024MCARN} & \xmark & \xmark & \cmark  & \xmark \\ 
FedHKD \cite{wang2024fedhkd}     & \xmark & \xmark & \xmark  & \xmark \\ 
FedMVD \cite{gao2025multimodal} & \xmark & \xmark & \cmark  &  \xmark \\
\PARSE \textbf{(ours)} & \cmark & \cmark & \cmark  & \cmark \\
\bottomrule
\end{tabular}
}
\end{table}

To answer this question, we draw inspiration from partial information decomposition (PID), which conceptually separates multimodal information into \textit{redundant} (shared across modalities), \textit{unique} (modality-specific), and \textit{synergistic} (emergent from modality interactions) components~\cite{williams2010nonnegative,bertschinger2014quantifying,liang2023quantifying}.
We propose \PARSE (\underline{P}artially \underline{A}ligned Featu\underline{R}e Fis\underline{S}ion for d\underline{E}centralized learning), {a server-free multimodal DFL overlay design} built around two system-compatible mechanisms:

\textbf{(i) Feature Fission.} Each agent structures its latent knowledge representation into three explicit slices, namely redundant ($z^{r}$), unique ($z^{u}$), and synergistic ($z^{s}$), turning a monolithic embedding into separately addressable sub-streams. This decomposition exposes a communication interface: an agent updates only the slices associated with its locally available modality branches, thereby confining interference to alignable parameter blocks and preparing each slice for overlay-level exchange.

\textbf{(ii) P2P Knowledge Sharing with Partial Alignment.} Knowledge sharing is performed at the slice level. For each overlay link, PARSE treats slices as separately exchangeable units and transmits only those that are semantically alignable between the two endpoints. Non-alignable slices remain local. This partial-alignment rule enables useful exchange between peers with partially overlapping modality sets while preventing each link from carrying updates that its receiver cannot meaningfully aggregate.

As summarized in Table \ref{tab:rw_summary}, compared to representative multimodal centralized and decentralized FL methods, \PARSE offers the following properties, which constitute our main contributions:

\begin{itemize} 
\item \textbf{Coordination-free (server-free) slice-aware gossip.}
\PARSE removes the central orchestrator: each agent applies a purely local rule and exchanges \emph{only} modality-matched (slice-specific) model branches with neighboring peers that possess the corresponding modality, so slice-level alignment emerges over P2P links without any global scheduling or rebalancing. 

\item\textbf{Topology-agnostic operation via modality-conditioned subgraphs.}
\PARSE runs on arbitrary (including time-varying) P2P overlays by inducing \emph{per-modality} communication subgraphs: modality-conditioned consensus is performed only among agents that can meaningfully align the corresponding slices, making the protocol compatible with fixed graphs and gossip-style/random graphs.  Moreover, our evaluation shows \PARSE is robust across a range of overlay families (e.g., ring/chordal-ring and random-gossip graphs), and consistently benefits from faster-mixing topologies.

\item \textbf{Conflict avoidance by protocol design (no gradient surgery).}
Instead of resolving conflicting updates through centralized regularization or gradient ``surgery", \PARSE avoids destructive collisions structurally: agents never co-update or co-synchronize slices that are non-alignable under their modality sets, eliminating extra coordination rounds and costly gradient manipulation.

\item \textbf{Synergy-preserving collaboration under heterogeneity.}
Beyond redundant/unique disentanglement, \PARSE explicitly models the \emph{synergistic} slice that arises from cross-modal interaction and keeps it aligned with the agents capable of learning it, yielding improved accuracy over decomposition-only baselines while maintaining coordination-free P2P training.
\end{itemize} 

With these novel designs, \PARSE effectively facilitates collaborative and compatible learning among distributed agents with heterogeneous multimodal data. Its emphasis on decentralized architectures makes it inherently adaptable and robust to communication failures and peer churns, which is critical for large-scale distributed learning systems in real-world settings. 

The rest of this paper is organized as follows. Section~\ref{sec:related} reviews related work on multimodal federated and decentralized learning. Section~\ref{sec:motivation} presents the motivation behind \PARSE and formalizes the problem setting. Section~\ref{sec:framework} details the design of the \PARSE framework, including its key components and training procedure. Section~\ref{sec:exp} reports extensive experimental results that evaluate \PARSE against state-of-the-art baselines across multiple benchmarks. Finally, Section~\ref{sec:conclusion} concludes the paper and discusses promising directions for future work.

\section{Related Work}\label{sec:related}
We review three closely related research directions: multimodal representation learning, multimodal federated learning, and the emerging field of multimodal \emph{decentralized} federated learning.

\textbf{Multimodal Representation Learning.}
Early work merges modalities via simple concatenation or cross-attention, yielding a single embedding that blends redundant, unique, and synergistic cues \cite{gao2020listen,avedataset,he2021vi}.
Later methods tighten alignment through contrastive pre-training, such as CLIP \cite{radford2021learning} and ALIGN \cite{jia2021scaling}, or via information-theoretic decomposition \cite{liang2023quantifying}. But all assume centralized, full-modality access and a monolithic backbone—impractical under privacy or bandwidth constraints. 
In contrast, our feature-fission perspective is orthogonal and complementary: rather than learning a single embedding, we separate each latent vector into three partial information decomposition (PID)-motivated components \cite{liang2023quantifying}. This disentanglement allows distributed agents share only the information that should be shared, while preserving modality-specific and synergistic knowledge locally.

\textbf{Multimodal Federated Learning.}
Extensions of FL to multimodal data are still largely \emph{server-centric}.
Task-partitioned methods (e.g., FedPercepNet \cite{xiong2022unified}, FedHGRL \cite{chen2022towards}) cluster agents that share the same modality set and run FedAvg inside each cluster, blocking cross-cluster knowledge transfer and suffering under modality imbalance.
Modality-partitioned methods (e.g., BalancedMS \cite{fan2023balanced}) treat each modality as a virtual agent, so synergistic cross-modal cues remain under-utilized.
Hybrid schemes (e.g., FedHGB \cite{chen2022towards}, FedCLIP \cite{lu2023fedclip}, FedMSplit \cite{chen2022fedmsplit}, FedMBridge \cite{chen2024fedmbridge}) blend the two ideas but still rely on a central server to resolve gradient conflicts and rebalance agents. Knowledge decomposition approaches (e.g., MCARN~\cite{yang2024MCARN}, FedHKD~\cite{wang2024fedhkd}) effectively exploit sharing among modality-heterogeneous agents; however, realizing modality interaction still requires centralized control via a dedicated server. None of the above methods tackles gradient misalignment while being simultaneously coordination-free and topology-agnostic.

\textbf{Multimodal Decentralized Federated Learning.}
Eliminating the central server removes a single point of failure and alleviates privacy concerns, but it also makes cross-agent coordination substantially more difficult. DMML-KD~\cite{kd-dfl} is, to the best of our knowledge, the first algorithm expressly designed for multimodal DFL. It aligns modality-shared knowledge across agents by broadcasting a feature generator over a \emph{fully-connected} peer-to-peer network. However, it does not explicitly account for synergistic interactions and heavily relies on the presence of a sufficient number of multimodal agents.

\begin{figure}
    \centering
    \includegraphics[width=\linewidth]
{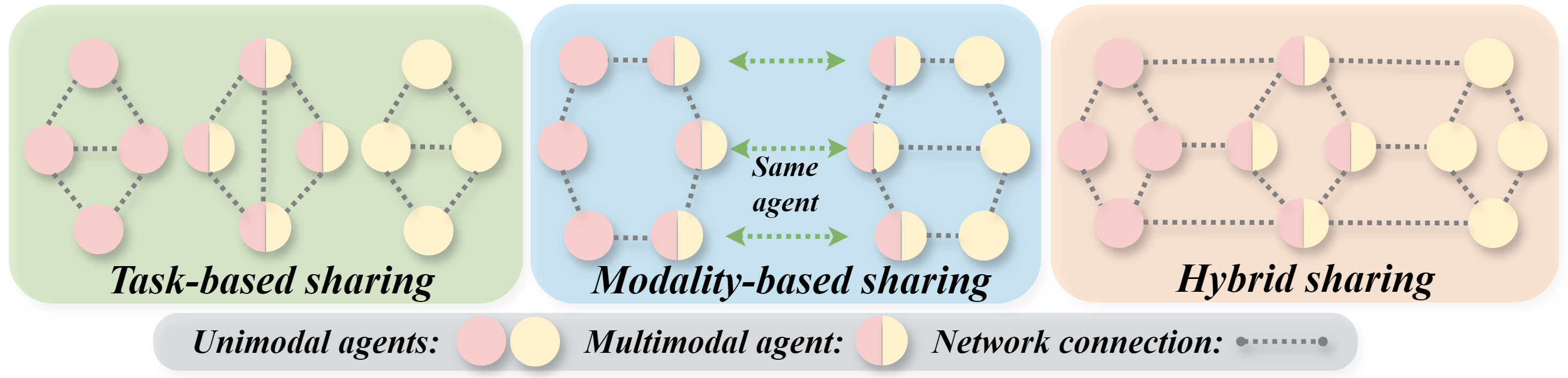}
  \caption{We study three knowledge-sharing strategies in DFL setting.  To illustrate, we consider a two-modality scenario involving three types of agents: \emph{Red:} modality~A only; \emph{yellow:} modality~B only; \emph{bicolored:} both.
}
  \label{fig:motivation-sharing}
\end{figure}

\begin{figure*}[t]
    \centering
    \scalebox{0.85}{
    \begin{subfigure}[t]{0.32\textwidth}
        \centering
        \includegraphics[width=\linewidth]{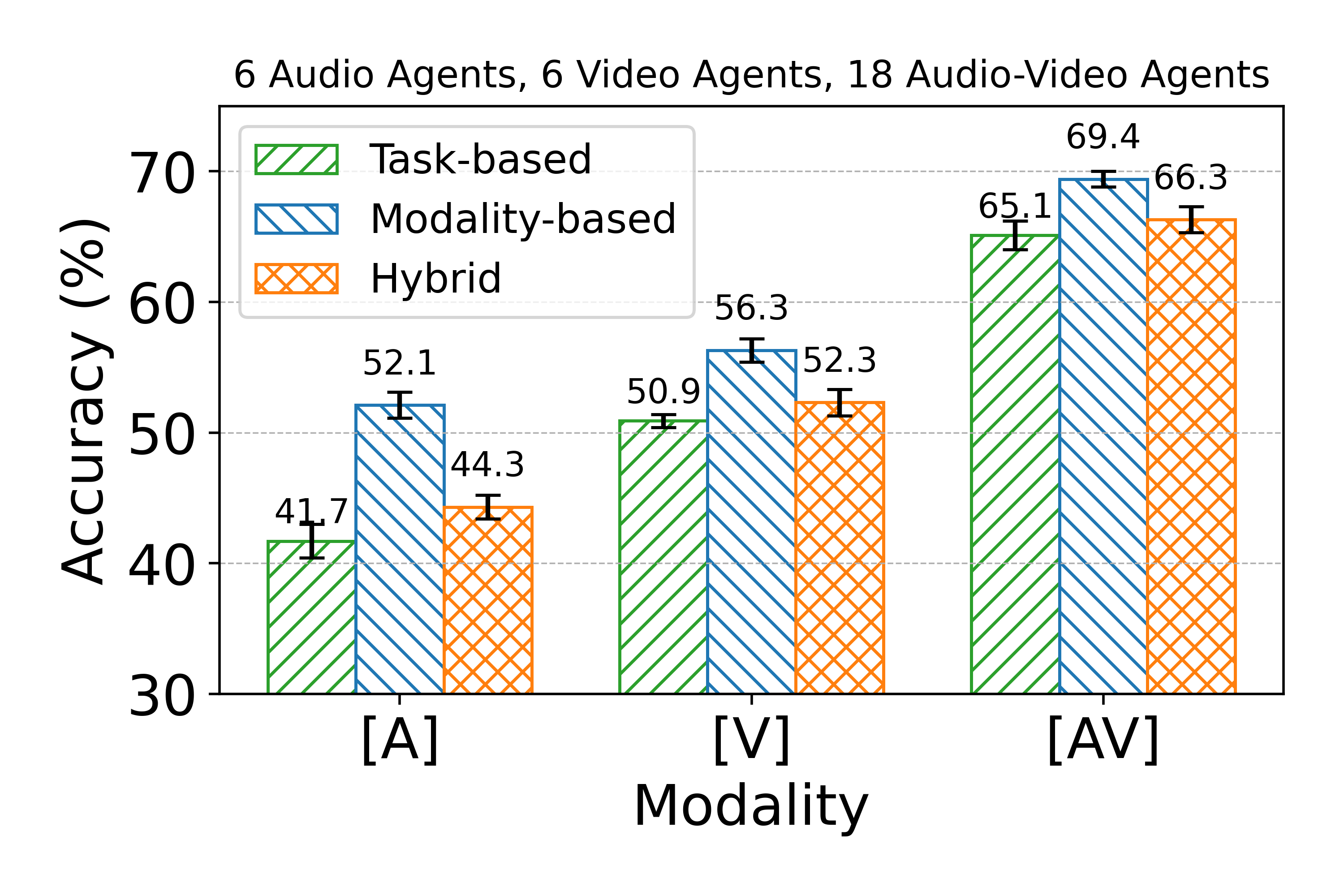}
        \caption{}
    \end{subfigure}
    \begin{subfigure}[t]{0.32\textwidth}
        \centering
        \includegraphics[width=\linewidth]{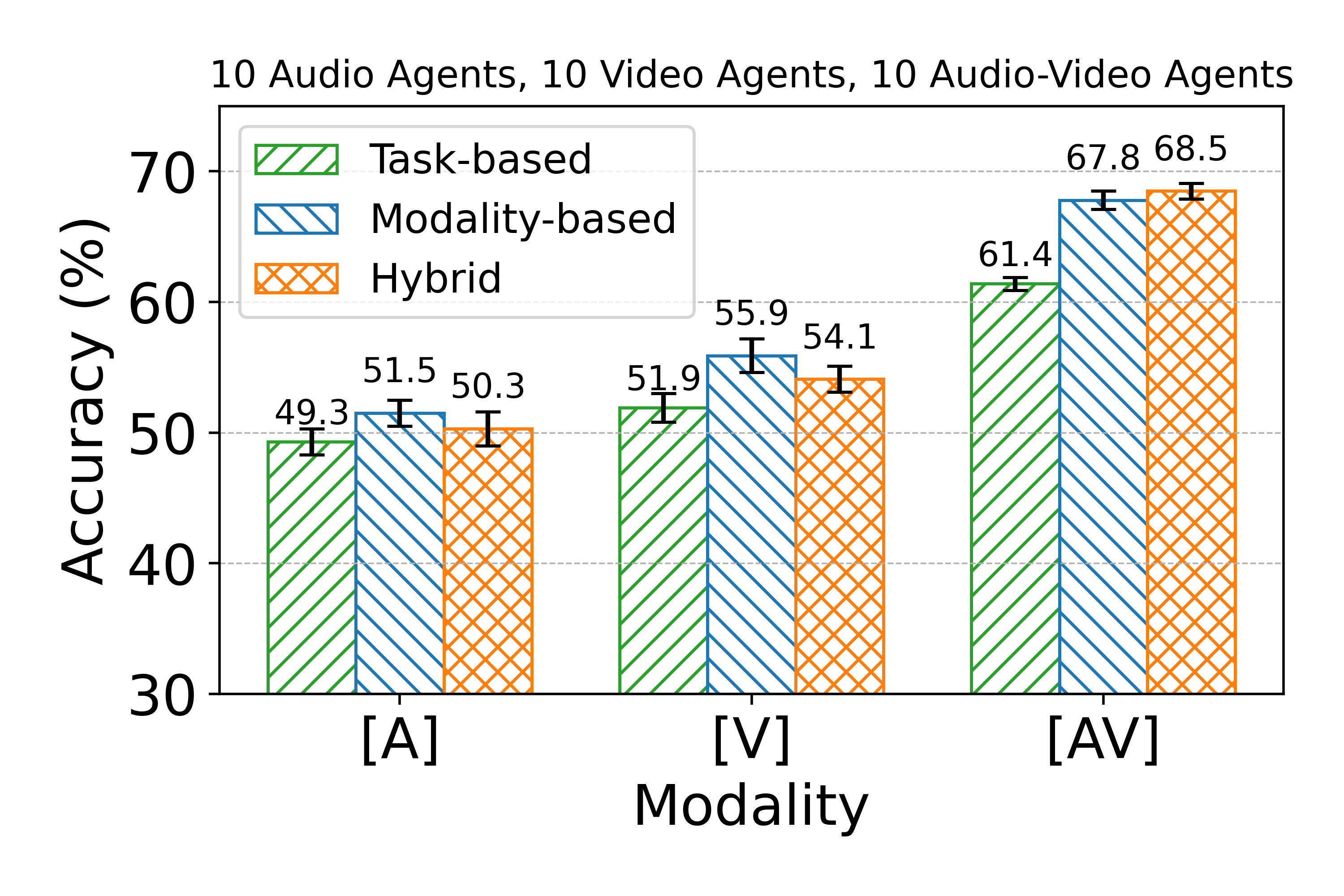}
        \caption{}
    \end{subfigure}
    \begin{subfigure}[t]{0.32\textwidth}
        \centering
        \includegraphics[width=\linewidth]{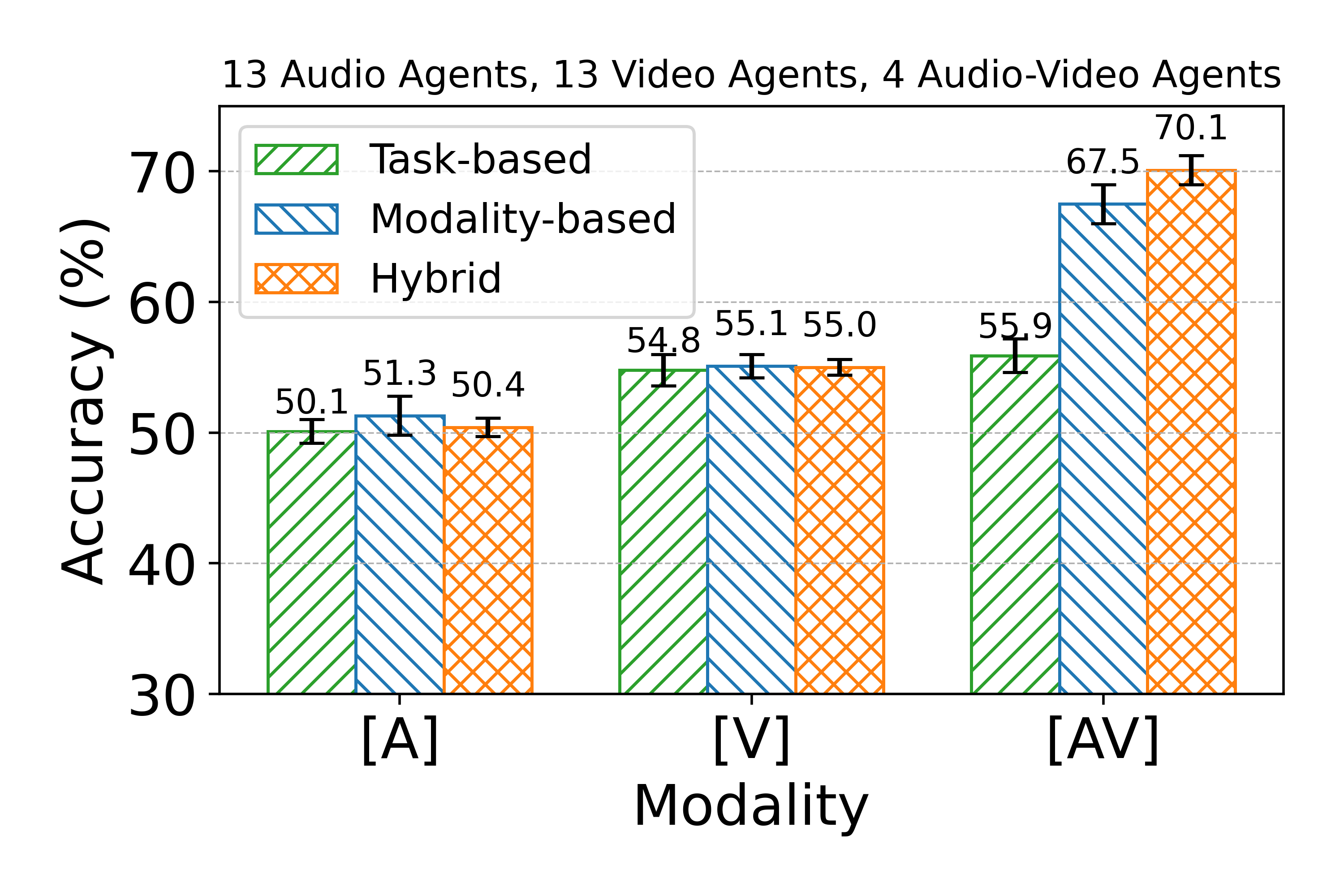}
        \caption{}
    \end{subfigure}
    }
    \caption{Test accuracy for unimodal and multimodal agents under task-, modality-, and hybrid sharing across three agent mixes: 6 audio, 6 video, and 18 multimodal agents; (b) 10 audio, 10 video, and 10 multimodal agents; and (c) 13 audio, 13 video, and 4 multimodal agents. 
    [A], [V], [AV] are averages over audio-only, video-only, and multimodal agents.
  }
  \label{fig:motivation-bars}
  \vspace{-0.1in}
\end{figure*}

\section{Motivation Study: Knowledge Alignment and Sharing on Multimodal Overlays}
\label{sec:motivation}

Multimodal DFL raises two coupled \emph{networking} challenges: \textit{(1) how to organize peer-to-peer overlays so that agents holding different modality subsets can still exchange useful updates, and (2) how to coordinate updates over those overlays without a central server to reconcile contributions from heterogeneous peers.} The two are tightly coupled: the choice of \emph{which agents form an overlay} determines \emph{what messages each overlay link must carry}, and without a server, inconsistencies between those messages must be resolved by the overlay itself. Learning a single shared embedding across all modalities and then gossiping it on an overlay shared by agents with different modality sets is fragile in server-free DFL, because no coordinator exists to absorb the gradient conflicts and training drifts caused by peer modality heterogeneity.

\subsection{Limitations of Existing Overlay Designs with Fully Aligned Representations}

A natural way to suppress multimodal gradient conflicts is to restrict gossip to overlays whose members share fully aligned representations. We examine three representative designs that follow this principle (Fig.~\ref{fig:motivation-sharing} from left to right):

\textbf{Task-based}~\cite{xiong2022unified}: agents with the same modality subset form an overlay and gossip a fully fused end-to-end model. Within an overlay the joint representation is fully aligned and captures cross-modal interaction, but the overlay decomposes into disjoint cliques — one per modality combination — and no messages flow between agents with partially overlapping modality subsets.

\textbf{Modality-based}~\cite{yuan2024communication}: for each modality, the agents that possess it form a per-modality overlay and gossip a common modality-specific representation. The representation is fully aligned within each overlay. But there is no overlay dedicated for learning cross-modal interactions.

\textbf{Hybrid}~\cite{chen2022towards}: identical per-modality overlays as the modality-based scheme, plus a local fusion head on multimodal agents. Encoder traffic on each overlay is unchanged, but multimodal peers now optimize a different local objective from unimodal peers on the \emph{same} overlay.

We evaluate the three designs with DSGD~\cite{lian2017dfl} on AVE~\cite{avedataset} (30 agents, three modality compositions, IID data):

$\bullet$ \textit{Task-based sharing has the worst performance.} Although intra-clique gossip is perfectly aligned, the absence of inter-clique links wastes the connectivity of the underlying P2P substrate: agents with partially overlapping modality subsets are not connected by overlays, severely limiting the learning efficiency of DFL.

$\bullet$ \textit{Modality-based sharing achieves strong unimodal performance, but weak multimodal performance.} The per-modality overlay maximizes within-modality reach and fully aligns local objectives along each link, but provides no overlay for cross-modal signal; multimodal accuracy trails the hybrid scheme, and the gap widens as the unimodal-agent ratio grows.

\begin{figure}
    \centering
    \includegraphics[width=0.8\linewidth]{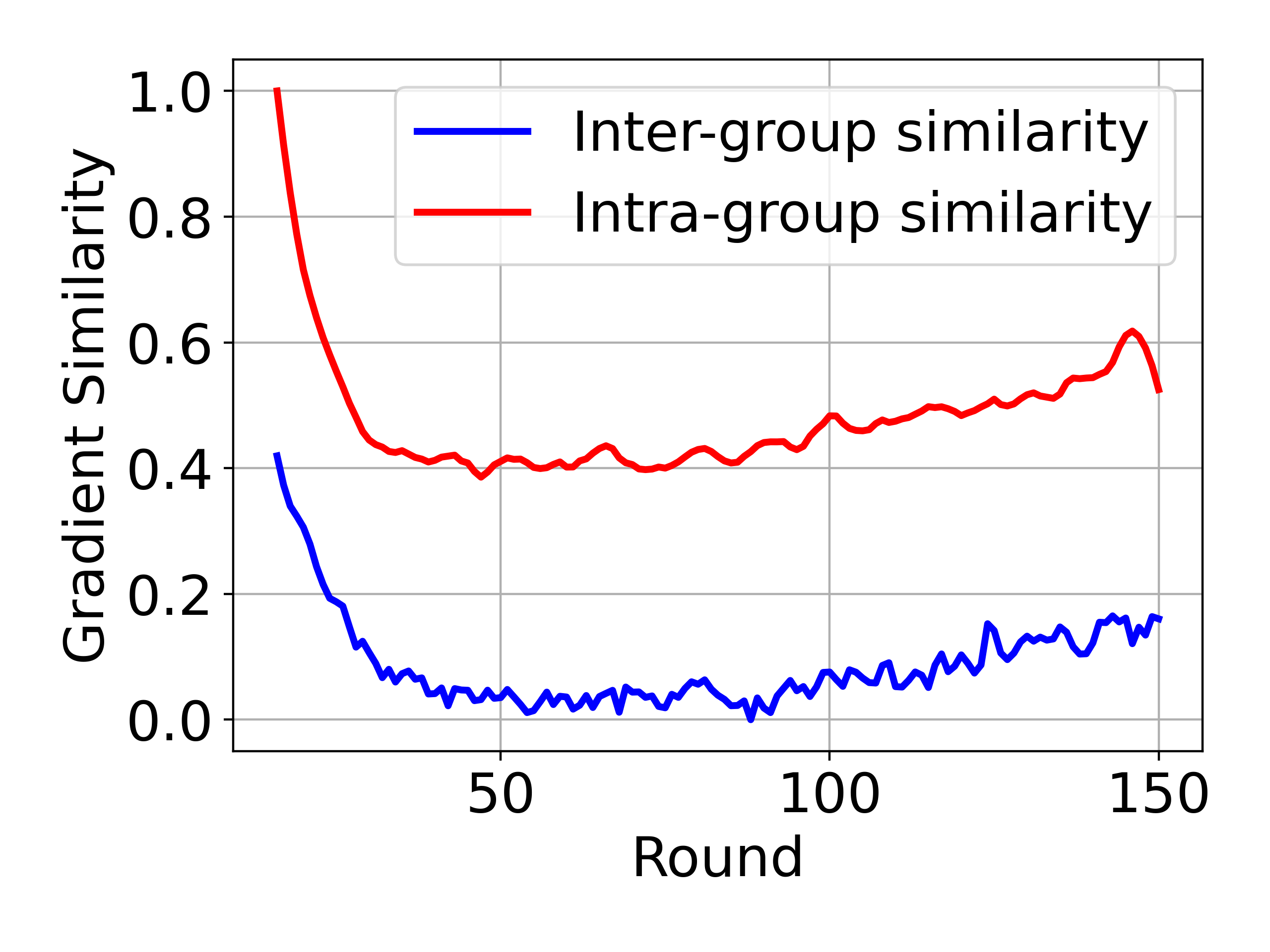}
    \vspace{-0.15in}
    \caption{Group agents by their modality subsets: inter-group and intra-group gradient cosine similarity (w.r.t.\ shared models) on the same per-modality overlay.}
    \label{fig:motivation-similarity}
\end{figure}

$\bullet$ \textit{Hybrid sharing fails to fully exploit multimodal interaction.} It yields only marginal gains for unimodal agents over task-based sharing, with modest multimodal improvements as the unimodal ratio increases (Fig.~\ref{fig:motivation-bars}(b),(c)). The cause is visible at the protocol level: peers with different modality subsets are admitted to the \emph{same} per-modality overlay but optimize against different local objectives, so each gossip step averages conflicting messages on every link rather than reinforcing a shared signal. Fig.~\ref{fig:motivation-similarity} confirms this: \emph{strong alignment} between agents with the same modality subset coexists with \emph{weak alignment} between agents with different modality subsets, even when both update the same modality branch on the same overlay.

\textbf{Key Takeaways.} \textit{1) To maximize DFL learning efficiency under sparse neighborhood connectivity, it is crucial to enable P2P exchange between heterogeneous agents that hold different subsets of data modalities, rather than fragmenting the overlay into homogeneous cliques. 2) Without a central coordinator, admitting heterogeneous agents to a single overlay that gossips a unified, modality-specific representation turns the overlay into a channel for inconsistent updates, limiting the effectiveness and stability of multimodal DFL.}

\subsection{Decoupled Local Training and Overlays via Feature Fission and Partial Alignment}

The diagnosis above motivates a different design point: rather than forcing heterogeneous peers to gossip a monolithic modality-specific representation on a single overlay, we \emph{decouple local training} so that each modality-specific representation is decomposed into multiple disentangled subspaces, or slices, each encoding a distinct facet of multimodal information (\textit{feature fission}); we then \emph{form per-slice overlays} among the agents that hold the corresponding component (\textit{partial alignment}), so every link only carries messages its endpoints can mutually average. This converts the overlay-level inconsistency exposed in Section~2.1 into a clean separation between local optimization and per-component communication.

\noindent\textbf{Feature Fission through Partial Information Decomposition.} For two modalities $X_1, X_2$ predicting $Y$, partial information decomposition (PID)~\cite{williams2010nonnegative,bertschinger2014quantifying,liang2023quantifying} separates
\begin{align}
I(X_1,X_2;Y)\!=\! \underbrace{U_{X_1}(Y)\!+\! U_{X_2}(Y)}_{\text{unique}} \!+\! \underbrace{R(Y)}_{\text{redundant}} \!+\! \underbrace{S(Y)}_{\text{synergistic}},
\end{align}
where $R(Y)$ denotes information about $Y$ shared by \emph{both} modalities, $U_{X_i}(Y)$ denotes information  \emph{specific} to $X_i$, and $S(Y)$ denotes information that emerges \emph{only} from joint observation. We illustrate PID with two modalities for clarity, but both PID and our design extend naturally to more than two modalities~\cite{griffith2014quantifying}.

\emph{Networking interpretation (overlay rule).} PID directly prescribes \emph{the overlay on which each component should be exchanged}, based on modality membership: (i) the \emph{unique} slice is modality-exclusive and is gossiped only on the per-modality overlay of agents that possess that modality; (ii) 
the \emph{redundant} slice represents the component intended to be compatible across modality branches, and is therefore exchanged on the per-modality overlay among all agents that possess that modality;
(iii) the \emph{synergistic} slice captures cross-modal complementarity, is learnable only by multimodal peers, and is therefore exchanged on a small overlay among agents that share the same modality set. Fissioning each modality representation into redundant/unique/synergistic slices and routing them on per-component sub-overlays is the core mechanism of \PARSE.

\noindent\textbf{P2P Knowledge Sharing with Partial Alignment.} Because each agent only updates the slices its modalities entitle it to, local training is decoupled across heterogeneous peers and no message arriving on a link ever touches a parameter the sender cannot meaningfully contribute to. Overlay-level alignment then emerges \emph{automatically} from the membership rule: any pair of peers gossips only on the sub-overlays whose slices both endpoints hold, and is never distracted by gradient conflicts on slices that cannot be aligned. The construction imposes no constraint on the underlying P2P topology and requires neither central coordination nor gradient surgery, so it composes with any DFL gossip protocol — properties detailed in the \PARSE framework (Section~\ref{sec:framework}).

\begin{figure*}[t]
  \centering
  \includegraphics[width=0.8\textwidth]{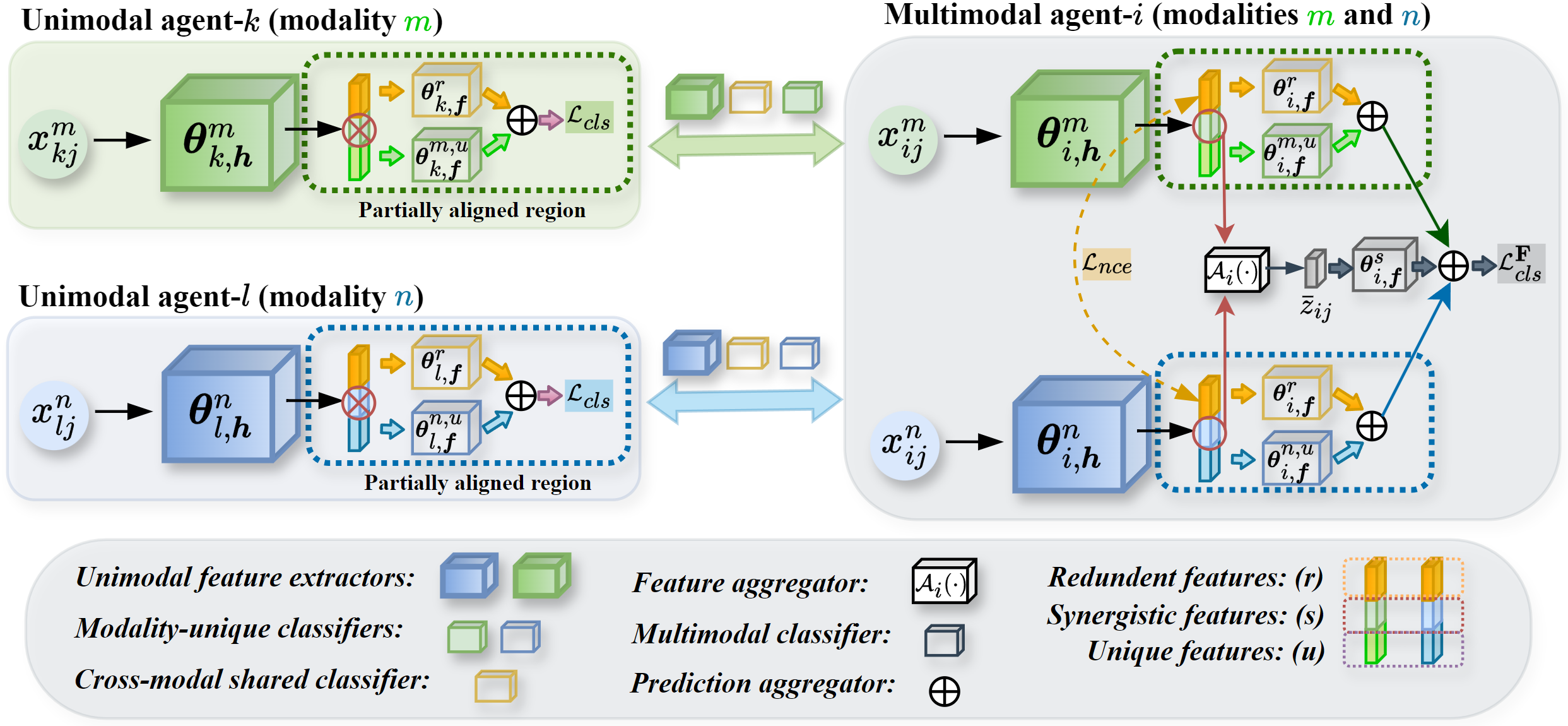}
  \caption{ \textbf{\PARSE at a glance (two modalities for illustration).} Agents that share a modality form a \emph{modality-specific} P2P subgraph. Each encoder output is \emph{fissioned} into redundant ($z^{r}$), synergistic ($z^{s}$), and unique ($z^{u}$) slices. Unimodal agents train on $z^{u}$ and $z^{r}$. Multimodal agents: (i) \emph{partially align} $z^{u}$ and $z^{r}$ across agents, and (ii) learn a multimodal classifier on the fused $z^{s}$. This routes shareable information while keeping modality-exclusive and joint-only components local.}  \label{img:design}
\end{figure*}

\section{The \PARSE Framework for Multimodal DFL}\label{sec:framework}

Motivated by Section~\ref{sec:motivation}, we formally present \PARSE, a server-free multimodal DFL framework for heterogeneous agents with arbitrary modality mixes. Fig.~\ref{img:design} illustrates the two-modality, three-agent case for clarity, while our framework supports any number of modalities and agents. \PARSE couples an encoder-classifier architecture with \emph{feature fission} and \emph{partial alignment}: each modality $m$ has a feature encoder $h^m$ whose output latent is factorized into redundant ($z^r$), unique ($z^u$), and synergistic ($z^s$) slices; unimodal agents train unique and redundant heads on their local data, while multimodal agents additionally fuse $z^s$ to train a synergistic head.
This design lets agents exchange what should be shared while preserving modality-exclusive capacity and learning synergy without a server.

\subsection{Problem setup}\label{sec:problem}

Let $\mathcal{M}$ be the global set of modalities. We consider a set of agents $\mathcal{N}$, where agent $i\in\mathcal{N}$ owns a subset of modalities $\mathcal{M}_i\subseteq\mathcal{M}$ and a local dataset
$S_{\mathcal{M}_i}=\{(\mathbf{x}_{ij},y_{ij})\}_{j=1}^{|S_{\mathcal{M}_i}|}$ with
$\mathbf{x}_{ij}=\{x^{m}_{ij}\}_{m\in\mathcal{M}_i}$ and $y_{ij}\in\mathcal{Y}$.
All agents share the same label space $\mathcal{Y}$, while both the input distributions and the available modality subsets $\mathcal{M}_i$ may be heterogeneous and non-IID across agents.

\textbf{Model Family.}
For each modality $m\in\mathcal{M}$, agents that possess $m$ use a (potentially shared) encoder
$h^m:\mathcal{X}^m\to\mathcal{Z}^m$ to produce latent features.%
\footnote{Architectures can differ across modalities (e.g., CNN for vision, Transformer for text) and, when desired, be shared across agents that own the same modality.}
For a multimodal agent $i$, an aggregator $\mathcal{A}_i:\prod_{m\in\mathcal{M}_i}\mathcal{Z}^m\to\mathcal{Z}$ maps per-modality features into a representation space $\mathcal{Z}$, and a classifier $f:\mathcal{Z}\to\mathcal{Y}$ outputs predictions.
In Section~\ref{sec:fission} we will \emph{instantiate} this generic family with \PARSE by factorizing each encoder’s latent into redundant/unique/synergistic slices and by using component-wise heads and partial alignment.

\textbf{Learning Objective.}
Let $\boldsymbol{\theta}$ collect all trainable parameters (encoders, aggregator(s), and classifier(s)).
The decentralized objective minimizes the average local loss \begin{align}\label{eq:objective}
\min_{\boldsymbol{\theta}}\;\frac{1}{|\mathcal{N}|}\sum_{i\in\mathcal{N}}
\mathbb{E}_{(\mathbf{x},y)\sim S_{\mathcal{M}_i}}\!\left[\,
\mathcal{L}_i(\mathbf{x},y;\boldsymbol{\theta})
\,\right],
\end{align}
where $\mathcal{L}_i$ is a task-appropriate loss (e.g., cross-entropy). Due to modality heterogeneity, agent $i$ updates only the parameters tied to $\mathcal{M}_i$ (its encoders for $m\in\mathcal{M}_i$ and any heads it hosts).

\textbf{Decentralized P2P Communication.}
Agents join multiple learning overlays and exchange parameters with overlay neighbors in a P2P fashion. There is one overlay for each modality. All agents possessing that modality join the overlay to collectively train models specific to that modality. 
In addition, multimodal agents sharing the same subset of modalities also form an overlay to jointly train a common multimodal classifier for the shared modality subset. The precise update rule (DSGD with mixing matrices) and the set of parameters exchanged are detailed in Section~\ref{sec:sharing}.

\subsection{Feature fission}\label{sec:fission}
To separate what is shareable from what must remain local (as motivated in Section~\ref{sec:motivation}), each encoder output is factorized into PID-motivated slices. For $x^m_{ij}$,
\begin{align}
z^m_{ij} \;=\; h^m(x^m_{ij};\boldsymbol{\theta}^m_{i,\bm{h}}) \;=\; \operatorname{Concat}\!\big(z^{m,r}_{ij},\,z^{m,s}_{ij},\,z^{m,u}_{ij}\big),
\end{align}
where $z^{m,r}$ captures \emph{redundant} information (present in all modalities), $z^{m,s}$ captures \emph{synergy} (available only under joint observation), and $z^{m,u}$ captures \emph{unique} modality-specific cues. We adopt an equal split by default, and Section~\ref{sec:exp} reports split-ratio sweeps.

This fission supplies a routing mechanism in DFL: neighbors exchange modality-specific parameters, with shareable information automatically aligned ($z^{u}$ and $z^{r}$).

\subsection{Losses and partial alignment}\label{sec:losses}

\textbf{Modality-unique and redundant Training.}
Each agent trains two heads per available modality $m\!\in\!\mathcal{M}_i$: a unique head $f_i^u(\cdot; \boldsymbol{\theta}^{m,u}_{i,\bm{f}})$ on $z^{m,u}$ and a redundant head $f_i^r(\cdot; \boldsymbol{\theta}^{r}_{i,\bm{f}})$ on $z^{m,r}$. The per-modality prediction is
$\hat{y}^m_{ij}=f_i^u(z^{m,u}_{ij})+f_i^r(z^{m,r}_{ij})$,
with classification loss
\begin{align}
\mathcal{L}_{cls}(\boldsymbol{\theta}^m_{i,\bm{h}},\boldsymbol{\theta}^{m,u}_{i,\bm{f}},\boldsymbol{\theta}^r_{i,\bm{f}})
=\frac{1}{|S_{\mathcal{M}_i}|}\sum_{j}\ell\big(\hat{y}^m_{ij},y_{ij}\big),
\end{align}
where $\ell:\mathcal{Y\times Y}\rightarrow\mathbb{R}$ is a task-appropriate loss function (e.g., cross-entropy).

\textbf{Contrastive Diversity for redundant Alignment.}
Only redundant slices are aligned across modalities of the \emph{same} sample, while unique/synergistic slices act as hard negatives to encourage orthogonality. To this end, we impose a contrastive objective $\mathcal{L}_{nce}$ on the redundant features: $\mathcal{L}_{nce}(\Theta_{i,\bm{h}})=$
\begin{small}
\begin{align}
\label{eq:nce}
\frac{-1}{|S_{\mathcal{M}_i}|}
\sum_{j}\!\!\!\sum_{\substack{(m,m')\in\mathcal{M}_i\\ m\neq m'}}
\!\!\!\!\!\log\!\frac{\exp\!\bigg(\frac{\mathrm{sim}(z^{m,r}_{ij},z^{m',r}_{ij})}{\tau}\bigg)}
{\exp\!\bigg(\frac{\mathrm{sim}(z^{m,r}_{ij},z^{m,u}_{ij})}{\tau}\bigg)\!+\!\exp\!\bigg(\frac{\mathrm{sim}(z^{m,r}_{ij},z^{m,s}_{ij})}{\tau}\bigg)},
\end{align}
\end{small}
where $\Theta_{i,\bm{h}} = \{\boldsymbol{\theta}^{m}_{i,\bm{h}}\}_{m \in \mathcal{M}_i}$, $\text{sim}(\cdot, \cdot)$ denotes a similarity metric (e.g., cosine similarity), and $\tau$ is a temperature hyperparameter. 

\textbf{Synergy on Multimodal Agents.}
For multimodal agent $i$, synergistic features are fused by a parameter-free mean
$\bar{z}_{ij}=\mathcal{A}_i(\{z^{m,s}_{ij}\}_{m\in\mathcal{M}_i})=\frac{1}{|\mathcal{M}_i|}\sum_{m\in\mathcal{M}_i}z^{m,s}_{ij}$,
then classified by $f_i^s(\cdot; \boldsymbol{\theta}^s_{i,\bm{f}})$ with an ensemble loss that encourages cooperation with per-modality predictions: $\mathcal{L}^{\boldsymbol{F}}_{cls}(\Theta_{i,\bm{h}},\Theta^u_{i,\bm{f}},\boldsymbol{\theta}^r_{i,\bm{f}},\boldsymbol{\theta}^s_{i,\bm{f}})
=$
\begin{small}
\begin{align}
\label{eq:cls_f}
\frac{1}{|S_{\mathcal{M}_i}|}\sum_{j}\ell\left(f_i^s(\bar{z}_{ij})+\sum_{m\in\mathcal{M}_i}\hat{y}^m_{ij},y_{ij}\right),
\end{align}
\end{small}
with $\Theta^u_{i,\bm{f}}=\{\boldsymbol{\theta}^{m,u}_{i,\bm{f}}\}_{m\in\mathcal{M}_i}$ being the collection of all unimodal classifier parameters.

\textbf{Local Objectives.}
Unimodal agent $i$ minimizes $\mathcal{L}_{cls}$, while multimodal agent $i$ minimizes
\begin{align}
\label{eq:final_loss}
\mathcal{L}_i=\mathcal{L}^{\boldsymbol{F}}_{cls}(\cdot)+\beta\cdot\mathcal{L}_{nce}(\cdot),
\end{align}
where $\beta$ balances alignment. Inference uses the ensemble
$\hat{y}_{ij}=f_i^s(\bar{z}_{ij})+\sum_{m\in\mathcal{M}_i}\hat{y}^m_{ij}$.

\textbf{Partial Alignment.}
Rather than training the three branches in isolation, we optimize an \emph{ensemble} prediction (Eq.~(\ref{eq:cls_f})) in which each modality contributes a unique head $f^u$ and a shared redundant head $f^r$, and multimodal agents add a synergistic head $f^s$. The contrastive-diversity loss in Eq.~(\ref{eq:nce}) (i) \emph{pulls} redundant features across modalities for the \emph{same} sample (positives), and (ii) treats the sample’s own unique and synergistic components as \emph{hard negatives}, thereby encouraging $z^{r}$, $z^{u}$, and $z^{s}$ to capture complementary information.
Crucially, only the redundant slice $z^{r}$ (and its head $f^r$) participates in cross-agent synchronization; $z^{u}$ and $z^{s}$ are explicitly pushed away from $z^{r}$ by Eq.~(\ref{eq:nce}) and are optimized \emph{locally}. This ``partial'' alignment mitigates uni- vs. multi-modal gradient conflict, preserves modality-exclusive capacity, and lets synergy be learned where it exists (on multimodal agents) without contaminating unimodal updates. In practice, the ensemble objective allows agents with different modality sets to share a common optimization signal through the aligned redundant branch while keeping non-shared branches non-interfering, yielding stable improvements even if any single sub-predictor is imperfect (cf.  Eq.~(\ref{eq:final_loss})).

\textbf{Protocol invariant (slice-aware exchange).}
For every P2P link, agents exchange/mix only \emph{alignable} parameter blocks. In particular, the redundant branch is synchronized across agents, while non-alignable unique/synergy branches are never mixed.

\subsection{Overlay Construction}\label{sec:overlay_con}
The membership rule in Section~\ref{sec:motivation} specifies \emph{which} agents should exchange each slice, but not how the corresponding logical overlays are formed and maintained in a server-free deployment. \PARSE only requires a lightweight membership mechanism that lets agents discover peers sharing a modality or modality set. Below we describe one practical tracker-assisted instantiation for a single per-modality overlay $\mathcal{G}^m$; the same procedure can run independently for other modality and modality-set overlays.

\textbf{Tracker-assisted overlay management.} Each logical overlay elects one member as a lightweight \emph{tracker} that maintains only control-plane metadata: the current member set $\mathcal{N}^m$, the neighbor assignment inducing $\mathcal{E}^m$, and a deterministic succession order for backup trackers. The tracker is not a learning coordinator: it never observes model parameters or gradients and is not on the gossip path. Its role is limited to keeping the overlay membership and neighbor assignment well-defined as agents join, leave, or fail. New agents join via a bootstrap contact and are assigned neighbors according to the chosen base topology, such as ring, chordal ring, or random gossip. Membership updates are replicated to overlay members, and tracker failover follows the deterministic succession order, so a backup can resume the control-plane role without changing the gossip protocol.

\textbf{Topology compatibility.}
The overlay-management mechanism is separate from the learning protocol. Once $\mathcal{N}^m$ and $\mathcal{E}^m$ are established, \PARSE applies the same slice-aware DSGD update regardless of whether the overlay is a ring, chordal ring, random-gossip graph, or another connected topology. The resulting membership record provides the structural inputs used in Section~\ref{sec:sharing}: the per-modality node sets $\mathcal{N}^m$, edge sets $\mathcal{E}^m$, and mixing matrices $W^m$ derived from the assigned topology.

\subsection{Knowledge Sharing}\label{sec:sharing}

We adopt decentralized SGD (DSGD)~\cite{lian2017dfl}, where agents exchange parameters only with graph neighbors. To respect modality heterogeneity, each multimodal agent is instantiated as multiple \emph{unimodal virtual agents}, one per owned modality, and each virtual agent participates in a \textit{modality-specific} communication overlay.

\textbf{Per-modality Subgraphs.}
For each modality $m\in\mathcal{M}$, we denote the sub-overlay (or a subgraph) as
$\mathcal{G}^{m}=(\mathcal{N}^{m},\mathcal{E}^{m})$ with
$\mathcal{N}^{m}=\{i\in\mathcal{N}\mid m\in\mathcal{M}_i\}$.
An edge $(i,k)\in\mathcal{E}^{m}$ means agents $i$ and $k$ can exchange updates for modality $m$.
Each subgraph is equipped with a mixing matrix $\mathbf{W}^{m}\in\mathbb{R}^{|\mathcal{N}^{m}|\times|\mathcal{N}^{m}|}$ (row-stochastic, respecting the sparsity of $\mathcal{E}^{m}$), where $W^{m}_{ik}$ is the weight agent $i$ assigns to neighbor $k$ on modality $m$.

\textbf{Neighbor Mixing.}
At each communication round, agent $i$ performs a local step and then mixes with its neighbors:
\begin{align}
\boldsymbol{\theta}^{m}_i \;\leftarrow\;
\sum_{k\in\mathcal{N}^{m}} W^{m}_{ik}\big(\boldsymbol{\theta}^{m}_k - \eta\,\nabla_{\boldsymbol{\theta}^{m}_k}\mathcal{L}_k\big),
\end{align}
where $\eta$ is the learning rate, and
$\boldsymbol{\theta}^{m}_i=\{\boldsymbol{\theta}^{m}_{i,\bm{h}},\;\boldsymbol{\theta}^{m,u}_{i,\bm{f}},\;\boldsymbol{\theta}^{r}_{i,\bm{f}}\}$
collects the modality-$m$ encoder, the modality-unique head, and the shared redundant head maintained by agent $i$. Subgraphs $\mathcal{G}^{m}$ can use ring, exponential, or other sparse topologies without changing the update rule.

\textbf{Synergistic-head Subgraphs.} 
Parameters of the multimodal (synergistic) classifier $\boldsymbol{\theta}^{s}_{i,\bm{f}}$ are exchanged only among agents that share the same modality set $\mathcal{M}_i$ (a small subgraph per set). Since $f_i^{s}$ is a single linear layer, this additional message is negligible compared with exchanging encoders.

\noindent\textit{\textbf{Summary.}}
By communicating on per-modality subgraphs and mixing only the parameters tied to owned modalities, \PARSE routes shareable information across peers while preserving the integrity of unshared parts, enabling stable collaboration without a server.

\section{Experiments}\label{sec:exp}

\subsection{Implementation and Testbed}\label{sec:implementation}
We implement \PARSE in \texttt{Python3} as a containerized multi-agent system: each agent is instantiated as an isolated Docker container (based on a \texttt{Pytorch Cuda} runtime image\footnote{Version: pytorch:2.9.1-cuda13.0-cudnn9-runtime}) and executes local training plus slice-aware neighbor exchange according to the protocol.
All experiments are conducted on a server equipped with an AMD Ryzen Threadripper PRO 7955WX CPU and three NVIDIA RTX PRO 5000 GPUs.
To support many concurrently running agents while preventing resource contention, we enable NVIDIA Multi-Process Service (MPS) to partition GPU resources across containers; by default, we allocate $5\%$ of each GPU to a container and map containers to the available GPUs.
To emulate decentralized P2P communication, we construct virtual overlay networks among containers using Mininet~\cite{mininet}: given a target topology (defined below), Mininet instantiates the corresponding virtual links between container pairs, enabling controlled comparisons across different P2P graphs.
Consistent with our modality-conditioned subgraph design, \PARSE exchanges only modality-matched parameter blocks with neighbors during communication.

\subsection{Settings}

{\bf Datasets.} (1) KU-HAR \cite{kuhar} is a human-activity-recognition benchmark with 18 actions captured by two sensors, namely, accelerometer (A) and gyroscope (G). We perform an eight-class setting (walking, walking upstairs, walking downstairs, sitting, standing, laying, jumping, and running).
(2) ModelNet-40 \cite{modelnet} contains CAD models from 40 object categories. We treat two rendered views of each 3-D object as distinct modalities {V1, V2}.
(3) AVE \cite{avedataset} (Audio-Visual Event) comprises 10-s video clips from 28 event classes with synchronized audio and visual streams {A, V}.
(4) IEMOCAP \cite{iemocap} is an emotion-recognition corpus with audio, visual, and text modalities {A, V, T}. Following \cite{liang2020semi}, we keep four emotion labels, namely, \emph{happy}, \emph{sad}, \emph{angry}, and \emph{neutral}.

{\bf Communication setup.} We follow the modality- and task-based connection schemes described in Section \ref{sec:motivation}. Unless stated otherwise, agents are linked in a ring topology \cite{ring}. Each ring is formed either (i) by agents that share a given modality (modality-specific ring) or (ii) by agents that have the exact same modality set (task-specific ring). During every communication round, each agent runs one local training epoch and then exchanges its modality-specific parameters only with its immediate ring neighbors.

{\bf Baselines.} Because multimodal DFL is still nascent, we adapt several server-based multimodal FL methods to the peer-to-peer setting, selecting those that do not fundamentally depend on a central server: (1) Harmony \cite{ouyang2023harmony}: a two-stage scheme that we instantiate as modality-based sharing followed by task-based sharing. (2) FedHGB \cite{chen2022towards}: employs hierarchical gradient blending; we run it with a hybrid topology. (3) DMML-KD \cite{kd-dfl}: uses a shared feature generator across modality-heterogeneous agents; we implement it under the same hybrid graph for fair comparison. We also include three lightweight baselines derived from DSGD (see Section \ref{sec:motivation}): (4) DSGD-Modality: modality-based sharing \cite{yuan2024communication}; (5) DSGD-Task:task-based sharing \cite{xiong2022unified}; (6) DSGD-Hybrid: hybrid sharing \cite{chen2022towards}. 

\subsection{Model Setup}
{\bf KU-HAR.} As summarized in Table \ref{tab:har_encoders}, we use a compact temporal encoder for processing inertial signals from the KU-HAR dataset. The model consists of a 1D convolutional encoder followed by a GRU layer. The convolutional block reduces the temporal resolution while expanding the feature dimension, and the GRU captures sequential dependencies across time. A final average pooling layer aggregates temporal information into a fixed-length embedding for downstream classification.

\begin{table}[t]
\centering
\caption{Architecture of the feature extractors for KU-HAR (accelerometer and gyroscope).}
\label{tab:har_encoders}
\small
\begin{tabularx}{\linewidth}{@{}lX@{}}
\toprule
\textbf{Module} & \textbf{Description} \\
\midrule
Input & Raw inertial sequence with 64 time steps and 3-axis sensor readings. $B$ is the batch size. \\
\midrule
Conv1dEncoder & 1D convolution that reduces time from 64 to 8, expands channels to 128, with dropout = 0.1. \\
\midrule
GRU & Single-layer, unidirectional GRU with input size 128 and hidden size 192. Dropout = 0.1. \\
\midrule
Average Pooling & Temporal average pooling across 8 time steps for fixed-length output. \\
\bottomrule
\end{tabularx}
\vspace{-0.1in}
\end{table}

{\bf ModelNet-40.}
For visual encoding, we use a MobileNetV3-Small backbone~\cite{MobileNet} (pretrained) and replace the classifier with a linear projection to a 300-d embedding.

{\bf AVE.}
We use MobileNetV3-Large backbones for both audio and vision. For audio spectrograms (1 channel), we modify the first conv from $3\!\rightarrow\!16$ to $1\!\rightarrow\!16$ channels (kernel $3$, stride $2$, padding $1$), keeping the rest unchanged. For both streams, we replace the classifier with a linear projection to a 384-d embedding (ImageNet-pretrained for vision).

{\bf IEMOCAP.}
We use modality-specific encoders (Table~\ref{tab:iemocap_encoders}). Audio (130-d/frame) and video (342-d/frame) are encoded by a 1-layer uni-directional LSTM (hidden size 384) followed by temporal max pooling. Text uses 1024-d BERT utterance embeddings, passed through parallel 1D CNN branches (kernel sizes 3/4/5), concatenated, dropout (0.5), and a ReLU FC projection to a 384-d embedding.

\begin{table}[ht]
\centering
\caption{Architectures of the feature extractors for IEMOCAP (where $B$ is the batch size).}
\label{tab:iemocap_encoders}
\small
\resizebox{\linewidth}{!}{
\begin{tabular}{@{}l|lc@{}}
\toprule
\textbf{Modality} & \textbf{Module} & \textbf{Description} \\
\midrule
\multirow{3}{*}{Audio} 
& Input & Frame-level acoustic features with 130 dimensions. \\
& LSTM & Single-layer, unidirectional LSTM with hidden size 384. \\
& Max Pooling & Temporal max pooling over $T$ time steps. \\
\midrule
\multirow{3}{*}{Video} 
& Input & Frame-level features (342 dimensions). \\
& LSTM & Single-layer, unidirectional LSTM with hidden size 384. \\
& Max Pooling & Temporal max pooling over $T$ time steps. \\
\midrule
\multirow{4}{*}{Text} 
& Input & BERT embeddings (1024 dimensions) over $T$ segments. \\
& Conv2D (3 branches) & 1D convolutions with kernel sizes 3, 4, and 5. \\
& Concat + Dropout & Concatenation followed by dropout with $p=0.5$. \\
& FC + ReLU & Final projection with ReLU activation. \\
\bottomrule
\end{tabular}}
\vspace{-0.1in}
\end{table}

\begin{table*}[t]
\centering
\scalebox{0.85}{
\begin{subtable}[t]{\textwidth}
\small
\renewcommand{\arraystretch}{1.2}
\resizebox{\textwidth}{!}{
\begin{tabular}{|c|l|ccc|ccc|ccc|}
\hline
\multicolumn{2}{|c|}{\multirow{1}{*}{Agent ratios}} & \multicolumn{3}{c|}{$6:6:18$} & \multicolumn{3}{c|}{$10:10:10$} & \multicolumn{3}{c|}{$13:13:4$} \\
\cline{1-11}
\multicolumn{2}{|c|}{\multirow{1}{*}{\textbf{Agent types}}} & [\textbf{A}] & [\textbf{G}] &[\textbf{AG}] &[\textbf{A}] & [\textbf{G}] &[\textbf{AG}]  &[\textbf{A}] & [\textbf{G}] &[\textbf{AG}]  \\
\hline
\multirow{7}{*}{\rotatebox{90}{\textbf{KU-HAR}}} 
& DSGD-Modality &
\heatmaplevel{6} 80.1$\pm$0.7 &
\heatmaplevel{6} {68.4$\pm$1.4} &
\heatmaplevel{4} 85.2$\pm$1.7 &
\heatmaplevel{4} 77.3$\pm$0.9 &
\heatmaplevel{6} {68.1$\pm$0.5} &
\heatmaplevel{2} 83.9$\pm$1.0 &
\heatmaplevel{5} 78.6$\pm$1.3 &
\heatmaplevel{6} {71.9$\pm$1.6} &
\heatmaplevel{2} 83.1$\pm$1.5 \\
& DSGD-Task &
\heatmaplevel{2} 74.8$\pm$0.6 &
\heatmaplevel{3} 64.4$\pm$2.3 &
\heatmaplevel{2} 83.2$\pm$0.6 &
\heatmaplevel{5} 77.4$\pm$0.4 &
\heatmaplevel{4} 67.4$\pm$1.2 &
\heatmaplevel{1} 80.2$\pm$0.5 &
\heatmaplevel{2} 76.9$\pm$0.5 &
\heatmaplevel{3} 70.7$\pm$1.4 &
\heatmaplevel{1} 77.8$\pm$0.9 \\
& DSGD-Hybrid &
\heatmaplevel{1} 74.3$\pm$0.8 &
\heatmaplevel{1} 62.9$\pm$1.4 &
\heatmaplevel{3} 84.4$\pm$1.3 &
\heatmaplevel{1} 75.3$\pm$0.7 &
\heatmaplevel{1} 64.9$\pm$1.2 &
\heatmaplevel{4} 85.3$\pm$0.4 &
\heatmaplevel{3} 77.3$\pm$0.3 &
\heatmaplevel{2} 70.6$\pm$0.5 &
\heatmaplevel{4} 86.0$\pm$1.2 \\
& Harmony &
\heatmaplevel{5} 78.9$\pm$0.6 &
\heatmaplevel{5} 67.3$\pm$0.7 &
\heatmaplevel{6} {87.4$\pm$0.6} &
\heatmaplevel{6} {77.8$\pm$1.2} &
\heatmaplevel{5} 67.7$\pm$1.2 &
\heatmaplevel{5} 87.5$\pm$0.8 &
\heatmaplevel{4} 77.6$\pm$0.6 &
\heatmaplevel{5} 71.5$\pm$0.7 &
\heatmaplevel{6} {88.1$\pm$0.5} \\
& FedHGB &
\heatmaplevel{3} 76.3$\pm$0.8 &
\heatmaplevel{4} 66.8$\pm$1.0 &
\heatmaplevel{1} 82.6$\pm$0.8 &
\heatmaplevel{2} 76.1$\pm$0.7 &
\heatmaplevel{2} 65.6$\pm$0.4 &
\heatmaplevel{3} 84.5$\pm$0.7 &
\heatmaplevel{1} 76.5$\pm$0.5 &
\heatmaplevel{4} 71.1$\pm$0.7 &
\heatmaplevel{3} 84.6$\pm$0.5 \\
& DMML-KD &
\heatmaplevel{4} 78.3$\pm$0.4 &
\heatmaplevel{2} 63.0$\pm$1.1 &
\heatmaplevel{5} 86.5$\pm$0.6 &
\heatmaplevel{3} 76.9$\pm$0.6 &
\heatmaplevel{3} 66.1$\pm$0.5 &
\heatmaplevel{6} {88.0$\pm$0.9} &
\heatmaplevel{6} {79.0$\pm$0.7} &
\heatmaplevel{1} 70.4$\pm$0.5 &
\heatmaplevel{5} 87.2$\pm$0.6 \\
& \PARSE &
\heatmaplevel{7} {80.6$\pm$1.0} &
\heatmaplevel{6} {68.4$\pm$1.1} &
\heatmaplevel{7} {88.1$\pm$0.4} &
\heatmaplevel{7} {79.4$\pm$0.2} &
\heatmaplevel{7} {68.4$\pm$0.2} &
\heatmaplevel{7} {88.6$\pm$0.5} &
\heatmaplevel{7} {80.9$\pm$0.4} &
\heatmaplevel{7} {73.1$\pm$0.7} &
\heatmaplevel{7} {88.4$\pm$0.5} \\
\cline{1-11}
\multicolumn{2}{|c|}{\multirow{1}{*}{\textbf{Agent types}}} & [\textbf{V1}] & [\textbf{V2}] &[\textbf{V1,V2}] &[\textbf{V1}] & [\textbf{V2}] &[\textbf{V1,V2}] & [\textbf{V1}] & [\textbf{V2}] &[\textbf{V1,V2}] \\
\hline
\multirow{7}{*}{\rotatebox{90}{\textbf{ModelNet-40}}} 
& DSGD-Modality &
\heatmaplevel{6} {77.8$\pm$0.6} &
\heatmaplevel{6} {71.6$\pm$1.5} &
\heatmaplevel{5} 75.7$\pm$0.8 &
\heatmaplevel{3} 72.4$\pm$2.0 &
\heatmaplevel{6} {67.9$\pm$1.8} &
\heatmaplevel{3} 73.4$\pm$1.0 &
\heatmaplevel{6} {75.2$\pm$2.0} &
\heatmaplevel{5} {69.1$\pm$0.9} &
\heatmaplevel{4} 76.4$\pm$0.7 \\
& DSGD-Task &
\heatmaplevel{2} 72.3$\pm$1.2 &
\heatmaplevel{3} 67.1$\pm$0.9 &
\heatmaplevel{2} 73.3$\pm$1.3 &
\heatmaplevel{4} 73.0$\pm$1.8 &
\heatmaplevel{1} 65.6$\pm$0.8 &
\heatmaplevel{1} 71.4$\pm$0.9 &
\heatmaplevel{4} 73.2$\pm$0.7 &
\heatmaplevel{4} 68.1$\pm$0.7 &
\heatmaplevel{1} 66.2$\pm$1.4 \\
& DSGD-Hybrid &
\heatmaplevel{1} 72.1$\pm$0.5 &
\heatmaplevel{1} 59.8$\pm$1.1 &
\heatmaplevel{1} 72.9$\pm$1.4 &
\heatmaplevel{2} 72.3$\pm$1.2 &
\heatmaplevel{2} 65.8$\pm$0.7 &
\heatmaplevel{2} 72.2$\pm$1.1 &
\heatmaplevel{2} 69.4$\pm$0.7 &
\heatmaplevel{3} 67.7$\pm$1.0 &
\heatmaplevel{2} 73.3$\pm$1.2 \\
& Harmony &
\heatmaplevel{4} 74.1$\pm$1.3 &
\heatmaplevel{5} 71.5$\pm$1.0 &
\heatmaplevel{6} {77.3$\pm$0.4} &
\heatmaplevel{5} 76.4$\pm$2.0 &
\heatmaplevel{5} 67.8$\pm$1.5 &
\heatmaplevel{6} {77.2$\pm$2.3} &
\heatmaplevel{5} 75.1$\pm$1.2 &
\heatmaplevel{5} 69.1$\pm$1.6 &
\heatmaplevel{5} 76.7$\pm$0.5 \\
& FedHGB &
\heatmaplevel{3} 72.6$\pm$0.6 &
\heatmaplevel{2} 60.6$\pm$0.4 &
\heatmaplevel{3} 74.4$\pm$0.5 &
\heatmaplevel{1} 72.2$\pm$1.4 &
\heatmaplevel{4} 67.2$\pm$1.0 &
\heatmaplevel{4} 74.1$\pm$1.6 &
\heatmaplevel{1} 69.3$\pm$1.7 &
\heatmaplevel{2} 65.1$\pm$2.0 &
\heatmaplevel{3} 74.9$\pm$0.9 \\
& DMML-KD &
\heatmaplevel{5} 76.4$\pm$1.0 &
\heatmaplevel{4} 67.6$\pm$1.3 &
\heatmaplevel{4} 75.1$\pm$1.2 &
\heatmaplevel{5} {76.4$\pm$1.4} &
\heatmaplevel{3} 66.9$\pm$2.2 &
\heatmaplevel{5} 76.4$\pm$1.6 &
\heatmaplevel{3} 72.2$\pm$1.5 &
\heatmaplevel{1} 64.3$\pm$0.5 &
\heatmaplevel{6} {78.4$\pm$0.8} \\
& \PARSE &
\heatmaplevel{7} {78.6$\pm$1.2} &
\heatmaplevel{7} {71.9$\pm$0.6} &
\heatmaplevel{7} {78.8$\pm$0.6} &
\heatmaplevel{7} {77.9$\pm$0.4} &
\heatmaplevel{7} {69.8$\pm$1.2} &
\heatmaplevel{7} {79.3$\pm$0.9} &
\heatmaplevel{7} {75.8$\pm$1.0} &
\heatmaplevel{7} {70.9$\pm$1.2} &
\heatmaplevel{7} {81.2$\pm$0.7} \\
\cline{1-11}
\multicolumn{2}{|c|}{\multirow{1}{*}{\textbf{Agent types}}} & [\textbf{A}] & [\textbf{V}] &[\textbf{AV}] &[\textbf{A}] & [\textbf{V}] &[\textbf{AV}]  &[\textbf{A}] & [\textbf{V}] &[\textbf{AV}]  \\
\hline
\multirow{7}{*}{\rotatebox{90}{\textbf{AVE}}} 
& DSGD-Modality &
\heatmaplevel{6} {46.1$\pm$0.3} &
\heatmaplevel{6} {52.4$\pm$0.2} &
\heatmaplevel{4} 63.4$\pm$0.7 &
\heatmaplevel{6} {44.8$\pm$0.6} &
\heatmaplevel{6} {52.2$\pm$0.4} &
\heatmaplevel{4} 61.4$\pm$1.0 &
\heatmaplevel{5} 43.7$\pm$0.3 &
\heatmaplevel{5} 50.3$\pm$0.2 &
\heatmaplevel{3} 60.9$\pm$0.6 \\
& DSGD-Task &
\heatmaplevel{1} 35.3$\pm$0.7 &
\heatmaplevel{2} 43.3$\pm$0.5 &
\heatmaplevel{1} 58.5$\pm$0.3 &
\heatmaplevel{2} 41.0$\pm$0.7 &
\heatmaplevel{2} 49.1$\pm$0.9 &
\heatmaplevel{1} 56.6$\pm$0.8 &
\heatmaplevel{2} 42.6$\pm$0.3 &
\heatmaplevel{2} 49.6$\pm$1.0 &
\heatmaplevel{1} 50.6$\pm$1.0 \\
& DSGD-Hybrid &
\heatmaplevel{2} 37.3$\pm$0.9 &
\heatmaplevel{3} 45.7$\pm$0.8 &
\heatmaplevel{2} 60.7$\pm$0.6 &
\heatmaplevel{1} 38.1$\pm$1.4 &
\heatmaplevel{3} 50.3$\pm$0.9 &
\heatmaplevel{3} 61.0$\pm$0.7 &
\heatmaplevel{1} 41.4$\pm$0.7 &
\heatmaplevel{3} 50.1$\pm$0.4 &
\heatmaplevel{5} 62.1$\pm$0.8 \\
& Harmony &
\heatmaplevel{5} 44.1$\pm$0.5 &
\heatmaplevel{5} 49.4$\pm$1.0 &
\heatmaplevel{6} {64.6$\pm$0.6} &
\heatmaplevel{5} 42.2$\pm$0.8 &
\heatmaplevel{5} 51.7$\pm$1.1 &
\heatmaplevel{6} {64.5$\pm$0.8} &
\heatmaplevel{2} 42.6$\pm$0.7 &
\heatmaplevel{3} 50.1$\pm$1.2 &
\heatmaplevel{2} 60.3$\pm$0.7 \\
& FedHGB &
\heatmaplevel{4} 42.0$\pm$0.4 &
\heatmaplevel{4} 47.6$\pm$0.5 &
\heatmaplevel{3} 63.1$\pm$0.3 &
\heatmaplevel{4} 41.4$\pm$1.3 &
\heatmaplevel{4} 51.1$\pm$0.7 &
\heatmaplevel{2} 60.6$\pm$0.9 &
\heatmaplevel{6} {44.6$\pm$0.4} &
\heatmaplevel{6} {50.6$\pm$0.5} &
\heatmaplevel{3} 60.9$\pm$0.8 \\
& DMML-KD &
\heatmaplevel{3} 38.8$\pm$1.1 &
\heatmaplevel{1} 41.2$\pm$0.4 &
\heatmaplevel{5} 64.1$\pm$0.5 &
\heatmaplevel{2} 41.3$\pm$0.9 &
\heatmaplevel{1} 43.5$\pm$0.7 &
\heatmaplevel{5} 63.7$\pm$0.5 &
\heatmaplevel{4} 43.4$\pm$0.9 &
\heatmaplevel{1} 43.1$\pm$0.6 &
\heatmaplevel{6} {62.8$\pm$0.8} \\
& \PARSE &
\heatmaplevel{7} {47.2$\pm2.1$} &
\heatmaplevel{7} {53.3$\pm$1.5} &
\heatmaplevel{7} {65.1$\pm$1.1} &
\heatmaplevel{7} {45.6$\pm$1.8} &
\heatmaplevel{7} {52.7$\pm$0.3} &
\heatmaplevel{7} {64.7$\pm$1.3} &
\heatmaplevel{7} {45.3$\pm$0.8} &
\heatmaplevel{7} {53.2$\pm$1.6} &
\heatmaplevel{7} {64.3$\pm$1.2} \\
\hline
\end{tabular}}
\end{subtable}}
\scalebox{0.85}{
\begin{subtable}[t]{0.998\textwidth}
\centering
\small
\renewcommand{\arraystretch}{1.2}
\resizebox{\textwidth}{!}{
\begin{tabular}{|c|l|cccc|cccc|cccc|}
\hline
\multicolumn{2}{|c|}{\multirow{1}{*}{Agent ratios}} & \multicolumn{4}{c|}{$6:6:6:22$} & \multicolumn{4}{c|}{$10:10:10:10$} & \multicolumn{4}{c|}{$12:12:12:4$} \\
\cline{1-14}
\multicolumn{2}{|c|}{\multirow{1}{*}{\textbf{Agent types}}} & [\textbf{A}] & [\textbf{V}] & [\textbf{T}] &[\textbf{AVT}] &[\textbf{A}] & [\textbf{V}] & [\textbf{T}] &[\textbf{AVT}]  &[\textbf{A}] & [\textbf{V}] & [\textbf{T}] &[\textbf{AVT}]  \\
\hline
\multirow{7}{*}{\rotatebox{90}{\textbf{IEMOCAP}}} 
& DSGD-Modality &
\heatmaplevel{6} 47.7$\pm$0.6 &
\heatmaplevel{4} 44.2$\pm$1.1 &
\heatmaplevel{6} 60.3$\pm$0.4 &
\heatmaplevel{1} 64.5$\pm$0.6 &
\heatmaplevel{5} 46.5$\pm$1.1 &
\heatmaplevel{6} {50.3$\pm$0.4} &
\heatmaplevel{6} {58.6$\pm$0.3} &
\heatmaplevel{1} 65.1$\pm$1.2 &
\heatmaplevel{1} 46.0$\pm$0.7 &
\heatmaplevel{6} {53.9$\pm$1.2} &
\heatmaplevel{6} {57.6$\pm$0.5} &
\heatmaplevel{1} 63.6$\pm$0.8 \\
& DSGD-Task &
\heatmaplevel{4} 47.2$\pm$0.3 &
\heatmaplevel{5} 44.9$\pm$0.5 &
\heatmaplevel{2} 57.2$\pm$0.8 &
\heatmaplevel{5} 68.8$\pm$0.6 &
\heatmaplevel{2} 46.0$\pm$1.2 &
\heatmaplevel{3} 50.1$\pm$0.3 &
\heatmaplevel{1} 55.4$\pm$1.1 &
\heatmaplevel{5} 68.4$\pm$0.6 &
\heatmaplevel{4} 48.1$\pm$0.3 &
\heatmaplevel{3} 53.3$\pm$0.5 &
\heatmaplevel{4} 57.1$\pm$0.2 &
\heatmaplevel{2} 64.6$\pm$0.4 \\
& DSGD-Hybrid &
\heatmaplevel{2} 45.8$\pm$0.5 &
\heatmaplevel{6} {45.5$\pm$0.3} &
\heatmaplevel{3} 60.1$\pm$0.9 &
\heatmaplevel{4} 68.3$\pm$1.1 &
\heatmaplevel{2} 46.0$\pm$0.6 &
\heatmaplevel{4} 50.2$\pm$0.5 &
\heatmaplevel{5} 58.3$\pm$0.9 &
\heatmaplevel{4} 68.3$\pm$1.3 &
\heatmaplevel{5} 48.3$\pm$0.6 &
\heatmaplevel{5} 53.8$\pm$0.4 &
\heatmaplevel{5} 57.4$\pm$0.8 &
\heatmaplevel{5} 68.4$\pm$0.7 \\
& Harmony &
\heatmaplevel{5} 47.6$\pm$1.3 &
\heatmaplevel{2} 43.9$\pm$0.7 &
\heatmaplevel{5} 60.2$\pm$0.6 &
\heatmaplevel{3} 65.7$\pm$0.8 &
\heatmaplevel{1} 45.6$\pm$1.5 &
\heatmaplevel{4} 50.2$\pm$0.7 &
\heatmaplevel{3} 57.8$\pm$1.8 &
\heatmaplevel{3} 67.1$\pm$1.3 &
\heatmaplevel{3} 47.5$\pm$0.8 &
\heatmaplevel{4} 53.4$\pm$0.5 &
\heatmaplevel{3} 57.0$\pm$0.8 &
\heatmaplevel{3} 68.0$\pm$1.2 \\
& FedHGB &
\heatmaplevel{3} 46.5$\pm$1.1 &
\heatmaplevel{1} 43.3$\pm$1.6 &
\heatmaplevel{3} 60.1$\pm$0.6 &
\heatmaplevel{2} 65.6$\pm$1.1 &
\heatmaplevel{4} 46.1$\pm$0.7 &
\heatmaplevel{2} 49.7$\pm$1.0 &
\heatmaplevel{3} 57.8$\pm$0.6 &
\heatmaplevel{2} 66.7$\pm$0.8 &
\heatmaplevel{2} 47.3$\pm$0.8 &
\heatmaplevel{2} 53.2$\pm$1.1 &
\heatmaplevel{2} 56.3$\pm$0.8 &
\heatmaplevel{4} 68.2$\pm$1.2 \\
& DMML-KD &
\heatmaplevel{1} 45.2$\pm$1.1 &
\heatmaplevel{2} 43.9$\pm$1.3 &
\heatmaplevel{1} 55.7$\pm$1.3 &
\heatmaplevel{6} {71.4$\pm$0.8} &
\heatmaplevel{6} {47.9$\pm$0.7} &
\heatmaplevel{1} 48.6$\pm$1.2 &
\heatmaplevel{2} 56.7$\pm$0.5 &
\heatmaplevel{6} {71.3$\pm$1.4} &
\heatmaplevel{6} {50.2$\pm$0.5} &
\heatmaplevel{1} 50.1$\pm$1.0 &
\heatmaplevel{1} 51.6$\pm$0.6 &
\heatmaplevel{6} {71.2$\pm$0.8} \\
& \PARSE &
\heatmaplevel{7} {48.2$\pm$0.6} &
\heatmaplevel{7} {47.2$\pm$0.8} &
\heatmaplevel{7} {61.4$\pm$0.4} &
\heatmaplevel{7} {73.6$\pm$0.2} &
\heatmaplevel{7} {48.9$\pm$1.2} &
\heatmaplevel{7} {50.9$\pm$1.2} &
\heatmaplevel{7} {60.3$\pm$0.2} &
\heatmaplevel{7} {73.2$\pm$0.6} &
\heatmaplevel{7} {50.5$\pm$1.3} &
\heatmaplevel{7} {54.3$\pm$1.4} &
\heatmaplevel{7} {58.8$\pm$1.0} &
\heatmaplevel{7} {74.3$\pm$0.3} \\
\hline
\end{tabular}}
\end{subtable}}
\caption{Performance (accuracy \%) of methods under varying agent-ratio scenarios (Dirichlet $\alpha=0.5$, ring topology). Cell shading indicates each method’s relative performance within each column.
}
\label{tab:main_results}
\vspace{-0.15in}
\end{table*}

\subsection{Training Setup}

{\bf Datasets.}
For KU-HAR/ModelNet40/AVE, we use SGD (momentum 0.9, weight decay $5{\times}10^{-4}$) with learning rates 0.1/0.01/0.005 and batch sizes 64/64/32, respectively. For IEMOCAP, we use Adam (lr $5{\times}10^{-5}$, $\beta{=}(0.9,0.999)$, batch size 64). Unless stated otherwise, we use an 80/20 train/test split.

{\bf Methods.}
\PARSE splits the feature dimension into three equal slices: 384$\rightarrow$128 (IEMOCAP, AVE), 300$\rightarrow$100 (ModelNet40), and 192$\rightarrow$64 (KU-HAR). We set the contrastive temperature to 0.2 for all experiments. For Harmony, we follow the two-stage schedule (150 rounds modality-independent + 50 rounds modality-joint). For FedHGB, we hold out 10\% of training data for validation. For DMML-KD, we use the same feature aggregator as in~\cite{kdfl}.

{\bf Non-IID Distribution.} 
We assign each agent a label distribution drawn from a Dirichlet distribution with concentration parameter $\alpha$ \cite{hsu2019measuring}. We simulate 30 agents (40 on IEMOCAP) and report our main results with $\alpha = 0.5$.

\begin{figure*}[t]
  \centering
  \begin{subfigure}[t]{0.25\textwidth}
    \centering\includegraphics[width=\linewidth]{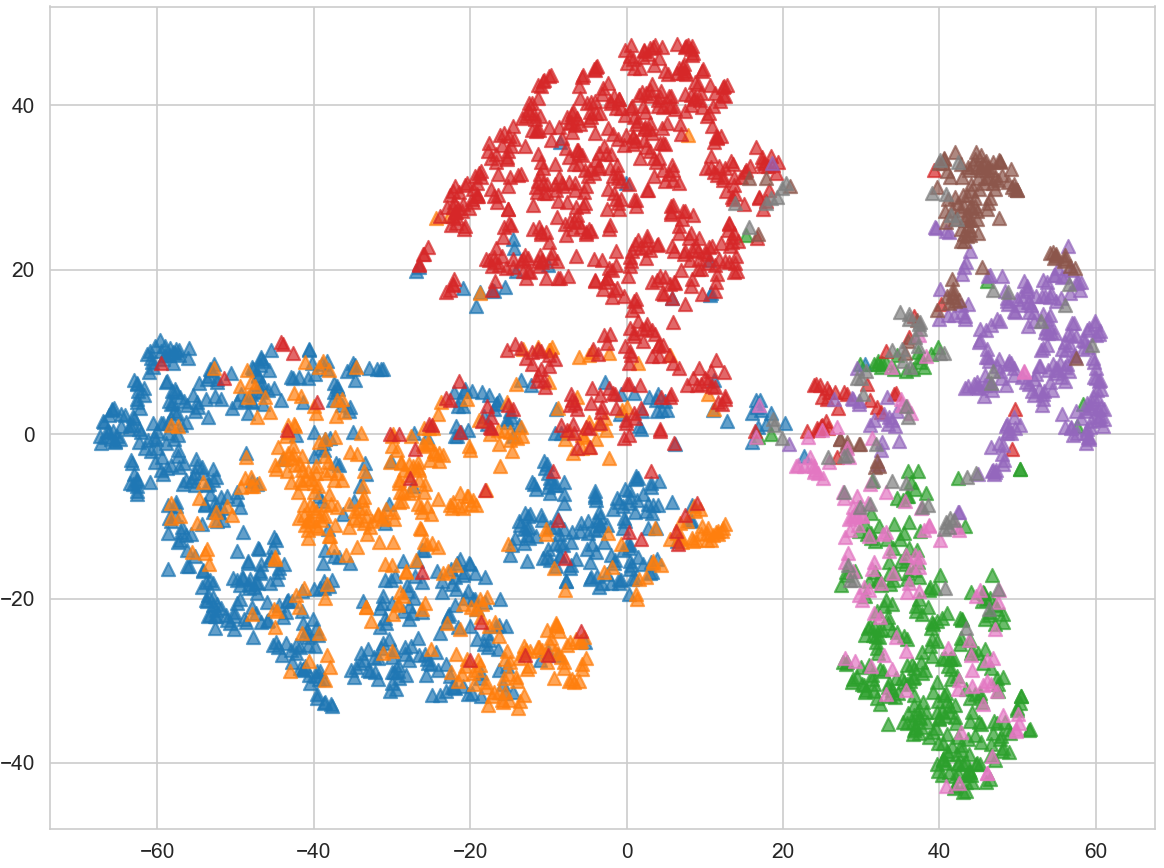}
    \subcaption{Accelerometer.}\label{fig:feat_har_acc}
  \end{subfigure}
  \begin{subfigure}[t]{0.25\textwidth}
    \centering\includegraphics[width=\linewidth]{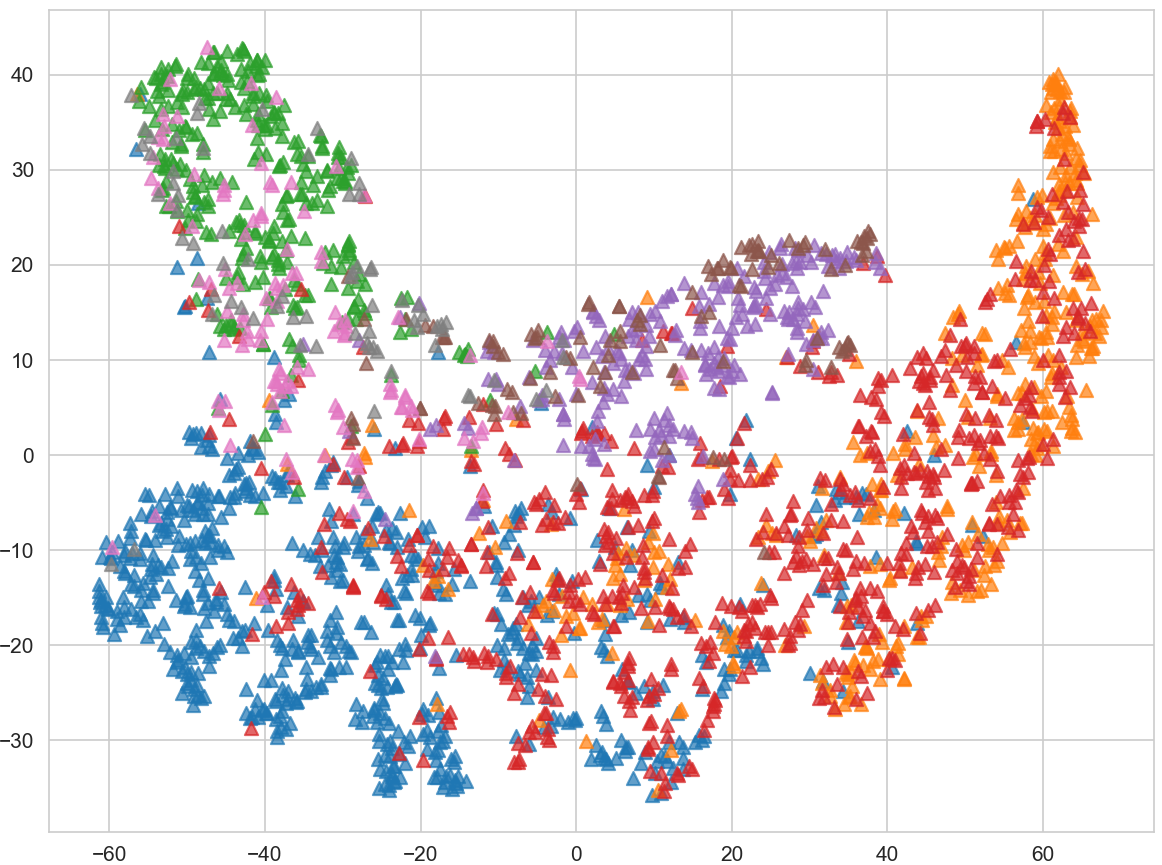}
    \subcaption{Gyroscope.}\label{fig:feat_har_gyro}
  \end{subfigure}
  \begin{subfigure}[t]{0.25\textwidth}
    \centering\includegraphics[width=\linewidth]{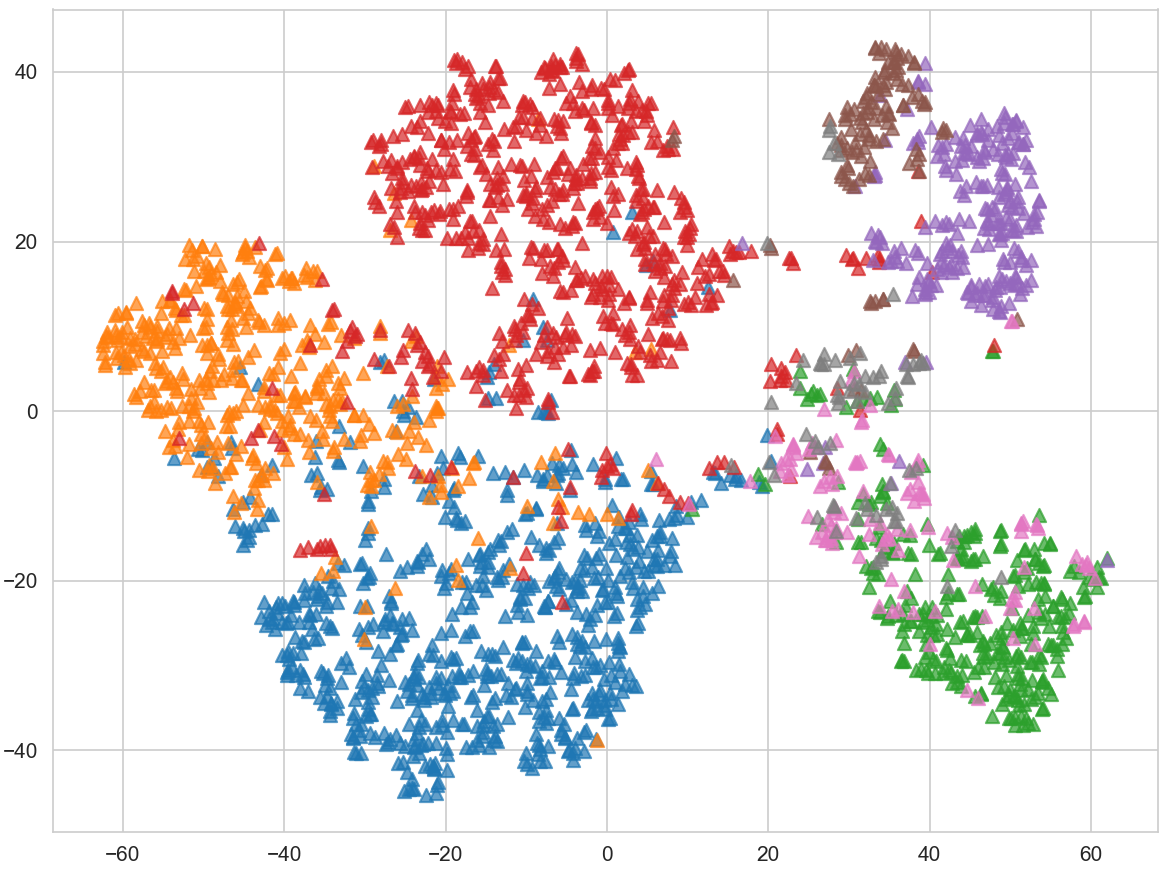}
    \subcaption{Fused features.}\label{fig:feat_har_both}
  \end{subfigure}
  \caption{t-SNE visualization of synergistic features from each modality and after fusion.}
  \label{fig:visualization_har}
  \vspace{-0.1in}
\end{figure*}

\subsection{Main Results}

We evaluate our \PARSE prototype and all baselines on four datasets and report accuracy by agent type. We denote types by their available modalities, e.g., [A] for audio-only and [AV] for audio and visual signals. Each experiment specifies an agent mix (the counts of each type).  
Table~\ref{tab:main_results} reports the mean test accuracy for every type. The results reveal several consistent trends:

\textbf{Consistent Improvement.} On KU-HAR, \PARSE outperforms all baselines by about 0.3–0.6\% across every agent type. On ModelNet-40 and AVE, gains grow when multimodal agents are scarce, reaching {1.8–2.8\%} for both multimodal and the most impacted unimodal groups.
On IEMOCAP, \PARSE yields {1.9–3.1\%} higher accuracy on multimodal agents across all splits. Whereas existing methods tend to favor either unimodal agents (e.g., DSGD-Modality) or multimodal agents (e.g., Harmony), \textit{\PARSE provides consistent benefits to every agent type.} Consistent improvements are observed under other Dirichlet parameters, e.g., $\alpha=0.1, 5$ and topologies, e.g., Chordal Ring and Random Gossip, and are omitted for brevity.

\textbf{Modality-mix generalization.}
Beyond the unimodal vs. fully-multimodal splits reported in Table~\ref{tab:main_results}, we further evaluate IEMOCAP under richer modality-mix regimes that enumerate all modality subsets (e.g., [A], [V], [T], [AV], [AT], [VT], [AVT]) and multiple agent-ratio configurations. \PARSE remains consistently competitive across modality subsets and provides the largest gains when multimodal agents are scarce, indicating that slice-aware partial alignment sustains transfer under fine-grained modality heterogeneity. Full modality-mix results are reported in Appendix~\ref{app:exp_results}.

\textbf{Modality Heterogeneity.}
Like DMML-KD, \PARSE disentangles redundant and unique features. In addition, it \emph{explicitly} models synergistic subspace. As a result, it consistently attains the highest multimodal accuracy  across all agent-ratio settings, while keeping unimodal performance stable as the overall modality mix varies. Concretely, Fig.~\ref{fig:visualization_har} shows t-SNE plots of per-modality synergistic features (pre-fusion) with their fused counterpart on KU-HAR: before fusion, classes remain interleaved (e.g., blue-yellow in Fig.~\ref{fig:visualization_har}(a), red-yellow in Fig.~\ref{fig:visualization_har}(b)), whereas fusion yields well-separated clusters with larger margins. This indicates \PARSE captures complementary cross-modal structure in synergistic slice rather than merely aligning features.  Similar observations hold for different modalities under AVE, ModelNet-40 and IEMOCAP datasets and hence are omitted here due to space constraint.

\subsection{Ablations}\label{sec:ablations}
Unless stated otherwise, all results in Sections~\ref{sec:ablations} and \ref{sec:sensitivity} are obtained on KU-HAR under the default setting: a 10:10:10 agent ratio and a non-IID Dirichlet split with $\alpha=0.5$. We observed the same qualitative trends on the other datasets and settings in Section~IV, and hence we omit the additional experimental results for brevity.

\textbf{Feature Split Ratios.} To access how split ratio affect performance, we run a split-sweep study with total feature dimension 192. In each sweep, we vary one branch's dimensionality  and divide the remaining budget equally between the other two.  Table \ref{tab:splits_har} reports 
(i) the overall accuracy of multimodal agents when the branch is included in the full model 
and (ii) the split-only accuracy when using that branch alone. Enlarging the unique branch from $32d$ to $128d$ improves its stand-alone accuracy by about $+3$ pp, but the overall accuracy remains flat (88.5–88.8).  Oversizing the redundant branch reduces overall accuracy (88.7 to 87.0).  The synergistic branch helps most at a moderate size (peak overall 89.5 at $96d$). A balanced allocation across unique, redundant, and synergistic slices (our default even split) offers a robust trade-off. 

\begin{table}
\centering
\caption{Accuracy vs. feature split size.}
\label{tab:splits_har}
\setlength{\tabcolsep}{4pt}
\resizebox{0.9\linewidth}{!}{
\begin{tabular}{llcccc}
\toprule
\textbf{Varying Split} & \textbf{Metric} & \textbf{32d} & \textbf{64d} & \textbf{96d} & \textbf{128d} \\
\midrule
\multirow{2}{*}{Unique}
  & Unique-only & 83.5 & 86.2 & \bf{86.7} & 86.1 \\
  & Combined    & 87.8 & 88.6 & \bf{88.8} & 88.5 \\
\midrule
\multirow{2}{*}{Redundant}
  & Redundant-only & 87.2 & \bf{87.8} & 87.4 & 87.6 \\
  & Combined       & \bf{88.7} & 88.6 & 87.6 & 87.0 \\
\midrule
\multirow{2}{*}{Synergistic}
  & Synergistic-only & 54.5 & 61.5 & 63.7 & \bf{67.2} \\
  & Combined         & 86.9 & 88.6 & \bf{89.5} & 88.3 \\
\bottomrule
\end{tabular}}
\end{table}

\begin{table}
\centering
\captionof{table}{Comparison of fusion methods.}
\label{tab:fusion_comparison_har}
\setlength{\tabcolsep}{2.5pt}
\resizebox{0.9\linewidth}{!}{
\begin{tabular}{lcc}
\toprule
\textbf{Fusion Method} & \textbf{Overall (\%)} & \textbf{Synergy (\%)} \\
\midrule
Mean (default)         & 88.6$\pm$0.5 & 61.5$\pm$1.0 \\
Concatenation + Linear & 87.1$\pm$0.9 & 60.3$\pm$0.8 \\
Summation + Linear     & 89.0$\pm$1.1 & 62.2$\pm$2.1 \\
Gated Fusion           & 89.0$\pm$1.4 & \bf{64.7}$\pm$1.5 \\
Cross-Attention        & \bf{89.2}$\pm$1.2 & 62.1$\pm$0.7 \\
Hadamard Product       & 88.6$\pm$1.3 & 58.7$\pm$1.8 \\
\bottomrule
\end{tabular}}
\end{table}

\textbf{Synergy Fusion Operator.} By default, we fuse synergistic features by simple averaging. To test whether it is too crude, we compare mean fusion with five stronger operators: concatenation+Linear, summation+Linear, gated fusion \cite{Xue_2023_CVPR}, cross attention \cite{zhang_2022_MM}, and Hadamard product \cite{Kim2017}, where ``Linear'' denotes a trainable linear layer. 
As reported in Table~\ref{tab:fusion_comparison_har}, gated fusion yields the highest synergistic-only accuracy, and cross-attention give a slight overall gain over mean. However, these operators introduce extra  parameters that must be exchanged and synchronized across agents each round. Mean fusion avoids this coordination cost while remaining competitive, making it a practical default for DFL. 

\begin{figure}
  \centering
  \includegraphics[width=0.8\linewidth]{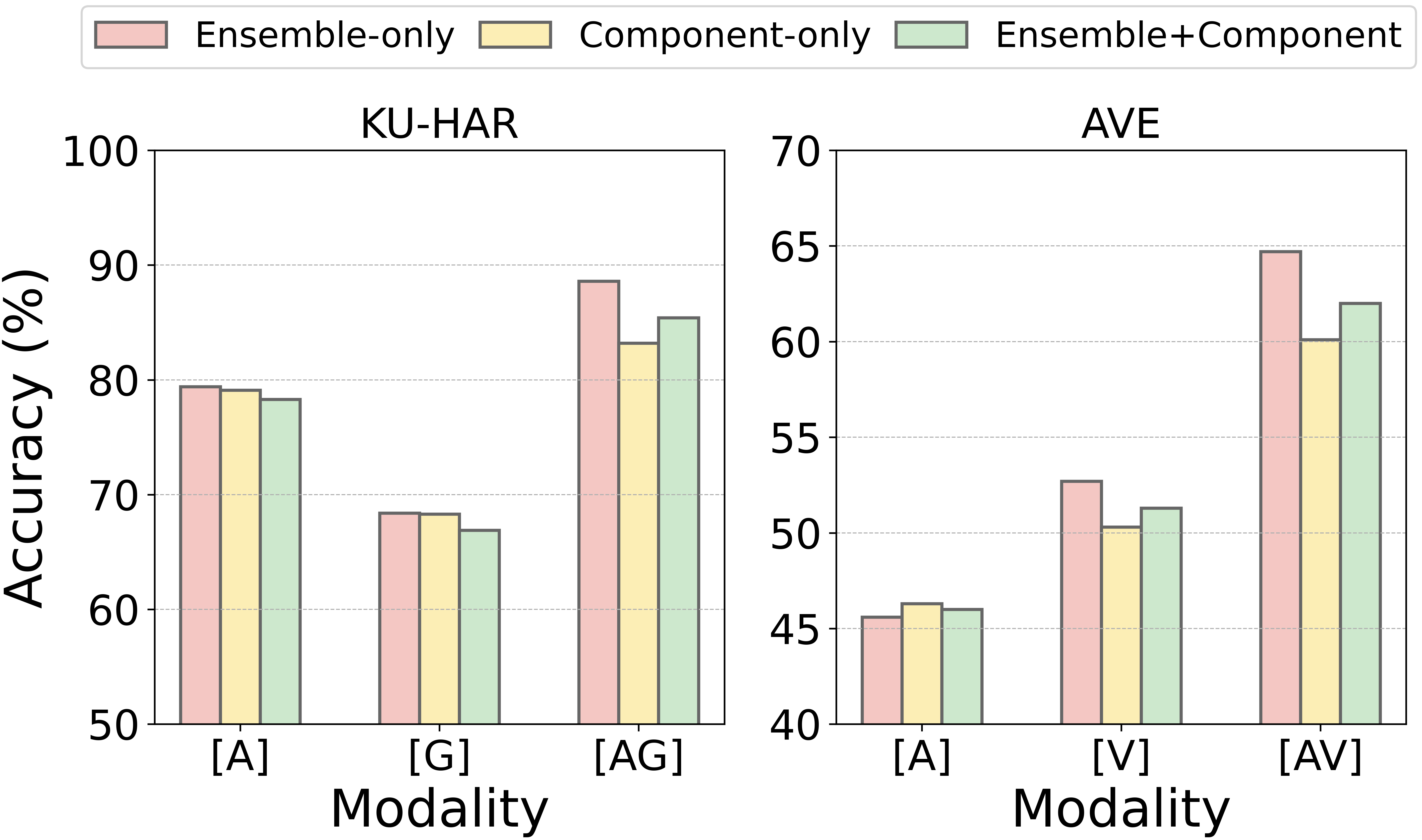}
  \caption{Impact of ensemble predictions on classification loss.
  }
  \vspace{-0.2in}
  \label{fig:ensemble_ablation}
\end{figure}

\textbf{Training Objective (Ensemble vs Component Losses).} We compare three schemes: \textit{Ensemble-only} (ours): a single loss on the sum of all three component predictions; \textit{Component-only}: independent losses for each component; and \textit{Ensemble+Component}: the ensemble loss plus three per-component losses. Fig.~\ref{fig:ensemble_ablation} reports the average accuracy. Ensemble-only achieves the highest accuracy for both unimodal and multimodal agents. Component-only slightly reduces unimodal accuracy and substantially degrades multimodal accuracy. Adding per-component losses recovers part of the multimodal drop but hurts unimodal performance. These results indicate that an ensemble-only objective provides the most stable signal by dampening gradient noise from unaligned components. 

\subsection{Sensitivity Analyses}\label{sec:sensitivity}

\textbf{Sensitivity to Data Heterogeneity (Dirichlet $\alpha$).} 
We assess robustness to data heterogeneity by varying the Dirichlet parameter $\alpha.$ In addition to the default $\alpha=0.5$, we consider a near-IID setting ($\alpha=5.0$) and a highly skewed non-IID setting ($\alpha=0.1$). Table~\ref{tab:ablation-non-iid} reports mean accuracies. When the data are near-IID or mildly heterogeneous ($\alpha=5.0$ and $0.5$), \PARSE exceeds the strongest competitor by 1.1\%--1.5\%. Under severe heterogeneity ($\alpha=0.1$), the margin widens to about 2.9\%, indicating that feature fission and partial alignment remains effective even in the most challenging non-IID regimes.

\begin{table}
\centering
\captionof{table}{Comparisons across non-IID settings.}
\label{tab:ablation-non-iid}
\resizebox{0.9\linewidth}{!}{
\begin{tabular}{@{}l|ccc}
\toprule
\multirow{1}{*}{\textbf{Methods}} & $\bm{\alpha=5.0}$ & $\bm{\alpha=0.5}$ & $\bm{\alpha=0.1}$ \\
\midrule
DSGD-Modality & 83.00 & 76.43 & 53.23 \\
DSGD-Task     & 82.33 & 75.00 & 52.07 \\
DSGD-Hybrid   & 81.33 & 75.17 & 50.67\\
Harmony       & \heatmaplevel{5}83.43 & \heatmaplevel{5}77.67 & \heatmaplevel{5}56.43 \\
FedHGB        & 82.17 & 75.40 & 49.57 \\
DMML-KD       & 81.67 & 77.00 & 55.00 \\
\PARSE         & \heatmaplevel{7} 84.97 & \heatmaplevel{7}78.80 & \heatmaplevel{7}59.23  \\
\hline
\end{tabular}}
\end{table}

\begin{table}
\centering
\caption{Accuracy under different $\beta$ values.}
\label{tab:beta_ablation}
\resizebox{0.9\linewidth}{!}{ 
\begin{tabular}{@{}l|ccc@{}}
\toprule
\multirow{2}{*}{$\beta$} & \multicolumn{3}{c}{\textbf{Accuracy (\%)}} \\
\cmidrule{2-4}
& Unimodal [\textbf{A}] & Unimodal [\textbf{G}] & Multimodal [\textbf{AG}] \\
\midrule
$0.0$      & 79.3$\pm$0.8 & 67.5$\pm$0.5 & 87.6$\pm$0.3 \\
$0.1$    & 79.1$\pm$0.4 & 67.9$\pm$0.5 & 87.8$\pm$0.2 \\
$0.2$ & 79.4$\pm$0.2 & \textbf{68.4}$\pm$0.2 & \textbf{88.6}$\pm$0.5 \\
$0.3$    & \textbf{79.6$\pm$0.5} & 67.7$\pm$0.3 & 88.4$\pm$0.6 \\
$0.4$    & 78.6$\pm$0.2 & 67.2$\pm$0.7 & 87.9$\pm$0.4 \\
\bottomrule
\end{tabular}}
\vspace{-0.1in}
\end{table}

\textbf{Sensitivity to Contrastive Weight $\beta.$}
We vary the coefficient $\beta$ in Eq.~\ref{eq:final_loss}, which scales the contrastive-diversity term for multimodal agents, from $0$ to $0.4$, and report accuracies in~Table~\ref{tab:beta_ablation}. Setting $\beta=0.2$ consistently improves both uni- and multimodal accuracy, indicating that a moderate contrastive signal strengthens cross-modal consistency. For $\beta>0.2$, multimodal accuracy declines because the contrastive objective begins to dominate, emphasizing modality-shared but task-irrelevant patterns and hindering supervised learning. 
\begin{figure}
  \centering
   \includegraphics[width=\linewidth]{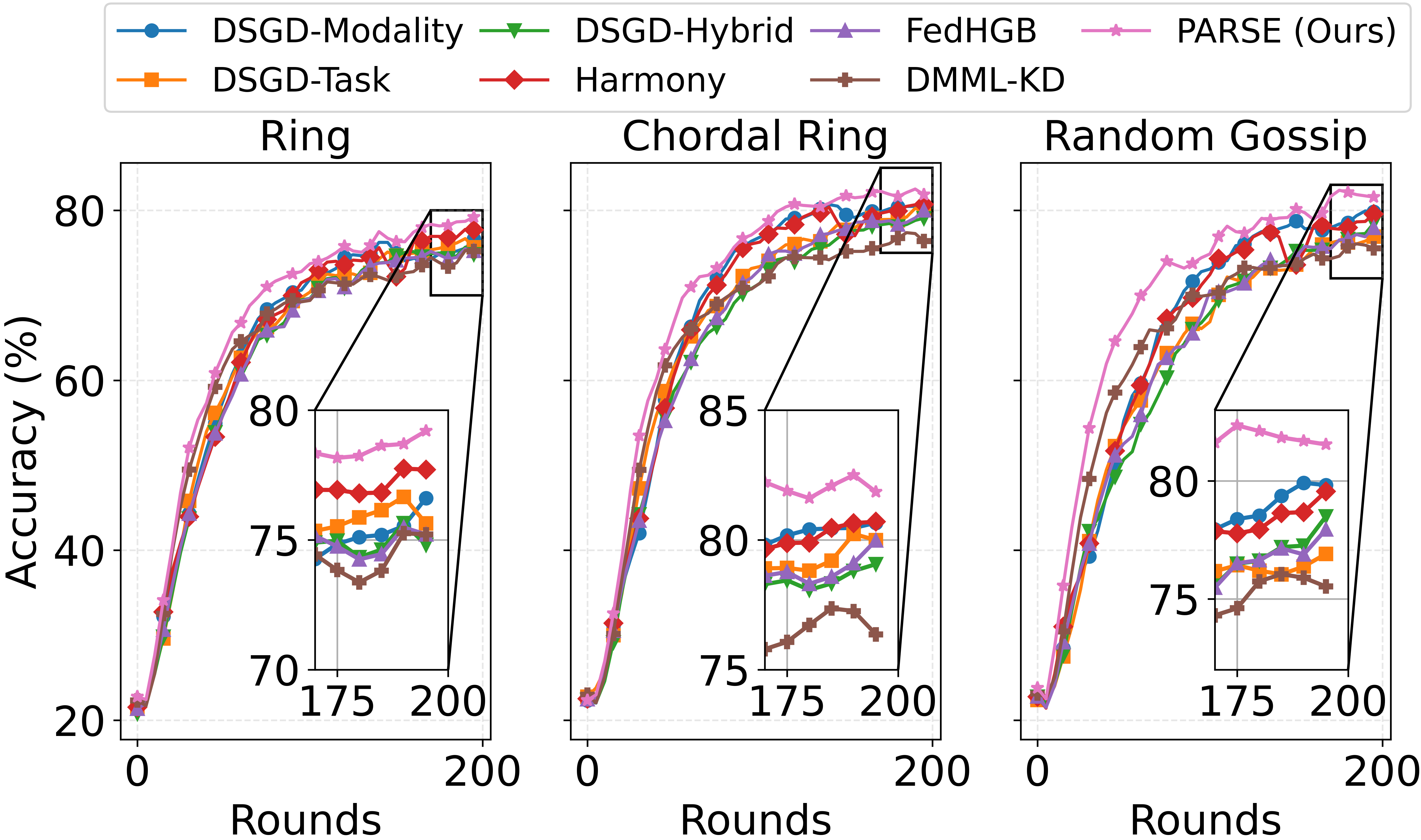}
  \caption{Comparison of different methods under various communication topologies.}
  \label{fig:topology_ablation}
\end{figure}

\textbf{Overlay Topology Sensitivity.}
We evaluate \PARSE and all baselines on three P2P overlay topologies:  \textit{Ring}~\cite{ring}; \textit{Chordal Ring}~\cite{ahmad2022studies}, which augments each node with a diametrically opposite link and roughly halves the diameter; and \textit{Random Gossip}~\cite{de2023epidemic}, where each  round each agent exchanges with two random peers from the same group.  Fig.~\ref{fig:topology_ablation} reports the average accuracy over agents. \PARSE converges faster and reaches higher final accuracy than all competitors on every topology. Gains are largest on Chordal Ring and Random-Gossip, indicating that \PARSE benefits from shorter path lengths and quicker consensus.

\subsection{Comparison with Centralized Federated Learning} \label{app:fl}

Although \PARSE is designed for server-free decentralized FL (DFL), it is natural to ask whether its slice-aware decomposition also benefits \emph{server-based} aggregation when a reliable coordinator is available. We therefore evaluate \PARSE in a centralized FL setting and compare it against representative server-coordinated multimodal FL methods, including FedMSplit~\cite{chen2022fedmsplit}, which learns inter-client relationship graphs for aggregation, FedMVD~\cite{gao2025multimodal}, which performs global alignment to mitigate modality-induced domain shifts, and FedMVC~\cite{chen2024bridging}, a multi-view approach for reducing modality heterogeneity.  

To isolate the effect of aggregation regime and for fair comparison, we keep the local training pipeline and model capacity comparable across methods and use the same client participation and round schedule; centralized methods additionally use a server for coordination as required by their designs.
As shown in Table~\ref{tab:fl_comp_har}, even without introducing any additional server-specific modules, \PARSE achieves the best performance on multimodal agents and remains competitive on unimodal agents across the evaluated benchmarks. These results suggest that \PARSE’s feature fission and partial alignment provide a \emph{portable slice interface} across aggregation regimes (P2P consensus vs. server aggregation), which is useful for real deployments that may switch between infrastructure modes (e.g., intermittent coordinator availability or fallback to server-free operation).

\begin{table}[t]
\caption{Comparing methods on four benchmarks in a federated learning setting. We report accuracy (ACC, higher is better) on different modalities (Non-IID Alpha=0.5).}
\label{tab:fl_comp_har}
\small
\centering
\setlength{\tabcolsep}{3.5pt}
\begin{tabular}{lccc}
\toprule
\textbf{Method} & \textbf{A} & \textbf{G} & \textbf{AG} \\
\midrule
Fed-Modality  & 85.2$\pm$0.8 & 79.1$\pm$1.3 & 91.3$\pm$0.3 \\
Fed-Task      & 83.2$\pm$1.2 & 76.5$\pm$1.4 & 85.7$\pm$0.8 \\
Fed-Hybrid    & 83.0$\pm$0.5 & 72.5$\pm$0.4 & 89.5$\pm$0.7 \\
Harmony       & 83.7$\pm$1.3 & 78.6$\pm$0.3 & 90.5$\pm$0.6 \\
FedHGB        & 82.2$\pm$1.2 & 75.4$\pm$0.5 & 88.3$\pm$0.9 \\
DMML-KD       & 85.2$\pm$0.6 & 77.6$\pm$1.1 & 92.1$\pm$0.4 \\
FedMVC        & 85.6$\pm$1.5 & 81.2$\pm$1.1 & 92.7$\pm$0.9 \\
FedMSplit     & 82.1$\pm$0.9 & 80.7$\pm$1.1 & 90.2$\pm$1.3 \\
FedMVD        & 85.1$\pm$1.4 & 78.8$\pm$1.0 & 92.5$\pm$0.8 \\
\PARSE        & \textbf{85.8}$\pm$0.9 & \textbf{81.5}$\pm$0.5 & \textbf{93.3}$\pm$1.2 \\
\bottomrule
\end{tabular}
\vspace{-0.1in}
\end{table}

\section{Conclusion} \label{sec:conclusion}

\PARSE introduces a PID-guided feature fission and partial-alignment framework for multimodal DFL, turning monolithic multimodal representations into slice-aware components that can be exchanged only over semantically compatible P2P overlays. By aligning shareable slices while keeping non-alignable components local or modality-set restricted, \PARSE avoids destructive gradient conflicts without centralized orchestration or gradient surgery. Across four benchmarks, diverse modality mixes, non-IID settings, and multiple overlay topologies, \PARSE consistently improves both unimodal and multimodal performance over representative multimodal DFL baselines while maintaining bounded per-link communication. These results show that representation structure and overlay design can be jointly optimized to support scalable, server-free multimodal learning among heterogeneous agents.

\bibliographystyle{IEEEtran}
\bibliography{ToN/references}

\appendices


\section{Modality-Mix Results on IEMOCAP}\label{app:exp_results}

As a complement to the main results in Section~\ref{sec:exp}, we further evaluate performance across a broader set of modality combinations on the IEMOCAP dataset. Specifically, we consider all possible subsets of the three modalities: [A], [V], [T], [AV], [AT], [VT], and [AVT], covering a total of seven distinct modality sets.

To simulate varying degrees of modality availability, we experiment with different agent ratio configurations, which correspond to the following settings:

\noindent $\bullet$ \textbf{(30\%:30\%:40\%)}. \textit{We select 30\% of agents to miss two modalities (i.e., unimodal agents), 30\% to miss one modality (i.e., bimodal agents), and 40\% to have full modality access.}

\noindent $\bullet$ \textbf{(43\%:43\%:14\%)}. \textit{We select 43\% of agents to miss two modalities, 43\% to miss one modality, and 14\% to have full modality access.}

\noindent $\bullet$ \textbf{(47\%:47\%:6\%)}. \textit{We select 47\% of agents to miss two modalities, 47\% to miss one modality, and 6\% to have full modality access.}

\begin{table}[!ht]
\centering
\small
\renewcommand{\arraystretch}{1.2}
\caption{Performance comparison on IEMOCAP across all possible agent modality combinations.}
\label{tab:iemocap-mix}
\scalebox{0.65}{
\begin{tabular}{@{}l|ccc|ccc|c@{}}
\toprule
\multicolumn{8}{c}{\textbf{Agent Ratio: \textbf{(30\%:30\%:40\%)}}} \\
\midrule
\textbf{Method} & [A] & [V] & [T] & [AV] & [AT] & [VT] & [AVT] \\
\midrule
DSGD-Modality &
\heatmaplevel{6}40.8$\pm$0.6 & \heatmaplevel{3}47.9$\pm$0.8 & \heatmaplevel{4}53.5$\pm$1.0 &
\heatmaplevel{3}56.6$\pm$1.1 & \heatmaplevel{4}58.8$\pm$1.4 & \heatmaplevel{1}56.7$\pm$0.7 &
\heatmaplevel{2}58.0$\pm$0.6 \\
DSGD-Task &
\heatmaplevel{3}40.7$\pm$2.5 & \heatmaplevel{2}47.3$\pm$1.9 & \heatmaplevel{1}48.0$\pm$0.8 &
\heatmaplevel{1}51.3$\pm$0.9 & \heatmaplevel{1}54.6$\pm$1.2 & \heatmaplevel{4}60.4$\pm$0.6 &
\heatmaplevel{4}63.5$\pm$0.9 \\
DSGD-Hybrid &
\heatmaplevel{2}38.0$\pm$1.4 & \heatmaplevel{1}47.1$\pm$1.5 & \heatmaplevel{3}52.7$\pm$1.0 &
\heatmaplevel{2}54.1$\pm$0.8 & \heatmaplevel{5}60.4$\pm$1.1 & \heatmaplevel{5}61.7$\pm$0.7 &
\heatmaplevel{5}64.0$\pm$0.8 \\
Harmony &
\heatmaplevel{1}37.6$\pm$2.2 & \heatmaplevel{2}47.5$\pm$1.3 & \heatmaplevel{2}51.3$\pm$1.0 &
\heatmaplevel{1}47.3$\pm$0.6 & \heatmaplevel{2}55.7$\pm$1.6 & \heatmaplevel{2}57.4$\pm$1.4 &
\heatmaplevel{1}54.6$\pm$1.0 \\
FedHGB &
\heatmaplevel{3}40.3$\pm$1.0 & \heatmaplevel{6}48.4$\pm$1.4 & \heatmaplevel{1}50.7$\pm$1.5 &
\heatmaplevel{1}51.6$\pm$1.7 & \heatmaplevel{3}58.4$\pm$1.8 & \heatmaplevel{3}59.5$\pm$1.1 &
\heatmaplevel{3}59.8$\pm$0.9 \\
DMML-KD &
\heatmaplevel{3}37.1$\pm$1.0 & \heatmaplevel{3}48.2$\pm$1.0 & \heatmaplevel{5}53.3$\pm$1.1 &
\heatmaplevel{5}60.6$\pm$0.8 & \heatmaplevel{6}61.9$\pm$1.1 & \heatmaplevel{6}63.1$\pm$1.4 &
\heatmaplevel{6}68.8$\pm$1.3 \\
\PARSE &
\heatmaplevel{7}40.9$\pm$1.4 & \heatmaplevel{7}48.8$\pm$0.6 & \heatmaplevel{7}54.7$\pm$0.9 &
\heatmaplevel{7}61.3$\pm$0.6 & \heatmaplevel{7}64.2$\pm$0.5 & \heatmaplevel{7}65.4$\pm$0.8 &
\heatmaplevel{7}70.4$\pm$0.7 \\
\midrule
\multicolumn{8}{c}{\textbf{Agent Ratio: \textbf{(43\%:43\%:14\%)}}} \\
\midrule
\textbf{Method} & [A] & [V] & [T] & [AV] & [AT] & [VT] & [AVT] \\
\midrule
DSGD-Modality &
\heatmaplevel{5}42.7$\pm$1.9 & \heatmaplevel{5}48.3$\pm$1.1 & \heatmaplevel{2}55.7$\pm$0.6 &
\heatmaplevel{2}52.2$\pm$1.2 & \heatmaplevel{2}59.7$\pm$1.3 & \heatmaplevel{1}55.8$\pm$0.7 &
\heatmaplevel{2}57.1$\pm$1.0 \\
DSGD-Task &
\heatmaplevel{1}41.5$\pm$1.7 & \heatmaplevel{2}47.2$\pm$1.7 & \heatmaplevel{1}53.9$\pm$0.5 &
\heatmaplevel{3}55.2$\pm$0.6 & \heatmaplevel{4}62.4$\pm$0.7 & \heatmaplevel{2}58.9$\pm$0.8 &
\heatmaplevel{3}61.0$\pm$0.5 \\
DSGD-Hybrid &
\heatmaplevel{3}42.1$\pm$1.3 & \heatmaplevel{4}47.8$\pm$0.6 & \heatmaplevel{5}56.2$\pm$0.9 &
\heatmaplevel{4}54.3$\pm$0.5 & \heatmaplevel{4}62.4$\pm$0.9 & \heatmaplevel{4}61.0$\pm$0.5 &
\heatmaplevel{4}62.3$\pm$0.7 \\
Harmony &
\heatmaplevel{1}41.4$\pm$1.2 & \heatmaplevel{1}46.2$\pm$0.8 & \heatmaplevel{2}55.7$\pm$0.3 &
\heatmaplevel{1}50.0$\pm$0.4 & \heatmaplevel{1}58.5$\pm$0.4 & \heatmaplevel{1}55.4$\pm$1.0 &
\heatmaplevel{1}54.1$\pm$1.8 \\
FedHGB &
\heatmaplevel{3}42.1$\pm$1.8 & \heatmaplevel{2}47.0$\pm$1.7 & \heatmaplevel{1}54.6$\pm$0.7 &
\heatmaplevel{2}52.7$\pm$0.8 & \heatmaplevel{3}60.5$\pm$1.3 & \heatmaplevel{2}57.1$\pm$1.3 &
\heatmaplevel{2}59.1$\pm$0.7 \\
DMML-KD &
\heatmaplevel{5}42.7$\pm$1.8 & \heatmaplevel{2}47.2$\pm$0.9 & \heatmaplevel{4}56.1$\pm$0.8 &
\heatmaplevel{5}58.5$\pm$1.3 & \heatmaplevel{5}64.7$\pm$0.5 & \heatmaplevel{5}62.3$\pm$0.9 &
\heatmaplevel{5}67.5$\pm$0.7 \\
\PARSE &
\heatmaplevel{7}43.2$\pm$1.0 & \heatmaplevel{7}48.8$\pm$0.9 & \heatmaplevel{7}56.4$\pm$0.7 &
\heatmaplevel{7}61.2$\pm$0.7 & \heatmaplevel{7}66.3$\pm$1.2 & \heatmaplevel{7}64.4$\pm$0.9 &
\heatmaplevel{7}69.0$\pm$0.7 \\
\midrule
\multicolumn{8}{c}{\textbf{Agent Ratio: \textbf{(47\%:47\%:6\%)}}} \\
\midrule
\textbf{Method} & [A] & [V] & [T] & [AV] & [AT] & [VT] & [AVT] \\
\midrule
DSGD-Modality &
\heatmaplevel{5}44.3$\pm$1.6 & \heatmaplevel{5}46.3$\pm$1.0 & \heatmaplevel{4}54.4$\pm$0.6 &
\heatmaplevel{2}49.3$\pm$0.7 & \heatmaplevel{2}58.3$\pm$0.9 & \heatmaplevel{1}58.3$\pm$0.7 &
\heatmaplevel{2}61.3$\pm$0.9 \\
DSGD-Task &
\heatmaplevel{4}44.1$\pm$1.4 & \heatmaplevel{3}45.5$\pm$0.6 & \heatmaplevel{1}53.5$\pm$0.5 &
\heatmaplevel{3}51.3$\pm$1.3 & \heatmaplevel{1}57.8$\pm$0.4 & \heatmaplevel{3}61.1$\pm$0.9 &
\heatmaplevel{3}62.9$\pm$0.8 \\
DSGD-Hybrid &
\heatmaplevel{6}44.7$\pm$0.7 & \heatmaplevel{2}45.3$\pm$1.1 & \heatmaplevel{3}54.3$\pm$1.0 &
\heatmaplevel{3}51.3$\pm$0.6 & \heatmaplevel{3}58.8$\pm$0.3 & \heatmaplevel{4}62.0$\pm$0.4 &
\heatmaplevel{4}66.4$\pm$0.6 \\
Harmony &
\heatmaplevel{1}42.6$\pm$1.1 & \heatmaplevel{1}43.3$\pm$0.6 & \heatmaplevel{2}54.1$\pm$0.9 &
\heatmaplevel{1}45.6$\pm$0.8 & \heatmaplevel{1}54.3$\pm$0.5 & \heatmaplevel{1}58.3$\pm$1.0 &
\heatmaplevel{1}58.2$\pm$1.1 \\
FedHGB &
\heatmaplevel{2}43.3$\pm$1.4 & \heatmaplevel{6}46.6$\pm$0.4 & \heatmaplevel{1}53.1$\pm$0.6 &
\heatmaplevel{2}49.6$\pm$0.5 & \heatmaplevel{4}59.1$\pm$1.0 & \heatmaplevel{2}60.4$\pm$0.7 &
\heatmaplevel{2}60.6$\pm$1.3 \\
DMML-KD &
\heatmaplevel{3}43.1$\pm$1.2 & \heatmaplevel{4}45.7$\pm$0.7 & \heatmaplevel{6}54.7$\pm$1.1 &
\heatmaplevel{5}57.2$\pm$0.7 & \heatmaplevel{5}63.3$\pm$0.7 & \heatmaplevel{5}63.4$\pm$1.6 &
\heatmaplevel{5}69.0$\pm$1.5 \\
\PARSE &
\heatmaplevel{7}45.8$\pm$1.1 & \heatmaplevel{7}47.6$\pm$1.0 & \heatmaplevel{7}55.3$\pm$0.6 &
\heatmaplevel{7}58.3$\pm$0.7 & \heatmaplevel{7}65.1$\pm$0.6 & \heatmaplevel{7}65.5$\pm$0.5 &
\heatmaplevel{7}71.7$\pm$0.5 \\
\bottomrule
\end{tabular}}
\end{table}

Under a default non-IID setting ($\alpha=0.5$) and a ring topology, the corresponding results are reported in Table~\ref{tab:iemocap-mix}. The key difference between this full combination setting and our earlier unimodal vs. all-modal comparison is that methods primarily optimized for multimodal performance, such as DSGD-Task and Harmony, struggle to capture effective cross-modal interactions, leading to degraded performance even for multimodal agents. In contrast, methods like DMML-KD and \PARSE, which explicitly specialize feature learning, achieve a distinct performance profile from other methods. Notably, \PARSE, our proposed approach, not only outperforms all baselines across the board, but the performance gap becomes especially pronounced for multimodal agents, highlighting its advantage in capturing synergistic and shareable representations.

\begin{table*}[!t]
\centering
\small
\renewcommand{\arraystretch}{1.2}
\caption{Performance (accuracy \%) of methods under varying agent-ratio scenarios (Dirichlet $\alpha=0.5$, Chordal Ring topology).}
\label{tab:0.5-chordal-agent}
\scalebox{0.75}{
\begin{subtable}[t]{\textwidth}
\resizebox{\textwidth}{!}{
\begin{tabular}{|c|l|ccc|ccc|ccc|}
\hline
\multicolumn{2}{|c|}{\multirow{1}{*}{Agent ratios}} & \multicolumn{3}{c|}{$6:6:18$} & \multicolumn{3}{c|}{$10:10:10$} & \multicolumn{3}{c|}{$13:13:4$} \\
\cline{1-11}
\multicolumn{2}{|c|}{\multirow{1}{*}{\textbf{Agent types}}} & [\textbf{A}] & [\textbf{G}] &[\textbf{AG}] &[\textbf{A}] & [\textbf{G}] &[\textbf{AG}]  &[\textbf{A}] & [\textbf{G}] &[\textbf{AG}]  \\
\hline
\multirow{7}{*}{\rotatebox{90}{\textbf{KU-HAR}}} 
& DSGD-Modality 
& \heatmaplevel{6} 82.2$\pm$1.0 & \heatmaplevel{6} 75.6$\pm$1.5 & \heatmaplevel{4} 85.4$\pm$1.2 
& \heatmaplevel{5} 82.0$\pm$0.5 & \heatmaplevel{6} 75.8$\pm$0.9 & \heatmaplevel{4} 88.2$\pm$0.8 
& \heatmaplevel{4} 81.2$\pm$1.3 & \heatmaplevel{5} 76.1$\pm$1.3 & \heatmaplevel{2} 88.0$\pm$1.2 \\
& DSGD-Task 
& \heatmaplevel{3} 77.8$\pm$0.9 & \heatmaplevel{3} 67.5$\pm$1.2 & \heatmaplevel{1} 84.5$\pm$0.9 
& \heatmaplevel{5} 82.0$\pm$1.8 & \heatmaplevel{3} 72.4$\pm$0.9 & \heatmaplevel{1} 84.0$\pm$1.5 
& \heatmaplevel{5} 81.5$\pm$0.7 & \heatmaplevel{1} 74.7$\pm$0.6 & \heatmaplevel{1} 80.2$\pm$2.0 \\
& DSGD-Hybrid 
& \heatmaplevel{1} 77.1$\pm$0.6 & \heatmaplevel{1} 62.4$\pm$0.8 & \heatmaplevel{2} 85.0$\pm$1.6 
& \heatmaplevel{1} 78.2$\pm$1.5 & \heatmaplevel{1} 70.2$\pm$0.7 & \heatmaplevel{2} 87.9$\pm$0.5 
& \heatmaplevel{1} 80.2$\pm$0.9 & \heatmaplevel{3} 75.5$\pm$0.8 & \heatmaplevel{4} 88.7$\pm$0.7 \\
& Harmony 
& \heatmaplevel{5} 81.9$\pm$1.0 & \heatmaplevel{5} 75.2$\pm$1.9 & \heatmaplevel{6} 91.0$\pm$0.4 
& \heatmaplevel{5} 81.0$\pm$1.7 & \heatmaplevel{5} 74.5$\pm$0.4 & \heatmaplevel{6} 90.4$\pm$0.9 
& \heatmaplevel{2} 80.6$\pm$1.5 & \heatmaplevel{4} 75.9$\pm$0.8 & \heatmaplevel{5} 89.7$\pm$1.0 \\
& FedHGB 
& \heatmaplevel{2} 77.5$\pm$0.7 & \heatmaplevel{4} 70.2$\pm$0.8 & \heatmaplevel{3} 85.1$\pm$0.6 
& \heatmaplevel{2} 79.3$\pm$1.0 & \heatmaplevel{4} 72.8$\pm$0.5 & \heatmaplevel{3} 88.0$\pm$0.9 
& \heatmaplevel{3} 80.7$\pm$0.5 & \heatmaplevel{6} 77.1$\pm$0.4 & \heatmaplevel{3} 88.3$\pm$1.0 \\
& DMML-KD 
& \heatmaplevel{4} 80.4$\pm$1.5 & \heatmaplevel{2} 67.1$\pm$2.0 & \heatmaplevel{5} 88.2$\pm$0.6 
& \heatmaplevel{3} 81.5$\pm$1.1 & \heatmaplevel{1} 70.2$\pm$1.7 & \heatmaplevel{5} 89.3$\pm$0.7 
& \heatmaplevel{6} 82.1$\pm$1.4 & \heatmaplevel{2} 75.4$\pm$1.5 & \heatmaplevel{6} 90.0$\pm$1.7 \\
& \PARSE 
& \heatmaplevel{7} 82.4$\pm$1.2 & \heatmaplevel{7} 76.1$\pm$1.3 & \heatmaplevel{7} 91.8$\pm$0.5 
& \heatmaplevel{7} 82.6$\pm$1.1 & \heatmaplevel{7} 76.5$\pm$1.8 & \heatmaplevel{7} 90.8$\pm$0.9 
& \heatmaplevel{7} 83.4$\pm$0.7 & \heatmaplevel{7} 78.4$\pm$0.8 & \heatmaplevel{7} 90.3$\pm$1.2 \\
\cline{1-11}
\multicolumn{2}{|c|}{\multirow{1}{*}{\textbf{Agent types}}} & [\textbf{V1}] & [\textbf{V2}] &[\textbf{V1,V2}] &[\textbf{V1}] & [\textbf{V2}] &[\textbf{V1,V2}] & [\textbf{V1}] & [\textbf{V2}] &[\textbf{V1,V2}] \\
\hline
\multirow{7}{*}{\rotatebox{90}{\textbf{ModelNet-40}}} 
& DSGD-Modality 
& \heatmaplevel{6} 80.9$\pm$1.0 & \heatmaplevel{6} 72.8$\pm$0.8 & \heatmaplevel{6} 81.6$\pm$1.1 
& \heatmaplevel{3} 78.1$\pm$0.5 & \heatmaplevel{5} 71.9$\pm$1.3 & \heatmaplevel{4} 80.8$\pm$1.0 
& \heatmaplevel{6} 80.3$\pm$1.0 & \heatmaplevel{5} 72.9$\pm$2.1 & \heatmaplevel{6} 82.6$\pm$1.3 \\
& DSGD-Task 
& \heatmaplevel{1} 71.1$\pm$0.7 & \heatmaplevel{4} 69.1$\pm$1.1 & \heatmaplevel{2} 75.4$\pm$0.7 
& \heatmaplevel{2} 75.6$\pm$2.1 & \heatmaplevel{3} 71.2$\pm$0.7 & \heatmaplevel{1} 71.7$\pm$0.6 
& \heatmaplevel{1} 76.4$\pm$0.9 & \heatmaplevel{4} 71.6$\pm$1.3 & \heatmaplevel{1} 66.1$\pm$1.6 \\
& DSGD-Hybrid 
& \heatmaplevel{3} 72.2$\pm$1.5 & \heatmaplevel{2} 63.3$\pm$1.6 & \heatmaplevel{1} 74.7$\pm$1.3 
& \heatmaplevel{1} 74.8$\pm$0.8 & \heatmaplevel{2} 67.3$\pm$1.1 & \heatmaplevel{2} 77.3$\pm$0.7 
& \heatmaplevel{4} 78.8$\pm$2.1 & \heatmaplevel{2} 71.2$\pm$1.7 & \heatmaplevel{4} 80.1$\pm$1.2 \\
& Harmony 
& \heatmaplevel{4} 78.8$\pm$1.3 & \heatmaplevel{5} 71.7$\pm$1.4 & \heatmaplevel{5} 81.2$\pm$0.4 
& \heatmaplevel{5} 80.1$\pm$1.1 & \heatmaplevel{6} 75.2$\pm$1.3 & \heatmaplevel{5} 81.0$\pm$0.8 
& \heatmaplevel{2} 77.1$\pm$0.6 & \heatmaplevel{3} 71.5$\pm$1.9 & \heatmaplevel{2} 79.0$\pm$1.7 \\
& FedHGB 
& \heatmaplevel{2} 72.0$\pm$0.7 & \heatmaplevel{3} 65.5$\pm$1.2 & \heatmaplevel{3} 77.2$\pm$2.1 
& \heatmaplevel{4} 78.6$\pm$0.9 & \heatmaplevel{4} 71.6$\pm$1.1 & \heatmaplevel{3} 77.6$\pm$0.5 
& \heatmaplevel{3} 77.3$\pm$0.8 & \heatmaplevel{6} 73.1$\pm$1.8 & \heatmaplevel{3} 79.3$\pm$1.1 \\
& DMML-KD 
& \heatmaplevel{5} 79.6$\pm$1.2 & \heatmaplevel{1} 59.3$\pm$1.1 & \heatmaplevel{4} 79.7$\pm$0.9 
& \heatmaplevel{6} 80.4$\pm$1.1 & \heatmaplevel{1} 57.8$\pm$1.5 & \heatmaplevel{6} 81.3$\pm$1.1 
& \heatmaplevel{5} 78.9$\pm$1.5 & \heatmaplevel{1} 69.1$\pm$1.7 & \heatmaplevel{5} 82.2$\pm$0.9 \\
& \PARSE 
& \heatmaplevel{7} 81.6$\pm$1.2 & \heatmaplevel{7} 76.4$\pm$0.9 & \heatmaplevel{7} 82.1$\pm$0.6 
& \heatmaplevel{7} 80.9$\pm$0.5 & \heatmaplevel{7} 76.0$\pm$0.8 & \heatmaplevel{7} 81.6$\pm$0.7 
& \heatmaplevel{7} 81.4$\pm$1.3 & \heatmaplevel{7} 75.2$\pm$1.5 & \heatmaplevel{7} 83.0$\pm$0.7 \\
\cline{1-11}
\multicolumn{2}{|c|}{\multirow{1}{*}{\textbf{Agent types}}} & [\textbf{A}] & [\textbf{V}] &[\textbf{AV}] &[\textbf{A}] & [\textbf{V}] &[\textbf{AV}]  &[\textbf{A}] & [\textbf{V}] &[\textbf{AV}]  \\
\hline
\multirow{7}{*}{\rotatebox{90}{\textbf{AVE}}} 
& DSGD-Modality 
& \heatmaplevel{6} 47.2$\pm$1.2 & \heatmaplevel{6} 53.8$\pm$1.7 & \heatmaplevel{6} 67.5$\pm$1.1 
& \heatmaplevel{6} 46.6$\pm$1.4 & \heatmaplevel{5} 53.3$\pm$1.4 & \heatmaplevel{2} 65.3$\pm$1.0 
& \heatmaplevel{3} 45.1$\pm$1.6 & \heatmaplevel{3} 52.2$\pm$1.2 & \heatmaplevel{3} 64.9$\pm$1.5 \\
& DSGD-Task 
& \heatmaplevel{1} 38.5$\pm$0.9 & \heatmaplevel{1} 47.0$\pm$0.8 & \heatmaplevel{1} 61.5$\pm$0.7 
& \heatmaplevel{1} 40.7$\pm$1.0 & \heatmaplevel{1} 50.8$\pm$0.5 & \heatmaplevel{1} 57.7$\pm$1.5 
& \heatmaplevel{1} 42.7$\pm$1.8 & \heatmaplevel{2} 51.5$\pm$1.8 & \heatmaplevel{1} 51.3$\pm$0.6 \\
& DSGD-Hybrid 
& \heatmaplevel{2} 39.1$\pm$0.6 & \heatmaplevel{2} 50.2$\pm$0.9 & \heatmaplevel{2} 64.7$\pm$0.8 
& \heatmaplevel{2} 42.6$\pm$0.5 & \heatmaplevel{3} 51.6$\pm$0.8 & \heatmaplevel{3} 65.4$\pm$0.7 
& \heatmaplevel{2} 44.3$\pm$1.6 & \heatmaplevel{4} 52.6$\pm$1.3 & \heatmaplevel{4} 65.6$\pm$1.6 \\
& Harmony 
& \heatmaplevel{3} 42.3$\pm$0.8 & \heatmaplevel{3} 51.2$\pm$1.4 & \heatmaplevel{3} 65.6$\pm$1.2 
& \heatmaplevel{4} 45.0$\pm$0.7 & \heatmaplevel{2} 51.3$\pm$1.6 & \heatmaplevel{5} 66.6$\pm$1.3 
& \heatmaplevel{4} 45.2$\pm$0.5 & \heatmaplevel{1} 51.0$\pm$1.5 & \heatmaplevel{2} 64.3$\pm$1.0 \\
& FedHGB 
& \heatmaplevel{5} 46.2$\pm$0.6 & \heatmaplevel{5} 53.6$\pm$0.6 & \heatmaplevel{5} 67.1$\pm$0.5 
& \heatmaplevel{5} 46.5$\pm$0.8 & \heatmaplevel{6} 53.7$\pm$1.1 & \heatmaplevel{4} 66.1$\pm$0.7 
& \heatmaplevel{6} 46.8$\pm$0.9 & \heatmaplevel{6} 53.4$\pm$1.1 & \heatmaplevel{6} 66.4$\pm$1.3 \\
& DMML-KD 
& \heatmaplevel{4} 43.3$\pm$1.7 & \heatmaplevel{4} 52.7$\pm$1.2 & \heatmaplevel{4} 67.0$\pm$0.6 
& \heatmaplevel{3} 44.3$\pm$1.5 & \heatmaplevel{4} 53.2$\pm$1.3 & \heatmaplevel{6} 66.7$\pm$0.6 
& \heatmaplevel{5} 45.8$\pm$1.2 & \heatmaplevel{5} 52.8$\pm$1.1 & \heatmaplevel{5} 66.2$\pm$1.3 \\
& \PARSE 
& \heatmaplevel{7} 47.6$\pm$1.1 & \heatmaplevel{7} 55.4$\pm$1.3 & \heatmaplevel{7} 68.1$\pm$0.5 
& \heatmaplevel{7} 47.0$\pm$0.5 & \heatmaplevel{7} 54.3$\pm$0.7 & \heatmaplevel{7} 67.2$\pm$0.8 
& \heatmaplevel{7} 47.1$\pm$0.3 & \heatmaplevel{7} 53.7$\pm$1.0 & \heatmaplevel{7} 67.3$\pm$0.6 \\
\hline
\end{tabular}}
\end{subtable}}
\scalebox{0.75}{
\begin{subtable}[t]{\textwidth}
\renewcommand{\arraystretch}{1.2}
\resizebox{\textwidth}{!}{
\begin{tabular}{|c|l|cccc|cccc|cccc|}
\hline
\multicolumn{2}{|c|}{\multirow{1}{*}{Agent ratios}} & \multicolumn{4}{c|}{$6:6:6:22$} & \multicolumn{4}{c|}{$10:10:10:10$} & \multicolumn{4}{c|}{$12:12:12:4$} \\
\cline{1-14}
\multicolumn{2}{|c|}{\multirow{1}{*}{\textbf{Agent types}}} & [\textbf{A}] & [\textbf{V}] &[\textbf{T}] &[\textbf{AVT}] &[\textbf{A}] & [\textbf{V}] &[\textbf{T}] &[\textbf{AVT}]  &[\textbf{A}] & [\textbf{V}] &[\textbf{T}] &[\textbf{AVT}]  \\
\hline
\multirow{7}{*}{\rotatebox{90}{\textbf{IEMOCAP}}} 
& DSGD-Modality 
& \heatmaplevel{5} 47.5$\pm$0.6 & \heatmaplevel{6} 48.8$\pm$1.6 & \heatmaplevel{4} 62.6$\pm$0.8 & \heatmaplevel{2} 70.4$\pm$0.2
& \heatmaplevel{2} 46.3$\pm$0.5 & \heatmaplevel{6} 52.2$\pm$0.8 & \heatmaplevel{4} 60.2$\pm$0.7 & \heatmaplevel{2} 70.3$\pm$0.2
& \heatmaplevel{3} 51.1$\pm$0.3 & \heatmaplevel{6} 53.1$\pm$0.4 & \heatmaplevel{3} 58.0$\pm$0.4 & \heatmaplevel{5} 71.8$\pm$0.6 \\
& DSGD-Task 
& \heatmaplevel{1} 48.5$\pm$0.8 & \heatmaplevel{1} 45.1$\pm$0.6 & \heatmaplevel{1} 58.5$\pm$0.4 & \heatmaplevel{4} 72.3$\pm$0.6
& \heatmaplevel{1} 44.8$\pm$0.5 & \heatmaplevel{4} 51.9$\pm$2.1 & \heatmaplevel{2} 58.7$\pm$0.7 & \heatmaplevel{1} 70.1$\pm$0.5
& \heatmaplevel{3} 51.1$\pm$0.3 & \heatmaplevel{3} 52.7$\pm$0.4 & \heatmaplevel{6} 58.5$\pm$0.8 & \heatmaplevel{1} 64.4$\pm$1.1 \\
& DSGD-Hybrid 
& \heatmaplevel{2} 47.3$\pm$0.6 & \heatmaplevel{3} 45.6$\pm$0.4 & \heatmaplevel{2} 62.3$\pm$0.6 & \heatmaplevel{3} 71.9$\pm$0.8
& \heatmaplevel{5} 48.0$\pm$1.3 & \heatmaplevel{3} 51.0$\pm$0.5 & \heatmaplevel{6} 60.6$\pm$0.8 & \heatmaplevel{4} 71.2$\pm$0.6
& \heatmaplevel{2} 51.0$\pm$0.4 & \heatmaplevel{4} 53.0$\pm$1.3 & \heatmaplevel{4} 58.1$\pm$0.6 & \heatmaplevel{3} 71.2$\pm$0.4 \\
& Harmony 
& \heatmaplevel{3} 47.4$\pm$1.2 & \heatmaplevel{4} 47.3$\pm$0.8 & \heatmaplevel{6} 63.3$\pm$0.2 & \heatmaplevel{5} 72.4$\pm$1.0
& \heatmaplevel{6} 48.3$\pm$1.5 & \heatmaplevel{4} 51.9$\pm$0.6 & \heatmaplevel{3} 60.1$\pm$0.5 & \heatmaplevel{5} 72.4$\pm$0.3
& \heatmaplevel{1} 50.9$\pm$0.6 & \heatmaplevel{4} 53.0$\pm$1.8 & \heatmaplevel{5} 58.3$\pm$1.8 & \heatmaplevel{2} 70.9$\pm$0.4 \\
& FedHGB 
& \heatmaplevel{3} 47.4$\pm$0.9 & \heatmaplevel{5} 48.5$\pm$1.6 & \heatmaplevel{3} 62.5$\pm$0.3 & \heatmaplevel{1} 70.3$\pm$0.5
& \heatmaplevel{2} 46.4$\pm$0.7 & \heatmaplevel{2} 50.2$\pm$1.0 & \heatmaplevel{4} 60.2$\pm$0.5 & \heatmaplevel{3} 70.7$\pm$0.5
& \heatmaplevel{6} 51.4$\pm$0.4 & \heatmaplevel{2} 52.0$\pm$1.7 & \heatmaplevel{1} 57.8$\pm$0.6 & \heatmaplevel{3} 71.2$\pm$0.8 \\
& DMML-KD 
& \heatmaplevel{7} 48.0$\pm$0.3 & \heatmaplevel{1} 45.1$\pm$0.6 & \heatmaplevel{5} 63.1$\pm$0.3 & \heatmaplevel{6} 74.3$\pm$0.2
& \heatmaplevel{4} 47.7$\pm$1.3 & \heatmaplevel{1} 49.2$\pm$0.5 & \heatmaplevel{1} 58.3$\pm$0.7 & \heatmaplevel{7} 74.7$\pm$0.5
& \heatmaplevel{5} 51.4$\pm$1.5 & \heatmaplevel{1} 50.5$\pm$0.3 & \heatmaplevel{2} 57.9$\pm$0.6 & \heatmaplevel{6} 74.1$\pm$0.6 \\
& \PARSE 
& \heatmaplevel{7} 48.0$\pm$0.3 & \heatmaplevel{7} 52.0$\pm$0.4 & \heatmaplevel{7} 64.3$\pm$0.5 & \heatmaplevel{7} 75.8$\pm$0.5
& \heatmaplevel{7} 48.5$\pm$0.2 & \heatmaplevel{7} 52.6$\pm$0.4 & \heatmaplevel{7} 60.7$\pm$0.8 & \heatmaplevel{7} 74.7$\pm$0.2
& \heatmaplevel{7} 51.7$\pm$0.5 & \heatmaplevel{7} 53.2$\pm$0.5 & \heatmaplevel{7} 58.7$\pm$0.9 & \heatmaplevel{7} 74.3$\pm$0.4 \\
\hline
\end{tabular}}
\end{subtable}}
\end{table*}
\begin{table*}[!ht]
\centering
\small
\renewcommand{\arraystretch}{1.2}
\caption{Performance (accuracy \%) of methods under varying agent-ratio scenarios (Dirichlet $\alpha=0.5$, Random Gossip Topology).}
\label{tab:0.5-random-agent}
\scalebox{0.75}{
\begin{subtable}[t]{\textwidth}
\resizebox{\textwidth}{!}{
\begin{tabular}{|c|l|ccc|ccc|ccc|}
\hline
\multicolumn{2}{|c|}{\multirow{1}{*}{Agent ratios}} & \multicolumn{3}{c|}{$6:6:18$} & \multicolumn{3}{c|}{$10:10:10$} & \multicolumn{3}{c|}{$13:13:4$} \\
\cline{1-11}
\multicolumn{2}{|c|}{\multirow{1}{*}{\textbf{Agent types}}} & [\textbf{A}] & [\textbf{G}] &[\textbf{AG}] &[\textbf{A}] & [\textbf{G}] &[\textbf{AG}]  &[\textbf{A}] & [\textbf{G}] &[\textbf{AG}]  \\
\hline
\multirow{7}{*}{\rotatebox{90}{\textbf{KU-HAR}}} 
& DSGD-Modality 
& \heatmaplevel{6} 81.2$\pm$1.7 & \heatmaplevel{6} 74.3$\pm$1.1 & \heatmaplevel{4} 86.1$\pm$0.6 
& \heatmaplevel{7} 80.5$\pm$1.5 & \heatmaplevel{6} 72.6$\pm$1.8 & \heatmaplevel{2} 85.8$\pm$1.3 
& \heatmaplevel{5} 79.2$\pm$1.1 & \heatmaplevel{3} 72.4$\pm$2.1 & \heatmaplevel{2} 84.9$\pm$1.2 \\
& DSGD-Task 
& \heatmaplevel{1} 74.6$\pm$0.8 & \heatmaplevel{2} 63.9$\pm$1.7 & \heatmaplevel{1} 83.9$\pm$0.6 
& \heatmaplevel{1} 77.8$\pm$1.9 & \heatmaplevel{2} 68.4$\pm$1.6 & \heatmaplevel{1} 83.4$\pm$0.8 
& \heatmaplevel{1} 77.2$\pm$1.0 & \heatmaplevel{1} 69.4$\pm$2.5 & \heatmaplevel{1} 77.4$\pm$0.7 \\
& DSGD-Hybrid 
& \heatmaplevel{2} 76.2$\pm$0.8 & \heatmaplevel{3} 64.5$\pm$1.0 & \heatmaplevel{3} 85.7$\pm$0.4 
& \heatmaplevel{2} 78.5$\pm$1.5 & \heatmaplevel{3} 70.8$\pm$1.3 & \heatmaplevel{4} 86.3$\pm$1.4 
& \heatmaplevel{3} 78.9$\pm$0.5 & \heatmaplevel{2} 71.9$\pm$0.7 & \heatmaplevel{6} 87.1$\pm$1.7 \\
& Harmony 
& \heatmaplevel{5} 81.0$\pm$0.9 & \heatmaplevel{5} 73.2$\pm$1.1 & \heatmaplevel{6} 87.5$\pm$1.7 
& \heatmaplevel{5} 80.4$\pm$1.3 & \heatmaplevel{5} 72.5$\pm$1.4 & \heatmaplevel{5} 86.4$\pm$1.8 
& \heatmaplevel{4} 79.0$\pm$1.5 & \heatmaplevel{4} 72.8$\pm$2.0 & \heatmaplevel{3} 85.2$\pm$1.2 \\
& FedHGB 
& \heatmaplevel{3} 78.3$\pm$1.7 & \heatmaplevel{4} 69.2$\pm$1.0 & \heatmaplevel{2} 85.2$\pm$1.3 
& \heatmaplevel{3} 79.6$\pm$1.5 & \heatmaplevel{4} 71.8$\pm$1.0 & \heatmaplevel{3} 86.1$\pm$0.7 
& \heatmaplevel{2} 77.7$\pm$1.2 & \heatmaplevel{6} 73.4$\pm$1.3 & \heatmaplevel{5} 86.7$\pm$0.8 \\
& DMML-KD 
& \heatmaplevel{4} 80.6$\pm$1.6 & \heatmaplevel{1} 62.6$\pm$1.5 & \heatmaplevel{5} 87.3$\pm$1.3 
& \heatmaplevel{4} 80.2$\pm$1.9 & \heatmaplevel{1} 68.2$\pm$1.6 & \heatmaplevel{6} 87.3$\pm$1.7 
& \heatmaplevel{6} 80.4$\pm$1.7 & \heatmaplevel{5} 73.2$\pm$0.8 & \heatmaplevel{4} 86.1$\pm$2.2 \\
& \PARSE 
& \heatmaplevel{7} 82.2$\pm$0.8 & \heatmaplevel{7} 77.2$\pm$1.6 & \heatmaplevel{7} 88.0$\pm$1.1 
& \heatmaplevel{7} 80.5$\pm$1.9 & \heatmaplevel{7} 76.4$\pm$1.5 & \heatmaplevel{7} 89.7$\pm$0.7 
& \heatmaplevel{7} 80.7$\pm$0.6 & \heatmaplevel{7} 75.7$\pm$1.7 & \heatmaplevel{7} 87.4$\pm$1.2 \\
\cline{1-11}
\multicolumn{2}{|c|}{\multirow{1}{*}{\textbf{Agent types}}} & [\textbf{V1}] & [\textbf{V2}] &[\textbf{V1,V2}] &[\textbf{V1}] & [\textbf{V2}] &[\textbf{V1,V2}] & [\textbf{V1}] & [\textbf{V2}] &[\textbf{V1,V2}] \\
\hline
\multirow{7}{*}{\rotatebox{90}{\textbf{ModelNet-40}}} 
& DSGD-Modality 
& \heatmaplevel{6} 82.4$\pm$1.7 & \heatmaplevel{6} 78.4$\pm$1.2 & \heatmaplevel{5} 83.0$\pm$1.4 
& \heatmaplevel{5} 80.6$\pm$1.6 & \heatmaplevel{6} 76.4$\pm$0.5 & \heatmaplevel{6} 84.1$\pm$0.6 
& \heatmaplevel{4} 80.4$\pm$1.2 & \heatmaplevel{6} 77.7$\pm$0.8 & \heatmaplevel{5} 82.1$\pm$0.7 \\
& DSGD-Task 
& \heatmaplevel{1} 74.4$\pm$1.8 & \heatmaplevel{2} 69.3$\pm$0.9 & \heatmaplevel{1} 78.1$\pm$1.8 
& \heatmaplevel{1} 78.8$\pm$1.6 & \heatmaplevel{2} 72.0$\pm$1.5 & \heatmaplevel{1} 77.6$\pm$1.5 
& \heatmaplevel{2} 79.2$\pm$0.9 & \heatmaplevel{4} 74.9$\pm$1.9 & \heatmaplevel{1} 67.5$\pm$1.3 \\
& DSGD-Hybrid 
& \heatmaplevel{2} 79.0$\pm$0.9 & \heatmaplevel{3} 70.3$\pm$2.5 & \heatmaplevel{2} 79.7$\pm$1.6 
& \heatmaplevel{3} 79.7$\pm$1.9 & \heatmaplevel{4} 72.7$\pm$1.7 & \heatmaplevel{2} 78.3$\pm$1.3 
& \heatmaplevel{3} 80.2$\pm$0.8 & \heatmaplevel{3} 74.6$\pm$1.0 & \heatmaplevel{3} 79.4$\pm$1.2 \\
& Harmony 
& \heatmaplevel{5} 81.2$\pm$0.6 & \heatmaplevel{5} 77.1$\pm$1.4 & \heatmaplevel{6} 83.8$\pm$0.5 
& \heatmaplevel{2} 79.0$\pm$1.7 & \heatmaplevel{5} 74.9$\pm$1.2 & \heatmaplevel{5} 83.7$\pm$0.7 
& \heatmaplevel{4} 80.4$\pm$1.0 & \heatmaplevel{5} 77.3$\pm$1.1 & \heatmaplevel{4} 80.6$\pm$1.0 \\
& FedHGB 
& \heatmaplevel{3} 80.1$\pm$0.9 & \heatmaplevel{4} 72.0$\pm$1.3 & \heatmaplevel{3} 80.5$\pm$0.6 
& \heatmaplevel{3} 79.7$\pm$1.6 & \heatmaplevel{3} 72.4$\pm$0.6 & \heatmaplevel{3} 78.8$\pm$0.9 
& \heatmaplevel{1} 78.7$\pm$1.7 & \heatmaplevel{2} 73.5$\pm$2.2 & \heatmaplevel{2} 78.6$\pm$1.3 \\
& DMML-KD 
& \heatmaplevel{4} 80.8$\pm$1.7 & \heatmaplevel{1} 67.7$\pm$1.9 & \heatmaplevel{4} 81.2$\pm$1.7 
& \heatmaplevel{6} 80.7$\pm$1.3 & \heatmaplevel{1} 68.9$\pm$1.4 & \heatmaplevel{4} 81.8$\pm$1.6 
& \heatmaplevel{6} 81.2$\pm$0.8 & \heatmaplevel{1} 71.5$\pm$1.3 & \heatmaplevel{6} 83.4$\pm$1.0 \\
& \PARSE 
& \heatmaplevel{7} 84.0$\pm$0.8 & \heatmaplevel{7} 81.4$\pm$0.7 & \heatmaplevel{7} 84.4$\pm$0.5 
& \heatmaplevel{7} 82.8$\pm$1.4 & \heatmaplevel{7} 78.2$\pm$1.1 & \heatmaplevel{7} 84.9$\pm$1.8 
& \heatmaplevel{7} 82.4$\pm$0.8 & \heatmaplevel{7} 78.3$\pm$1.4 & \heatmaplevel{7} 84.2$\pm$0.7 \\
\cline{1-11}
\multicolumn{2}{|c|}{\multirow{1}{*}{\textbf{Agent types}}} & [\textbf{A}] & [\textbf{V}] &[\textbf{AV}] &[\textbf{A}] & [\textbf{V}] &[\textbf{AV}]  &[\textbf{A}] & [\textbf{V}] &[\textbf{AV}]  \\
\hline
\multirow{7}{*}{\rotatebox{90}{\textbf{AVE}}} 
& DSGD-Modality 
& \heatmaplevel{6} 51.6$\pm$1.2 & \heatmaplevel{6} 55.8$\pm$0.4 & \heatmaplevel{3} 67.8$\pm$1.1 
& \heatmaplevel{6} 51.2$\pm$0.6 & \heatmaplevel{5} 54.8$\pm$1.6 & \heatmaplevel{3} 67.1$\pm$0.8 
& \heatmaplevel{6} 48.6$\pm$1.4 & \heatmaplevel{4} 53.1$\pm$1.1 & \heatmaplevel{5} 66.9$\pm$0.9 \\
& DSGD-Task 
& \heatmaplevel{1} 37.3$\pm$0.6 & \heatmaplevel{1} 45.3$\pm$0.7 & \heatmaplevel{1} 62.4$\pm$1.2 
& \heatmaplevel{2} 43.6$\pm$1.7 & \heatmaplevel{2} 51.1$\pm$1.3 & \heatmaplevel{1} 57.5$\pm$2.0 
& \heatmaplevel{2} 46.7$\pm$1.2 & \heatmaplevel{1} 51.4$\pm$0.9 & \heatmaplevel{1} 52.7$\pm$1.6 \\
& DSGD-Hybrid 
& \heatmaplevel{2} 41.3$\pm$0.7 & \heatmaplevel{2} 50.3$\pm$0.6 & \heatmaplevel{2} 66.2$\pm$1.2 
& \heatmaplevel{1} 42.4$\pm$1.0 & \heatmaplevel{3} 51.5$\pm$2.1 & \heatmaplevel{2} 65.6$\pm$0.8 
& \heatmaplevel{3} 47.3$\pm$1.8 & \heatmaplevel{2} 51.9$\pm$0.8 & \heatmaplevel{2} 63.4$\pm$0.5 \\
& Harmony 
& \heatmaplevel{4} 49.3$\pm$1.2 & \heatmaplevel{3} 51.9$\pm$1.8 & \heatmaplevel{6} 70.9$\pm$0.5 
& \heatmaplevel{3} 45.1$\pm$1.0 & \heatmaplevel{6} 55.3$\pm$0.7 & \heatmaplevel{5} 67.9$\pm$1.4 
& \heatmaplevel{5} 48.2$\pm$1.1 & \heatmaplevel{3} 52.3$\pm$0.7 & \heatmaplevel{3} 66.2$\pm$1.6 \\
& FedHGB 
& \heatmaplevel{5} 51.3$\pm$0.9 & \heatmaplevel{5} 55.0$\pm$1.3 & \heatmaplevel{5} 69.3$\pm$0.6 
& \heatmaplevel{5} 50.0$\pm$0.9 & \heatmaplevel{4} 54.1$\pm$1.4 & \heatmaplevel{5} 67.9$\pm$1.6 
& \heatmaplevel{4} 48.0$\pm$1.9 & \heatmaplevel{6} 54.3$\pm$0.6 & \heatmaplevel{3} 66.2$\pm$0.8 \\
& DMML-KD 
& \heatmaplevel{3} 45.8$\pm$0.7 & \heatmaplevel{4} 54.4$\pm$1.2 & \heatmaplevel{4} 68.4$\pm$0.8 
& \heatmaplevel{4} 45.4$\pm$1.3 & \heatmaplevel{1} 44.8$\pm$1.5 & \heatmaplevel{4} 67.5$\pm$0.6 
& \heatmaplevel{1} 45.9$\pm$2.2 & \heatmaplevel{5} 53.6$\pm$0.9 & \heatmaplevel{5} 66.9$\pm$1.5 \\
& \PARSE 
& \heatmaplevel{7} 51.7$\pm$1.1 & \heatmaplevel{7} 57.3$\pm$0.8 & \heatmaplevel{7} 71.2$\pm$0.4 
& \heatmaplevel{7} 51.9$\pm$1.5 & \heatmaplevel{7} 57.1$\pm$1.4 & \heatmaplevel{7} 68.4$\pm$0.8 
& \heatmaplevel{7} 49.3$\pm$1.0 & \heatmaplevel{7} 54.6$\pm$1.3 & \heatmaplevel{7} 68.3$\pm$1.8 \\
\hline
\end{tabular}}
\end{subtable}}
\scalebox{0.75}{
\begin{subtable}[t]{\textwidth}
\renewcommand{\arraystretch}{1.2}
\resizebox{\textwidth}{!}{
\begin{tabular}{|c|l|cccc|cccc|cccc|}
\hline
\multicolumn{2}{|c|}{\multirow{1}{*}{Agent ratios}} & \multicolumn{4}{c|}{$6:6:6:22$} & \multicolumn{4}{c|}{$10:10:10:10$} & \multicolumn{4}{c|}{$12:12:12:4$} \\
\cline{1-14}
\multicolumn{2}{|c|}{\multirow{1}{*}{\textbf{Agent types}}} & [\textbf{A}] & [\textbf{V}] &[\textbf{T}] &[\textbf{AVT}] &[\textbf{A}] & [\textbf{V}] &[\textbf{T}] &[\textbf{AVT}]  &[\textbf{A}] & [\textbf{V}] &[\textbf{T}] &[\textbf{AVT}]  \\
\hline
\multirow{7}{*}{\rotatebox{90}{IEMOCAP}} 
& DSGD-Modality 
& \heatmaplevel{3} 50.4$\pm$0.9 & \heatmaplevel{5} 50.2$\pm$1.9 & \heatmaplevel{4} 62.1$\pm$0.6 & \heatmaplevel{1} 70.3$\pm$1.2 
& \heatmaplevel{2} 48.9$\pm$1.2 & \heatmaplevel{5} 50.8$\pm$1.4 & \heatmaplevel{6} 61.9$\pm$0.5 & \heatmaplevel{4} 71.2$\pm$0.9 
& \heatmaplevel{1} 50.1$\pm$0.5 & \heatmaplevel{6} 52.0$\pm$1.3 & \heatmaplevel{3} 59.7$\pm$1.0 & \heatmaplevel{5} 71.4$\pm$0.5 \\
& DSGD-Task 
& \heatmaplevel{3} 50.4$\pm$1.0 & \heatmaplevel{4} 48.7$\pm$1.5 & \heatmaplevel{1} 57.7$\pm$0.7 & \heatmaplevel{2} 71.3$\pm$0.6 
& \heatmaplevel{4} 51.1$\pm$0.7 & \heatmaplevel{1} 47.0$\pm$0.9 & \heatmaplevel{1} 58.3$\pm$0.8 & \heatmaplevel{1} 68.1$\pm$0.9 
& \heatmaplevel{2} 51.3$\pm$0.6 & \heatmaplevel{2} 51.3$\pm$1.4 & \heatmaplevel{5} 60.4$\pm$1.3 & \heatmaplevel{1} 62.5$\pm$1.2 \\
& DSGD-Hybrid 
& \heatmaplevel{1} 44.6$\pm$0.5 & \heatmaplevel{2} 46.3$\pm$0.8 & \heatmaplevel{3} 61.5$\pm$1.0 & \heatmaplevel{4} 71.7$\pm$1.3 
& \heatmaplevel{1} 46.3$\pm$0.6 & \heatmaplevel{2} 47.3$\pm$0.5 & \heatmaplevel{4} 61.1$\pm$0.9 & \heatmaplevel{2} 70.3$\pm$0.5 
& \heatmaplevel{3} 52.1$\pm$1.4 & \heatmaplevel{4} 51.6$\pm$2.1 & \heatmaplevel{1} 59.5$\pm$0.9 & \heatmaplevel{2} 69.3$\pm$1.2 \\
& Harmony 
& \heatmaplevel{6} 50.8$\pm$1.7 & \heatmaplevel{3} 47.6$\pm$1.5 & \heatmaplevel{2} 61.1$\pm$1.4 & \heatmaplevel{3} 71.6$\pm$0.7 
& \heatmaplevel{6} 51.8$\pm$0.6 & \heatmaplevel{4} 49.1$\pm$0.7 & \heatmaplevel{2} 59.4$\pm$1.5 & \heatmaplevel{4} 71.2$\pm$0.5 
& \heatmaplevel{3} 52.1$\pm$0.7 & \heatmaplevel{3} 51.5$\pm$1.1 & \heatmaplevel{2} 59.6$\pm$1.4 & \heatmaplevel{4} 70.0$\pm$1.2 \\
& FedHGB 
& \heatmaplevel{2} 50.1$\pm$1.4 & \heatmaplevel{6} 51.2$\pm$1.8 & \heatmaplevel{5} 62.2$\pm$0.8 & \heatmaplevel{5} 72.2$\pm$1.0 
& \heatmaplevel{3} 49.0$\pm$1.2 & \heatmaplevel{6} 51.0$\pm$1.2 & \heatmaplevel{4} 61.1$\pm$0.9 & \heatmaplevel{3} 70.5$\pm$0.7 
& \heatmaplevel{5} 53.1$\pm$0.6 & \heatmaplevel{5} 51.9$\pm$1.3 & \heatmaplevel{4} 60.2$\pm$0.7 & \heatmaplevel{3} 69.6$\pm$1.2 \\
& DMML-KD 
& \heatmaplevel{5} 50.6$\pm$0.5 & \heatmaplevel{1} 44.7$\pm$1.2 & \heatmaplevel{6} 62.8$\pm$0.9 & \heatmaplevel{6} 74.7$\pm$1.4 
& \heatmaplevel{5} 51.6$\pm$0.7 & \heatmaplevel{3} 48.3$\pm$0.6 & \heatmaplevel{3} 61.0$\pm$1.3 & \heatmaplevel{6} 74.2$\pm$0.9 
& \heatmaplevel{6} 53.5$\pm$0.4 & \heatmaplevel{1} 50.2$\pm$0.6 & \heatmaplevel{6} 60.6$\pm$1.6 & \heatmaplevel{6} 73.1$\pm$0.7 \\
& \PARSE 
& \heatmaplevel{7} 53.5$\pm$0.8 & \heatmaplevel{7} 52.3$\pm$1.4 & \heatmaplevel{7} 64.6$\pm$1.2 & \heatmaplevel{7} 76.2$\pm$0.9 
& \heatmaplevel{7} 53.7$\pm$0.5 & \heatmaplevel{7} 51.4$\pm$1.2 & \heatmaplevel{7} 63.2$\pm$0.6 & \heatmaplevel{7} 75.1$\pm$0.7 
& \heatmaplevel{7} 54.2$\pm$1.6 & \heatmaplevel{7} 52.1$\pm$0.4 & \heatmaplevel{7} 61.0$\pm$1.7 & \heatmaplevel{7} 75.2$\pm$1.2 \\
\hline
\end{tabular}}
\end{subtable}}
\end{table*}

\section{Additional Results of Different Topologies} \label{app:exp_results}
We also conduct experiments under the default non-IID setting ($\alpha=0.5$) using different topology configurations and agent ratios. All methods are trained for 200 rounds, and the final test accuracies are reported in Table \ref{tab:0.5-chordal-agent} and Table \ref{tab:0.5-random-agent}.

The observations under the Chordal Ring topology (Table \ref{tab:0.5-chordal-agent}) are consistent with those in Section \ref{sec:exp}. \PARSE outperforms both DSGD-Modality and Harmony, which specialize in unimodal and multimodal agents, respectively. While DSGD-Modality often achieves strong performance on unimodal agents, and Harmony excels on multimodal ones, \PARSE achieves state-of-the-art accuracy across all agent types and ratios. 

The results demonstrate that \PARSE does not rely on a specific communication topology between agents; instead, it maintains strong performance across different topological configurations.

\section{Additional Results for Varying Non-IID Configurations}

\begin{table*}[!ht]
\centering
\small
\renewcommand{\arraystretch}{1.2}
\caption{Performance (accuracy \%) of methods under varying agent-ratio scenarios (Dirichlet $\alpha=5.0$, ring topology).}
\label{tab:5.0-ring-agent}
\scalebox{0.75}{
\begin{subtable}[t]{\textwidth}
\resizebox{\textwidth}{!}{
\begin{tabular}{|c|l|ccc|ccc|ccc|}
\hline
\multicolumn{2}{|c|}{\multirow{1}{*}{Agent ratios}} & \multicolumn{3}{c|}{$6:6:18$} & \multicolumn{3}{c|}{$10:10:10$} & \multicolumn{3}{c|}{$13:13:4$} \\
\cline{1-11}
\multicolumn{2}{|c|}{\multirow{1}{*}{\textbf{Agent types}}} & [\textbf{A}] & [\textbf{G}] &[\textbf{AG}] &[\textbf{A}] & [\textbf{G}] &[\textbf{AG}]  &[\textbf{A}] & [\textbf{G}] &[\textbf{AG}]  \\
\hline
\multirow{7}{*}{\rotatebox{90}{\textbf{KU-HAR}}} 
& DSGD-Modality &  
\heatmaplevel{7} 84.5$\pm$1.0 & \heatmaplevel{6} 73.2$\pm$0.9 & \heatmaplevel{4} 88.8$\pm$0.7 & 
\heatmaplevel{4} 84.3$\pm$0.1 & \heatmaplevel{5} 75.0$\pm$0.5 & \heatmaplevel{4} 89.7$\pm$0.2 & 
\heatmaplevel{4} 83.9$\pm$0.6 & \heatmaplevel{5} 77.4$\pm$0.7 & \heatmaplevel{3} 90.5$\pm$0.4 \\
& DSGD-Task & 
\heatmaplevel{3} 82.8$\pm$0.9 & \heatmaplevel{3} 71.5$\pm$1.9 & \heatmaplevel{2} 88.2$\pm$0.7 & 
\heatmaplevel{2} 84.0$\pm$0.3 & \heatmaplevel{4} 74.7$\pm$0.4 & \heatmaplevel{1} 88.3$\pm$0.2 &  
\heatmaplevel{5} 84.0$\pm$0.6 & \heatmaplevel{3} 76.4$\pm$0.8 & \heatmaplevel{1} 86.8$\pm$0.5 \\
& DSGD-Hybrid & 
\heatmaplevel{1} 82.5$\pm$1.2 & \heatmaplevel{2} 69.0$\pm$1.0 & \heatmaplevel{3} 88.5$\pm$0.7 & 
\heatmaplevel{1} 82.6$\pm$0.8 & \heatmaplevel{2} 72.1$\pm$0.6 & \heatmaplevel{3} 89.3$\pm$0.9 & 
\heatmaplevel{3} 82.9$\pm$0.5 & \heatmaplevel{1} 74.1$\pm$0.8 & \heatmaplevel{4} 90.8$\pm$0.4 \\
& Harmony & 
\heatmaplevel{4} 84.1$\pm$0.7 & \heatmaplevel{5} 72.9$\pm$0.7 & \heatmaplevel{6} 91.0$\pm$0.4 & 
\heatmaplevel{6} 84.5$\pm$0.3 & \heatmaplevel{5} 75.0$\pm$0.4 & \heatmaplevel{5} 90.8$\pm$0.2 &  
\heatmaplevel{7} 84.6$\pm$1.0 & \heatmaplevel{6} 77.5$\pm$0.9 & \heatmaplevel{4} 90.8$\pm$0.4 \\
& FedHGB & 
\heatmaplevel{1} 82.5$\pm$0.8 & \heatmaplevel{4} 71.7$\pm$1.2 & \heatmaplevel{1} 88.1$\pm$0.6 & 
\heatmaplevel{2} 84.0$\pm$0.8 & \heatmaplevel{3} 73.7$\pm$1.0 & \heatmaplevel{2} 88.8$\pm$1.2 & 
\heatmaplevel{1} 81.4$\pm$0.6 & \heatmaplevel{2} 74.3$\pm$1.5 & \heatmaplevel{2} 90.1$\pm$0.3 \\
& DMML-KD & 
\heatmaplevel{7} 84.5$\pm$0.4 & \heatmaplevel{1} 68.7$\pm$1.0 & \heatmaplevel{5} 89.8$\pm$0.7 &
\heatmaplevel{5} 84.4$\pm$1.0 & \heatmaplevel{1} 69.6$\pm$1.1 & \heatmaplevel{6} 91.0$\pm$0.9  & 
\heatmaplevel{2} 82.6$\pm$1.2 & \heatmaplevel{5} 77.0$\pm$0.4 & \heatmaplevel{6} 91.0$\pm$0.5 \\
& \PARSE & 
\heatmaplevel{7} 84.5$\pm$1.4 & \heatmaplevel{7} 78.6$\pm$1.0 & \heatmaplevel{7} 91.2$\pm$0.8  & 
\heatmaplevel{7} 86.4$\pm$0.9 & \heatmaplevel{7} 77.1$\pm$0.4 & \heatmaplevel{7} 91.4$\pm$0.3 & 
\heatmaplevel{7} 84.6$\pm$1.2 & \heatmaplevel{7} 78.3$\pm$1.3 & \heatmaplevel{7} 91.5$\pm$0.8 \\
\cline{1-11}
\multicolumn{2}{|c|}{\multirow{1}{*}{\textbf{Agent types}}} & [\textbf{V1}] & [\textbf{V2}] &[\textbf{V1,V2}] &[\textbf{V1}] & [\textbf{V2}] &[\textbf{V1,V2}] & [\textbf{V1}] & [\textbf{V2}] &[\textbf{V1,V2}] \\
\hline
\multirow{7}{*}{\rotatebox{90}{\textbf{ModelNet-40}}} 
& DSGD-Modality & 
\heatmaplevel{5} 77.4$\pm$0.4 & \heatmaplevel{6} 73.2$\pm$1.2 & \heatmaplevel{4} 81.4$\pm$0.5 & 
\heatmaplevel{6} 77.2$\pm$1.4 & \heatmaplevel{4} 69.6$\pm$0.8 & \heatmaplevel{4} 81.4$\pm$1.4 & 
\heatmaplevel{4} 77.2$\pm$1.2 & \heatmaplevel{4} 75.2$\pm$0.8 & \heatmaplevel{2} 76.3$\pm$0.9 \\
& DSGD-Task & 
\heatmaplevel{4} 77.0$\pm$0.9 & \heatmaplevel{1} 66.2$\pm$1.4 & \heatmaplevel{2} 77.5$\pm$1.3 & 
\heatmaplevel{2} 76.4$\pm$1.5 & \heatmaplevel{1} 66.4$\pm$0.7 & \heatmaplevel{1} 76.3$\pm$1.3 & 
\heatmaplevel{1} 75.4$\pm$1.2 & \heatmaplevel{5} 76.3$\pm$0.8 & \heatmaplevel{1} 75.3$\pm$0.6 \\
& DSGD-Hybrid & 
\heatmaplevel{1} 76.4$\pm$0.4 & \heatmaplevel{2} 70.3$\pm$1.2 & \heatmaplevel{1} 74.5$\pm$0.3 & 
\heatmaplevel{4} 76.9$\pm$1.8 & \heatmaplevel{2} 67.5$\pm$1.4 & \heatmaplevel{3} 81.3$\pm$1.3 & 
\heatmaplevel{3} 77.0$\pm$0.6 & \heatmaplevel{1} 68.9$\pm$1.3 & \heatmaplevel{4} 78.6$\pm$0.6 \\
& Harmony & 
\heatmaplevel{2} 76.5$\pm$1.2 & \heatmaplevel{5} 72.1$\pm$1.2 & \heatmaplevel{5} 81.6$\pm$0.7 & 
\heatmaplevel{5} 77.1$\pm$1.3 & \heatmaplevel{6} 75.2$\pm$1.8 & \heatmaplevel{2} 81.0$\pm$1.2 & 
\heatmaplevel{4} 77.2$\pm$0.7 & \heatmaplevel{6} 76.4$\pm$0.3 & \heatmaplevel{5} 80.6$\pm$0.5 \\
& FedHGB & 
\heatmaplevel{3} 76.9$\pm$0.5 & \heatmaplevel{3} 71.0$\pm$1.3 & \heatmaplevel{5} 81.6$\pm$1.8 & 
\heatmaplevel{1} 73.4$\pm$1.3 & \heatmaplevel{5} 70.4$\pm$2.3 & \heatmaplevel{4} 81.4$\pm$1.5 & 
\heatmaplevel{6} 77.5$\pm$0.7 & \heatmaplevel{3} 72.7$\pm$1.5 & \heatmaplevel{6} 80.1$\pm$0.9 \\
& DMML-KD & 
\heatmaplevel{5} 77.4$\pm$0.8 & \heatmaplevel{4} 71.8$\pm$2.0 & \heatmaplevel{3} 80.8$\pm$1.2 & 
\heatmaplevel{3} 76.6$\pm$1.0 & \heatmaplevel{3} 68.8$\pm$1.9 & \heatmaplevel{6} 80.8$\pm$1.0 & 
\heatmaplevel{2} 76.2$\pm$0.8 & \heatmaplevel{2} 71.1$\pm$1.4 & \heatmaplevel{3} 78.2$\pm$0.7 \\
& \PARSE & 
\heatmaplevel{7} 81.1$\pm$0.6 & \heatmaplevel{7} 76.3$\pm$1.0 & \heatmaplevel{7} 82.7$\pm$0.4 & 
\heatmaplevel{7} 77.5$\pm$1.0 & \heatmaplevel{7} 76.3$\pm$1.3 & \heatmaplevel{7} 82.4$\pm$1.1 & 
\heatmaplevel{7} 78.3$\pm$1.6 & \heatmaplevel{7} 76.5$\pm$0.7 & \heatmaplevel{7} 80.8$\pm$0.7 \\
\cline{1-11}
\multicolumn{2}{|c|}{\multirow{1}{*}{\textbf{Agent types}}} & [\textbf{A}] & [\textbf{V}] &[\textbf{AV}] &[\textbf{A}] & [\textbf{V}] &[\textbf{AV}]  &[\textbf{A}] & [\textbf{V}] &[\textbf{AV}]  \\
\hline
\multirow{7}{*}{\rotatebox{90}{\textbf{AVE}}} 
& DSGD-Modality & 
\heatmaplevel{6} 50.2$\pm$0.9 & \heatmaplevel{5} 56.6$\pm$0.5 & \heatmaplevel{3} 68.1$\pm$1.2 & 
\heatmaplevel{6} 51.6$\pm$0.6 & \heatmaplevel{5} 56.0$\pm$0.6 & \heatmaplevel{3} 67.8$\pm$0.2 & 
\heatmaplevel{2} 50.2$\pm$0.3 & \heatmaplevel{4} 55.3$\pm$0.4 & \heatmaplevel{2} 69.0$\pm$0.5 \\
& DSGD-Task & 
\heatmaplevel{1} 46.2$\pm$1.4 & \heatmaplevel{1} 51.4$\pm$0.4 & \heatmaplevel{1} 65.3$\pm$1.0 & 
\heatmaplevel{1} 49.3$\pm$0.5 & \heatmaplevel{1} 51.9$\pm$0.4 & \heatmaplevel{1} 61.5$\pm$0.2 & 
\heatmaplevel{1} 49.3$\pm$0.4 & \heatmaplevel{4} 55.3$\pm$0.6 & \heatmaplevel{1} 57.0$\pm$0.7 \\
& DSGD-Hybrid & 
\heatmaplevel{2} 46.3$\pm$0.5 & \heatmaplevel{2} 52.7$\pm$0.9 & \heatmaplevel{2} 66.4$\pm$0.6 & 
\heatmaplevel{3} 50.3$\pm$0.8 & \heatmaplevel{3} 54.1$\pm$0.3 & \heatmaplevel{5} 68.5$\pm$0.2 & 
\heatmaplevel{3} 50.4$\pm$1.4 & \heatmaplevel{2} 54.8$\pm$0.3 & \heatmaplevel{3} 69.4$\pm$1.0 \\
& Harmony & 
\heatmaplevel{4} 49.2$\pm$0.6 & \heatmaplevel{4} 56.2$\pm$0.9 & \heatmaplevel{4} 69.2$\pm$1.9 & 
\heatmaplevel{5} 51.0$\pm$0.1 & \heatmaplevel{4} 54.7$\pm$0.2 & \heatmaplevel{6} 70.8$\pm$1.0 & 
\heatmaplevel{5} 51.1$\pm$0.7 & \heatmaplevel{3} 55.2$\pm$0.6 & \heatmaplevel{3} 69.4$\pm$1.1 \\
& FedHGB & 
\heatmaplevel{4} 49.2$\pm$1.5 & \heatmaplevel{3} 55.4$\pm$0.6 & \heatmaplevel{4} 69.2$\pm$1.0 & 
\heatmaplevel{4} 50.6$\pm$0.8 & \heatmaplevel{5} 56.0$\pm$0.8 & \heatmaplevel{2} 66.7$\pm$0.7 & 
\heatmaplevel{6} 51.7$\pm$0.4 & \heatmaplevel{6} 55.7$\pm$0.8 & \heatmaplevel{5} 69.6$\pm$0.6 \\
& DMML-KD & 
\heatmaplevel{3} 47.8$\pm$0.4 & \heatmaplevel{6} 56.7$\pm$0.5 & \heatmaplevel{7} 69.4$\pm$1.1 & 
\heatmaplevel{2} 49.5$\pm$0.6 & \heatmaplevel{2} 52.2$\pm$1.3 & \heatmaplevel{4} 68.3$\pm$1.5 & 
\heatmaplevel{4} 51.0$\pm$0.8 & \heatmaplevel{1} 54.2$\pm$0.4 & \heatmaplevel{5} 69.6$\pm$0.7 \\
& \PARSE & 
\heatmaplevel{7} 50.4$\pm$0.7 & \heatmaplevel{7} 57.0$\pm$0.4 & \heatmaplevel{7} 69.4$\pm$0.7 & 
\heatmaplevel{7} 51.9$\pm$1.7 & \heatmaplevel{7} 56.7$\pm$1.3 & \heatmaplevel{7} 71.4$\pm$1.6 & 
\heatmaplevel{7} 52.1$\pm$1.1 & \heatmaplevel{7} 55.8$\pm$1.7 & \heatmaplevel{7} 69.8$\pm$0.6 \\
\hline
\end{tabular}}
\end{subtable}}
\scalebox{0.75}{
\begin{subtable}[t]{\textwidth}
\renewcommand{\arraystretch}{1.2}
\resizebox{\textwidth}{!}{
\begin{tabular}{|c|l|cccc|cccc|cccc|}
\hline
\multicolumn{2}{|c|}{\multirow{1}{*}{Agent ratios}} & \multicolumn{4}{c|}{$6:6:6:22$} & \multicolumn{4}{c|}{$10:10:10:10$} & \multicolumn{4}{c|}{$12:12:12:4$} \\
\cline{1-14}
\multicolumn{2}{|c|}{\multirow{1}{*}{\textbf{Agent types}}} & [\textbf{A}] & [\textbf{V}] &[\textbf{T}] &[\textbf{AVT}] &[\textbf{A}] & [\textbf{V}] &[\textbf{T}] &[\textbf{AVT}]  &[\textbf{A}] & [\textbf{V}] &[\textbf{T}] &[\textbf{AVT}]  \\
\hline
\multirow{7}{*}{\rotatebox{90}{IEMOCAP}} 
& DSGD-Modality & 
\heatmaplevel{6} 51.2$\pm$0.2 & \heatmaplevel{6} 56.8$\pm$1.1 & \heatmaplevel{5} 61.2$\pm$0.4 & \heatmaplevel{4} 74.3$\pm$0.2 & 
\heatmaplevel{5} 51.1$\pm$0.5 & \heatmaplevel{4} 56.8$\pm$0.4 & \heatmaplevel{6} 62.5$\pm$0.4 & \heatmaplevel{4} 74.0$\pm$0.7 & 
\heatmaplevel{1} 51.0$\pm$0.5 & \heatmaplevel{2} 56.2$\pm$0.7 & \heatmaplevel{2} 62.2$\pm$0.3 & \heatmaplevel{3} 72.1$\pm$0.4 \\
& DSGD-Task & 
\heatmaplevel{1} 50.2$\pm$0.5 & \heatmaplevel{3} 53.6$\pm$0.8 & \heatmaplevel{1} 56.3$\pm$0.2 & \heatmaplevel{1} 72.6$\pm$0.2 & 
\heatmaplevel{3} 50.7$\pm$0.4 & \heatmaplevel{6} 57.1$\pm$0.4 & \heatmaplevel{1} 60.3$\pm$0.5 & \heatmaplevel{1} 69.8$\pm$0.2 & 
\heatmaplevel{4} 52.4$\pm$0.2 & \heatmaplevel{5} 56.6$\pm$0.3 & \heatmaplevel{1} 61.7$\pm$0.3 & \heatmaplevel{1} 64.1$\pm$0.5 \\
& DSGD-Hybrid & 
\heatmaplevel{3} 50.7$\pm$1.0 & \heatmaplevel{2} 49.9$\pm$0.4 & \heatmaplevel{6} 61.5$\pm$0.4 & \heatmaplevel{2} 73.6$\pm$0.3 & 
\heatmaplevel{1} 50.2$\pm$0.7 & \heatmaplevel{3} 56.7$\pm$0.5 & \heatmaplevel{4} 62.1$\pm$0.9 & \heatmaplevel{2} 73.5$\pm$0.7 & 
\heatmaplevel{3} 52.3$\pm$0.6 & \heatmaplevel{6} 56.7$\pm$0.5 & \heatmaplevel{4} 62.4$\pm$0.2 & \heatmaplevel{4} 72.6$\pm$0.4 \\
& Harmony & 
\heatmaplevel{2} 50.4$\pm$0.7 & \heatmaplevel{4} 54.3$\pm$0.6 & \heatmaplevel{2} 60.1$\pm$0.4 & \heatmaplevel{3} 74.2$\pm$0.2 & 
\heatmaplevel{4} 50.9$\pm$0.8 & \heatmaplevel{4} 56.8$\pm$0.3 & \heatmaplevel{5} 62.4$\pm$0.1 & \heatmaplevel{3} 73.6$\pm$0.5 & 
\heatmaplevel{4} 52.4$\pm$0.6 & \heatmaplevel{4} 56.3$\pm$0.4 & \heatmaplevel{4} 62.4$\pm$0.5 & \heatmaplevel{2} 69.9$\pm$0.3 \\
& FedHGB & 
\heatmaplevel{5} 51.0$\pm$0.4 & \heatmaplevel{5} 55.2$\pm$0.8 & \heatmaplevel{2} 60.1$\pm$0.7 & \heatmaplevel{5} 75.1$\pm$0.3 & 
\heatmaplevel{2} 50.4$\pm$1.0 & \heatmaplevel{2} 56.1$\pm$0.7 & \heatmaplevel{3} 61.8$\pm$0.5 & \heatmaplevel{5} 74.3$\pm$1.2 & 
\heatmaplevel{2} 52.1$\pm$0.3 & \heatmaplevel{2} 56.2$\pm$0.5 & \heatmaplevel{4} 62.4$\pm$0.3 & \heatmaplevel{5} 74.4$\pm$0.6 \\
& DMML-KD & 
\heatmaplevel{4} 50.9$\pm$0.2 & \heatmaplevel{1} 49.3$\pm$0.4 & \heatmaplevel{3} 60.8$\pm$0.8 & \heatmaplevel{6} 75.3$\pm$0.3 & 
\heatmaplevel{6} 51.9$\pm$1.4 & \heatmaplevel{1} 54.3$\pm$1.1 & \heatmaplevel{2} 61.7$\pm$0.3 & \heatmaplevel{6} 74.8$\pm$0.8 & 
\heatmaplevel{5} 53.2$\pm$0.5 & \heatmaplevel{3} 55.1$\pm$0.8 & \heatmaplevel{5} 62.5$\pm$0.3 & \heatmaplevel{6} 74.5$\pm$0.6 \\
& \PARSE & 
\heatmaplevel{7} 53.0$\pm$0.4 & \heatmaplevel{7} 57.0$\pm$1.0 & \heatmaplevel{7} 62.0$\pm$0.5 & \heatmaplevel{7} 76.5$\pm$0.3 & 
\heatmaplevel{7} 53.2$\pm$0.7 & \heatmaplevel{7} 57.3$\pm$0.4 & \heatmaplevel{7} 63.3$\pm$0.3 & \heatmaplevel{7} 77.2$\pm$0.5 & 
\heatmaplevel{7} 53.9$\pm$0.3 & \heatmaplevel{7} 56.9$\pm$0.5 & \heatmaplevel{7} 62.7$\pm$0.2 & \heatmaplevel{7} 76.7$\pm$0.4 \\

\hline
\end{tabular}}
\end{subtable}}

\end{table*}
\begin{table*}[!ht]
\centering
\small
\renewcommand{\arraystretch}{1.2}
\caption{Performance (accuracy \%) of methods under varying agent-ratio scenarios (Dirichlet $\alpha=0.1$, ring topology).}
\label{tab:0.1-ring-agent}
\scalebox{0.75}{
\begin{subtable}[t]{\textwidth}
\resizebox{\textwidth}{!}{
\begin{tabular}{|c|l|ccc|ccc|ccc|}
\hline
\multicolumn{2}{|c|}{\multirow{1}{*}{Agent ratios}} & \multicolumn{3}{c|}{$6:6:18$} & \multicolumn{3}{c|}{$10:10:10$} & \multicolumn{3}{c|}{$13:13:4$} \\
\cline{1-11}
\multicolumn{2}{|c|}{\multirow{1}{*}{\textbf{Agent types}}} & [\textbf{A}] & [\textbf{G}] &[\textbf{AG}] &[\textbf{A}] & [\textbf{G}] &[\textbf{AG}]  &[\textbf{A}] & [\textbf{G}] &[\textbf{AG}]  \\
\hline
\multirow{7}{*}{\rotatebox{90}{\textbf{KU-HAR}}} 
& {DSGD-Modality} & 
\heatmaplevel{6} 57.4$\pm$1.4 & \heatmaplevel{5} 48.3$\pm$0.8 & \heatmaplevel{2} 52.3$\pm$1.3 & 
\heatmaplevel{6} 57.9$\pm$0.3 & \heatmaplevel{6} 45.8$\pm$0.4 & \heatmaplevel{1} 56.0$\pm$1.1 &
\heatmaplevel{5} 58.3$\pm$1.5 & \heatmaplevel{3} 44.2$\pm$0.9 & \heatmaplevel{1} 61.4$\pm$1.2 \\
& DSGD-Task & 
\heatmaplevel{1} 39.7$\pm$1.8 & \heatmaplevel{3} 47.1$\pm$2.1 & \heatmaplevel{5} 60.4$\pm$0.9 &
\heatmaplevel{3} 49.3$\pm$0.2 & \heatmaplevel{5} 45.5$\pm$0.5 & \heatmaplevel{3} 61.4$\pm$0.3 &
\heatmaplevel{2} 55.3$\pm$1.2 & \heatmaplevel{6} 44.6$\pm$1.7 & \heatmaplevel{3} 64.5$\pm$1.4 \\
& DSGD-Hybrid & 
\heatmaplevel{2} 42.9$\pm$0.8 & \heatmaplevel{1} 42.5$\pm$1.2 & \heatmaplevel{3} 56.7$\pm$1.1 &
\heatmaplevel{2} 48.6$\pm$0.9 & \heatmaplevel{2} 41.2$\pm$0.8 & \heatmaplevel{4} 62.2$\pm$1.3 &
\heatmaplevel{1} 51.3$\pm$1.8 & \heatmaplevel{1} 41.5$\pm$0.6 & \heatmaplevel{2} 62.4$\pm$2.0 \\
& Harmony & 
\heatmaplevel{5} 56.3$\pm$2.1 & \heatmaplevel{4} 47.7$\pm$1.5 & \heatmaplevel{4} 59.7$\pm$1.5 &
\heatmaplevel{4} 55.8$\pm$0.3 & \heatmaplevel{4} 44.7$\pm$0.4 & \heatmaplevel{6} 68.8$\pm$0.3 &
\heatmaplevel{4} 57.6$\pm$1.6 & \heatmaplevel{3} 44.2$\pm$1.7 & \heatmaplevel{5} 66.7$\pm$1.2 \\
& FedHGB & 
\heatmaplevel{3} 43.9$\pm$0.8 & \heatmaplevel{2} 42.6$\pm$1.1 & \heatmaplevel{1} 48.6$\pm$1.3 &
\heatmaplevel{1} 46.2$\pm$0.7 & \heatmaplevel{3} 42.4$\pm$1.0 & \heatmaplevel{2} 60.1$\pm$0.8 &
\heatmaplevel{3} 56.2$\pm$1.0 & \heatmaplevel{2} 43.1$\pm$1.6 & \heatmaplevel{4} 64.8$\pm$1.2 \\
& DMML-KD & 
\heatmaplevel{4} 54.0$\pm$1.2 & \heatmaplevel{6} 53.8$\pm$1.5 & \heatmaplevel{6} 60.5$\pm$0.9 &
\heatmaplevel{6} 56.7$\pm$0.6 & \heatmaplevel{1} 40.3$\pm$0.5 & \heatmaplevel{5} 68.0$\pm$0.7 &
\heatmaplevel{6} 58.8$\pm$1.7 & \heatmaplevel{4} 44.3$\pm$2.5 & \heatmaplevel{6} 67.7$\pm$2.3 \\
& \PARSE & 
\heatmaplevel{7} {58.7$\pm$0.9} & \heatmaplevel{7} {54.0$\pm$1.3} & \heatmaplevel{7} {60.9$\pm$1.7} & 
\heatmaplevel{7} {60.4$\pm$0.7} & \heatmaplevel{7} {46.6$\pm$0.4} & \heatmaplevel{7} {70.7$\pm$0.9} &
\heatmaplevel{7} {59.2$\pm$1.4} & \heatmaplevel{7} {46.8$\pm$1.3} & \heatmaplevel{7} {68.1$\pm$0.8} \\
\cline{1-11}
\multicolumn{2}{|c|}{\multirow{1}{*}{\textbf{Agent types}}} & [\textbf{V1}] & [\textbf{V2}] &[\textbf{V1,V2}] &[\textbf{V1}] & [\textbf{V2}] &[\textbf{V1,V2}] & [\textbf{V1}] & [\textbf{V2}] &[\textbf{V1,V2}] \\
\hline
\multirow{7}{*}{\rotatebox{90}{\textbf{ModelNet-40}}} 
& DSGD-Modality & 
\heatmaplevel{6} 52.3$\pm$1.6 & \heatmaplevel{6} 37.4$\pm$1.4 & \heatmaplevel{4} 51.2$\pm$1.1 &
\heatmaplevel{6} 46.2$\pm$0.8 & \heatmaplevel{6} 44.8$\pm$1.2 & \heatmaplevel{6} 55.7$\pm$1.3 &
\heatmaplevel{5} 41.6$\pm$1.6 & \heatmaplevel{5} 42.4$\pm$1.2 & \heatmaplevel{5} 54.2$\pm$1.5 \\
& DSGD-Task & 
\heatmaplevel{2} 36.8$\pm$1.6 & \heatmaplevel{4} 35.1$\pm$1.3 & \heatmaplevel{1} 43.5$\pm$2.4 &
\heatmaplevel{3} 42.7$\pm$1.4 & \heatmaplevel{3} 39.8$\pm$1.5 & \heatmaplevel{3} 49.3$\pm$0.9 &
\heatmaplevel{2} 39.5$\pm$2.6 & \heatmaplevel{3} 41.9$\pm$1.6 & \heatmaplevel{1} 39.4$\pm$1.1 \\
& DSGD-Hybrid & 
\heatmaplevel{1} 36.7$\pm$1.8 & \heatmaplevel{1} 32.9$\pm$2.3 & \heatmaplevel{2} 44.7$\pm$1.6 & 
\heatmaplevel{1} 36.9$\pm$0.7 & \heatmaplevel{2} 34.8$\pm$1.2 & \heatmaplevel{2} 48.4$\pm$0.5 &
\heatmaplevel{1} 39.3$\pm$1.1 & \heatmaplevel{6} 42.9$\pm$1.5 & \heatmaplevel{4} 48.1$\pm$1.4 \\
& Harmony & 
\heatmaplevel{5} 49.9$\pm$1.4 & \heatmaplevel{3} 34.2$\pm$0.9 & \heatmaplevel{5} 51.8$\pm$2.0 & 
\heatmaplevel{4} 44.2$\pm$0.9 & \heatmaplevel{5} 43.7$\pm$0.7 & \heatmaplevel{5} 54.5$\pm$1.3 &
\heatmaplevel{3} 40.4$\pm$2.7 & \heatmaplevel{4} 42.2$\pm$1.7 & \heatmaplevel{3} 44.8$\pm$2.2 \\
& FedHGB & 
\heatmaplevel{3} 41.4$\pm$1.7 & \heatmaplevel{2} 33.2$\pm$1.6 & \heatmaplevel{3} 47.2$\pm$2.2 & 
\heatmaplevel{2} 40.6$\pm$1.6 & \heatmaplevel{1} 34.0$\pm$1.4 & \heatmaplevel{1} 47.8$\pm$2.0 &
\heatmaplevel{4} 41.1$\pm$2.6 & \heatmaplevel{2} 39.3$\pm$1.8 & \heatmaplevel{2} 44.1$\pm$2.6 \\
& DMML-KD & 
\heatmaplevel{4} 43.6$\pm$2.0 & \heatmaplevel{5} 35.4$\pm$2.3 & \heatmaplevel{6} 56.2$\pm$2.1 & 
\heatmaplevel{5} 45.1$\pm$1.6 & \heatmaplevel{4} 41.3$\pm$2.7 & \heatmaplevel{4} 54.4$\pm$2.3 &
\heatmaplevel{6} 44.5$\pm$1.2 & \heatmaplevel{1} 34.8$\pm$1.8 & \heatmaplevel{6} 61.0$\pm$1.8 \\
& \PARSE & 
\heatmaplevel{7} {55.3$\pm$1.9} & \heatmaplevel{7} {42.3$\pm$1.6} & \heatmaplevel{7} {60.7$\pm$1.7} &
\heatmaplevel{7} {53.6$\pm$1.3} & \heatmaplevel{7} {45.0$\pm$0.4} & \heatmaplevel{7} {62.5$\pm$0.5} &
\heatmaplevel{7} {46.5$\pm$1.7} & \heatmaplevel{7} {47.1$\pm$1.9} & \heatmaplevel{7} {65.4$\pm$1.6} \\
\cline{1-11}
\multicolumn{2}{|c|}{\multirow{1}{*}{\textbf{Agent types}}} & [\textbf{A}] & [\textbf{V}] &[\textbf{AV}] &[\textbf{A}] & [\textbf{V}] &[\textbf{AV}]  &[\textbf{A}] & [\textbf{V}] &[\textbf{AV}]  \\
\hline
\multirow{7}{*}{\rotatebox{90}{\textbf{AVE}}} 
& DSGD-Modality & 
\heatmaplevel{6} 30.4$\pm$1.4 & \heatmaplevel{6} 39.3$\pm$1.5 & \heatmaplevel{4} 43.3$\pm$1.5 & 
\heatmaplevel{6} 34.4$\pm$1.3 & \heatmaplevel{4} 38.4$\pm$1.0 & \heatmaplevel{4} 44.5$\pm$0.7 &
\heatmaplevel{6} 32.6$\pm$1.5 & \heatmaplevel{3} 36.4$\pm$0.9 & \heatmaplevel{4} 46.7$\pm$0.8 \\
& DSGD-Task & 
\heatmaplevel{3} 27.4$\pm$1.0 & \heatmaplevel{1} 33.6$\pm$1.4 & \heatmaplevel{1} 38.0$\pm$0.9 & 
\heatmaplevel{4} 32.1$\pm$1.0 & \heatmaplevel{3} 37.5$\pm$1.1 & \heatmaplevel{1} 35.3$\pm$1.1 &
\heatmaplevel{3} 30.4$\pm$1.4 & \heatmaplevel{2} 35.5$\pm$2.0 & \heatmaplevel{1} 25.6$\pm$0.9 \\
& DSGD-Hybrid & 
\heatmaplevel{1} 25.7$\pm$1.1 & \heatmaplevel{2} 34.9$\pm$1.3 & \heatmaplevel{3} 42.5$\pm$0.7 & 
\heatmaplevel{1} 31.3$\pm$1.6 & \heatmaplevel{2} 35.8$\pm$0.7 & \heatmaplevel{3} 43.0$\pm$0.6 &
\heatmaplevel{1} 29.5$\pm$1.2 & \heatmaplevel{1} 34.9$\pm$1.5 & \heatmaplevel{2} 36.8$\pm$2.1 \\
& Harmony & 
\heatmaplevel{5} 29.9$\pm$1.2 & \heatmaplevel{5} 37.8$\pm$0.9 & \heatmaplevel{2} 41.9$\pm$1.5 & 
\heatmaplevel{5} 32.2$\pm$1.5 & \heatmaplevel{6} 39.3$\pm$0.6 & \heatmaplevel{2} 42.4$\pm$0.8 &
\heatmaplevel{5} 31.7$\pm$0.8 & \heatmaplevel{3} 36.4$\pm$1.0 & \heatmaplevel{5} 48.0$\pm$1.1 \\
& FedHGB & 
\heatmaplevel{2} 27.3$\pm$1.6 & \heatmaplevel{3} 37.4$\pm$1.9 & \heatmaplevel{5} 44.8$\pm$1.5 & 
\heatmaplevel{2} 31.6$\pm$2.0 & \heatmaplevel{5} 38.7$\pm$0.8 & \heatmaplevel{5} 44.7$\pm$1.3 &
\heatmaplevel{4} 30.7$\pm$1.2 & \heatmaplevel{5} 37.1$\pm$1.4 & \heatmaplevel{3} 39.0$\pm$0.7 \\
& DMML-KD & 
\heatmaplevel{4} 28.7$\pm$0.9 & \heatmaplevel{3} 37.4$\pm$1.0 & \heatmaplevel{5} 46.3$\pm$1.8 & 
\heatmaplevel{2} 31.9$\pm$0.5 & \heatmaplevel{1} 31.8$\pm$0.6 & \heatmaplevel{5} 47.5$\pm$1.3 &
\heatmaplevel{2} 29.7$\pm$1.9 & \heatmaplevel{2} 37.2$\pm$1.7 & \heatmaplevel{6} 51.3$\pm$1.1 \\
& \PARSE & 
\heatmaplevel{7} {31.1$\pm$0.8} & \heatmaplevel{7} {40.7$\pm$1.0} & \heatmaplevel{7} {46.6$\pm$1.5} &
\heatmaplevel{7} {34.9$\pm$1.4} & \heatmaplevel{7} {39.5$\pm$0.9} & \heatmaplevel{7} {51.8$\pm$0.8} &
\heatmaplevel{7} {33.6$\pm$0.8} & \heatmaplevel{7} {39.2$\pm$1.3} & \heatmaplevel{7} {52.7$\pm$1.4} \\
\hline
\end{tabular}}
\end{subtable}}
\scalebox{0.75}{
\begin{subtable}[t]{\textwidth}
\renewcommand{\arraystretch}{1.2}
\resizebox{\textwidth}{!}{
\begin{tabular}{|c|l|cccc|cccc|cccc|}
\hline
\multicolumn{2}{|c|}{\multirow{1}{*}{Agent ratios}} & \multicolumn{4}{c|}{$6:6:6:22$} & \multicolumn{4}{c|}{$10:10:10:10$} & \multicolumn{4}{c|}{$12:12:12:4$} \\
\cline{1-14}
\multicolumn{2}{|c|}{\multirow{1}{*}{\textbf{Agent types}}} & [\textbf{A}] & [\textbf{V}] &[\textbf{T}] &[\textbf{AVT}] &[\textbf{A}] & [\textbf{V}] &[\textbf{T}] &[\textbf{AVT}]  &[\textbf{A}] & [\textbf{V}] &[\textbf{T}] &[\textbf{AVT}]  \\
\hline
\multirow{7}{*}{\rotatebox{90}{IEMOCAP}} 
& DSGD-Modality &  
\heatmaplevel{4} 35.5$\pm$1.3 & \heatmaplevel{5} 41.2$\pm$1.1 & \heatmaplevel{2} 54.1$\pm$1.6 & \heatmaplevel{2} 49.8$\pm$1.6 & 
\heatmaplevel{5} 37.4$\pm$0.9 & \heatmaplevel{6} 50.1$\pm$0.7 & \heatmaplevel{3} 48.0$\pm$0.9 & \heatmaplevel{1} 51.5$\pm$1.7 &
\heatmaplevel{4} 37.5$\pm$1.7 & \heatmaplevel{2} 45.1$\pm$1.2 & \heatmaplevel{2} 47.6$\pm$0.8 & \heatmaplevel{5} 63.0$\pm$0.5 \\
& DSGD-Task & 
\heatmaplevel{1} 34.5$\pm$1.2 & \heatmaplevel{1} 37.8$\pm$1.2 & \heatmaplevel{1} 46.4$\pm$0.5 & \heatmaplevel{4} 55.4$\pm$1.3 & 
\heatmaplevel{6} 38.1$\pm$0.5 & \heatmaplevel{2} 49.1$\pm$0.4 & \heatmaplevel{2} 47.6$\pm$0.5 & \heatmaplevel{4} 54.4$\pm$0.3 &
\heatmaplevel{2} 37.0$\pm$2.3 & \heatmaplevel{5} 46.5$\pm$1.3 & \heatmaplevel{5} 48.4$\pm$0.7 & \heatmaplevel{1} 52.7$\pm$0.4 \\
& DSGD-Hybrid & 
\heatmaplevel{2} 34.7$\pm$1.4 & \heatmaplevel{2} 40.4$\pm$1.2 & \heatmaplevel{5} 55.8$\pm$1.3 & \heatmaplevel{5} 55.9$\pm$0.6 & 
\heatmaplevel{1} 35.4$\pm$1.5 & \heatmaplevel{4} 49.4$\pm$0.9 & \heatmaplevel{6} 50.9$\pm$0.8 & \heatmaplevel{5} 56.8$\pm$0.5 &
\heatmaplevel{3} 37.1$\pm$1.9 & \heatmaplevel{3} 45.6$\pm$0.7 & \heatmaplevel{4} 48.0$\pm$0.9 & \heatmaplevel{2} 61.2$\pm$0.8 \\
& Harmony & 
\heatmaplevel{3} 35.3$\pm$2.1 & \heatmaplevel{4} 40.8$\pm$0.9 & \heatmaplevel{3} 54.8$\pm$2.3 & \heatmaplevel{3} 52.7$\pm$1.7 & 
\heatmaplevel{2} 36.1$\pm$0.7 & \heatmaplevel{5} 49.5$\pm$1.8 & \heatmaplevel{5} 48.7$\pm$1.5 & \heatmaplevel{3} 53.2$\pm$0.9 &
\heatmaplevel{5} 37.6$\pm$1.5 & \heatmaplevel{5} 46.5$\pm$2.3 & \heatmaplevel{6} 48.5$\pm$1.4 & \heatmaplevel{4} 62.9$\pm$0.8 \\
& FedHGB & 
\heatmaplevel{6} 36.2$\pm$2.0 & \heatmaplevel{6} 41.3$\pm$1.9 & \heatmaplevel{4} 55.3$\pm$1.6 & \heatmaplevel{1} 49.3$\pm$1.1 & 
\heatmaplevel{4} 36.4$\pm$1.7 & \heatmaplevel{2} 49.1$\pm$1.2 & \heatmaplevel{1} 46.7$\pm$1.3 & \heatmaplevel{2} 53.0$\pm$0.8 &
\heatmaplevel{6} 38.4$\pm$1.2 & \heatmaplevel{3} 45.6$\pm$0.8 & \heatmaplevel{1} 47.0$\pm$0.9 & \heatmaplevel{2} 61.2$\pm$0.6 \\
& DMML-KD & 
\heatmaplevel{5} 36.0$\pm$1.3 & \heatmaplevel{3} 40.6$\pm$1.0 & \heatmaplevel{6} 56.0$\pm$1.9 & \heatmaplevel{6} 60.3$\pm$2.1 & 
\heatmaplevel{3} 36.1$\pm$1.0 & \heatmaplevel{1} 48.0$\pm$0.4 & \heatmaplevel{4} 48.6$\pm$0.6 & \heatmaplevel{6} 61.5$\pm$1.4 &
\heatmaplevel{1} 35.6$\pm$1.3 & \heatmaplevel{1} 45.0$\pm$0.9 & \heatmaplevel{3} 47.9$\pm$1.3 & \heatmaplevel{6} 69.8$\pm$1.0 \\
& \PARSE & 
\heatmaplevel{7} 36.7$\pm$1.8 & \heatmaplevel{7} 41.8$\pm$1.5 & \heatmaplevel{7} 56.0$\pm$0.7 & \heatmaplevel{7} 60.5$\pm$0.6 & 
\heatmaplevel{7} 38.4$\pm$0.4 & \heatmaplevel{7} 50.6$\pm$1.1 & \heatmaplevel{7} 51.7$\pm$1.8 & \heatmaplevel{7} 65.1$\pm$0.7 &
\heatmaplevel{7} 38.7$\pm$1.7 & \heatmaplevel{7} 47.1$\pm$0.6 & \heatmaplevel{7} 48.8$\pm$0.5 & \heatmaplevel{7} 70.4$\pm$1.2 \\
\hline
\end{tabular}}
\end{subtable}}
\end{table*}

To provide a more fine-grained analysis of our method under varying non-IID conditions, we conduct additional experiments comparing agent ratios across different levels of heterogeneity. These results, presented in Table \ref{tab:5.0-ring-agent} and Table \ref{tab:0.1-ring-agent}, complement the main findings in Section \ref{sec:exp}. Across all evaluated configurations, \PARSE consistently achieves the highest accuracy, demonstrating robust performance for multimodal agents and unimodal agents across varying agent ratios.

When the non-IID level is low ($\alpha=5.0$), \PARSE outperforms baseline methods by a notable margin. For instance, it surpasses other methods by at least 3.0\% on modality V2 of the ModelNet-40 dataset, when the proportion of unimodal agents is small. On the IEMOCAP dataset, \PARSE improves multimodal performance by approximately 1.2--2.5\% across various settings, demonstrating its effectiveness in learning modality-interactive features.
Even under settings where baseline methods struggle, \PARSE continues to deliver substantial performance gains, highlighting its robustness to agent heterogeneity and adverse modality configurations.
At a higher non-IID level ($\alpha=0.1$), the improvements become even more pronounced. \PARSE yields significant accuracy boosts for multimodal agents, particularly on the ModelNet-40 and AVE datasets. The results further suggest the strength of \PARSE in highly heterogeneous environments.

The results confirm that our method benefits both multimodal and unimodal agents simultaneously. The proposed \textbf{feature fission} and \textbf{partial alignment} mechanisms facilitate effective knowledge sharing across modality-heterogeneous agents and mitigate gradient misalignment, while the \textbf{contrastive loss} further distills cross-modality shared features to enhance the performance of unimodal agents.

\section{Feature Split Sweep} \label{app:splits}

To examine how the feature split ratio influences performance and to assess its potential for future work, we conduct a split-sweep study on all benchmarks: KU-HAR (total dim = 192), AVE (total dim = 384), ModelNet-40 (total dim = 384), and IEMOCAP (total dim = 384). In each sweep, we vary the dimensionality of one branch while dividing the remaining budget equally between the other two. For example, setting the unique branch of AVE to $64$ dimensions assigns $160$ dimensions each to the redundant and synergistic branches, preserving the total of $384$. All other settings follow our default ablation protocol.

\begin{table}[ht]
\centering
\caption{Performance (Acc.) vs.\ feature split size across datasets.}
\label{tab:splits-up}

\begin{subtable}{\linewidth}
\caption{KU-HAR}
\centering
\scriptsize
\setlength{\tabcolsep}{4pt}
\begin{tabular}{llcccc}
\toprule
\textbf{Varying Split} & \textbf{Metric} & \textbf{32d} & \textbf{64d} & \textbf{96d} & \textbf{128d} \\
\midrule
\multirow{2}{*}{Unique}
  & Unique-only & 83.5 & 86.2 & 86.7 & 86.1 \\
  & Combined    & 87.8 & 88.6 & 88.8 & 88.5 \\
\midrule
\multirow{2}{*}{Redundant}
  & Redundant-only & 87.2 & 87.8 & 87.4 & 87.6 \\
  & Combined       & 88.7 & 88.6 & 87.6 & 87.0 \\
\midrule
\multirow{2}{*}{Synergistic}
  & Synergistic-only & 54.5 & 61.5 & 63.7 & 67.2 \\
  & Combined         & 86.9 & 88.6 & 89.5 & 88.3 \\
\bottomrule
\end{tabular}
\end{subtable}

\vspace{0.8em}

\begin{subtable}{\linewidth}
\caption{AVE}
\centering
\scriptsize
\setlength{\tabcolsep}{4pt}
\begin{tabular}{llcccc}
\toprule
\textbf{Varying Split} & \textbf{Metric} & \textbf{64d} & \textbf{128d} & \textbf{192d} & \textbf{256d} \\
\midrule
\multirow{2}{*}{Unique}
  & Unique-only & 59.2 & 61.7 & 61.5 & 61.3 \\
  & Combined    & 63.6 & 64.7 & 62.8 & 61.9 \\
\midrule
\multirow{2}{*}{Redundant}
  & Redundant-only & 58.9 & 63.1 & 63.3 & 62.5 \\
  & Combined       & 63.1 & 64.7 & 63.9 & 62.6 \\
\midrule
\multirow{2}{*}{Synergistic}
  & Synergistic-only & 26.2 & 34.4 & 35.9 & 36.2 \\
  & Combined         & 61.4 & 64.7 & 65.5 & 62.3 \\
\bottomrule
\end{tabular}
\end{subtable}

\vspace{0.8em}

\begin{subtable}{\linewidth}
\caption{ModelNet-40}
\centering
\scriptsize
\setlength{\tabcolsep}{4pt}
\begin{tabular}{llcccc}
\toprule
\textbf{Varying Split} & \textbf{Metric} & \textbf{64d} & \textbf{128d} & \textbf{192d} & \textbf{256d} \\
\midrule
\multirow{2}{*}{Unique}
  & Unique-only & 72.3 & 75.7 & 76.2 & 76.6 \\
  & Combined    & 79.2 & 79.3 & 79.7 & 80.9 \\
\midrule
\multirow{2}{*}{Redundant}
  & Redundant-only & 74.1 & 75.3 & 76.8 & 77.2 \\
  & Combined       & 79.6 & 79.3 & 81.2 & 81.4 \\
\midrule
\multirow{2}{*}{Synergistic}
  & Synergistic-only & 48.0 & 51.5 & 56.3 & 61.8 \\
  & Combined         & 80.4 & 79.3 & 79.6 & 78.6 \\
\bottomrule
\end{tabular}
\end{subtable}

\vspace{0.8em}

\begin{subtable}{\linewidth}
\caption{IEMOCAP}
\centering
\scriptsize
\setlength{\tabcolsep}{4pt}
\begin{tabular}{llcccc}
\toprule
\textbf{Varying Split} & \textbf{Metric} & \textbf{64d} & \textbf{128d} & \textbf{192d} & \textbf{256d} \\
\midrule
\multirow{2}{*}{Unique}
  & Unique-only & 51.3 & 67.3 & 68.2 & 69.0 \\
  & Combined    & 69.8 & 73.2 & 71.7 & 70.3 \\
\midrule
\multirow{2}{*}{Redundant}
  & Redundant-only & 57.5 & 70.3 & 64.7 & 63.2 \\
  & Combined       & 68.4 & 73.2 & 70.5 & 70.2 \\
\midrule
\multirow{2}{*}{Synergistic}
  & Synergistic-only & 37.8 & 45.1 & 47.5 & 51.5 \\
  & Combined         & 69.6 & 73.2 & 70.7 & 68.3 \\
\bottomrule
\end{tabular}
\end{subtable}
\end{table}

The Table \ref{tab:splits-up} reports, on dataset KU-HAR, for each chosen split size, the overall accuracy of multimodal agents (when the split is integrated) and the split-only accuracy obtained when using only that split. We observe that enlarging the unique branch from $32d$ to $128d$ lifts its stand-alone accuracy ($+3$ pp) but the overall accuracy plateaus (88.5–88.8 \%).  Oversizing redundancy even degrades overall accuracy (88.7 to 87.0 \%).  Synergy helps most at a moderate size (overall peak 89.5 \% at $96d$).
Similarly, on AVE, we observe that unique-only and redundant-only scores saturate beyond $128d$, while synergy yields the best overall accuracy at a mid-range size (65.5 \% at $192d$).  Over-allocating any single branch hurts the combined model. On ModelNet-40 and IEMOCAP, the results are similar, and the overall performance is more sensitive to redundant feature size.

The default even split across unique, redundant, and synergistic features offers a robust trade-off.  Because the optimum is dataset-dependent, data-driven adaptive allocation is indeed an important avenue for future work.

\section{Fusion Method Comparison} \label{app:fusion}

By default, we use a simple averaging as the fusion method. To assess whether it is too crude for capturing synergistic information, we compared our default mean fusion with five stronger operators: concatenation+Linear, summation+Linear, gated fusion \cite{Xue_2023_CVPR}, cross attention \cite{zhang_2022_MM}, and Hadamard product \cite{Kim2017}, where "Linear" means a linear map that can be learned.  Results on all benchmarks are summarized in Table~\ref{tab:fusion_comparison_4up}.

\begin{table}[ht]
\centering
\caption{Comparison of fusion methods across datasets (mean $\pm$ std).}
\label{tab:fusion_comparison_4up}

\begin{subtable}{\linewidth}
\caption{KU-HAR}
\centering
\scriptsize
\setlength{\tabcolsep}{6pt}
\begin{tabular}{lcc}
\toprule
\textbf{Fusion Method} & \textbf{Overall (\%)} & \textbf{Synergy (\%)} \\
\midrule
Mean (default)         & 88.6$\pm$0.5 & 61.5$\pm$1.0 \\
Concatenation + Linear & 87.1$\pm$0.9 & 60.3$\pm$0.8 \\
Summation + Linear     & 89.0$\pm$1.1 & 62.2$\pm$2.1 \\
Gated Fusion           & 89.0$\pm$1.4 & 64.7$\pm$1.5 \\
Cross-Attention        & 89.2$\pm$1.2 & 62.1$\pm$0.7 \\
Hadamard Product       & 88.6$\pm$1.3 & 58.7$\pm$1.8 \\
\bottomrule
\end{tabular}
\end{subtable}

\vspace{0.8em}

\begin{subtable}{\linewidth}
\caption{AVE}
\centering
\scriptsize
\setlength{\tabcolsep}{6pt}
\begin{tabular}{lcc}
\toprule
\textbf{Fusion Method} & \textbf{Overall (\%)} & \textbf{Synergy (\%)} \\
\midrule
Mean (default)         & 64.7$\pm$1.3 & 34.4$\pm$1.6 \\
Concatenation + Linear & 63.2$\pm$1.5 & 33.0$\pm$0.9 \\
Summation + Linear     & 65.4$\pm$1.0 & 39.3$\pm$1.3 \\
Gated Fusion           & 63.8$\pm$1.1 & 33.9$\pm$1.7 \\
Cross-Attention        & 64.2$\pm$1.9 & 34.3$\pm$1.0 \\
Hadamard Product       & 65.8$\pm$1.6 & 38.5$\pm$1.4 \\
\bottomrule
\end{tabular}
\end{subtable}

\vspace{0.8em}

\begin{subtable}{\linewidth}
\caption{ModelNet-40}
\centering
\scriptsize
\setlength{\tabcolsep}{6pt}
\begin{tabular}{lcc}
\toprule
\textbf{Fusion Method} & \textbf{Overall (\%)} & \textbf{Synergy (\%)} \\
\midrule
Mean (default)         & 79.3$\pm$0.9 & 51.5$\pm$0.8 \\
Concatenation + Linear & 79.7$\pm$1.4 & 51.9$\pm$0.4 \\
Summation + Linear     & 80.8$\pm$0.6 & 53.2$\pm$1.2 \\
Gated Fusion           & 81.4$\pm$0.5 & 56.1$\pm$1.0 \\
Cross-Attention        & 79.6$\pm$0.8 & 51.7$\pm$0.6 \\
Hadamard Product       & 80.4$\pm$0.5 & 52.7$\pm$0.3 \\
\bottomrule
\end{tabular}
\end{subtable}

\vspace{0.8em}

\begin{subtable}{\linewidth}
\caption{IEMOCAP}
\centering
\scriptsize
\setlength{\tabcolsep}{6pt}
\begin{tabular}{lcc}
\toprule
\textbf{Fusion Method} & \textbf{Overall (\%)} & \textbf{Synergy (\%)} \\
\midrule
Mean (default)         & 73.2$\pm$0.6 & 42.3$\pm$1.2 \\
Concatenation + Linear & 71.7$\pm$1.3 & 41.1$\pm$0.7 \\
Summation + Linear     & 70.2$\pm$1.0 & 38.3$\pm$0.8 \\
Gated Fusion           & 74.8$\pm$0.7 & 46.5$\pm$1.4 \\
Cross-Attention        & 70.3$\pm$0.6 & 39.6$\pm$1.3 \\
Hadamard Product       & 67.5$\pm$0.9 & 37.2$\pm$0.4 \\
\bottomrule
\end{tabular}
\end{subtable}
\end{table}

As shown in the results, gated fusion yields the highest synergistic-only and overall performance, improving accuracy on ModelNet-40 and IEMOCAP (by respectively $+2$ and $+1.5$ pp compared to mean fusion). Notably, \PARSE performs slightly better with cross-attention on KU-HAR, while on AVE it achieves its best results with summation+Linear fusion. While some heavier operators can improve synergistic feature alignment on specific datasets, they also introduce learnable parameters that must be exchanged across agents each round. This exchange increases both bandwidth consumption and sensitivity to non-IID drift. By contrast, mean fusion minimizes coordination costs while still providing competitive accuracy, making it a practical baseline in server-free (DFL) environments. 

A key advantage of \PARSE is that it is \emph{fusion-agnostic}.Our default choice is mean fusion, which is lightweight, parameter-free, scales linearly with the number of neighbors, and avoids synchronization overhead in decentralized settings. The framework itself does not depend on this choice, and any richer operators can be adopted to the framework. 

\begin{figure*}[t]
  \centering
  \begin{subfigure}[t]{0.25\textwidth}
    \centering\includegraphics[width=\linewidth]{exp/feat_har_acc.png}
    \subcaption{KU-HAR accelerometer.}\label{fig:a}
  \end{subfigure}
  \begin{subfigure}[t]{0.25\textwidth}
    \centering\includegraphics[width=\linewidth]{exp/feat_har_gyro.png}
    \subcaption{KU-HAR gyroscope.}\label{fig:b}
  \end{subfigure}
  \begin{subfigure}[t]{0.25\textwidth}
    \centering\includegraphics[width=\linewidth]{exp/feat_har_both.png}
    \subcaption{KU-HAR fused features.}\label{fig:c}
  \end{subfigure}

  \medskip 

  \begin{subfigure}[t]{0.25\textwidth}
    \centering\includegraphics[width=\linewidth]{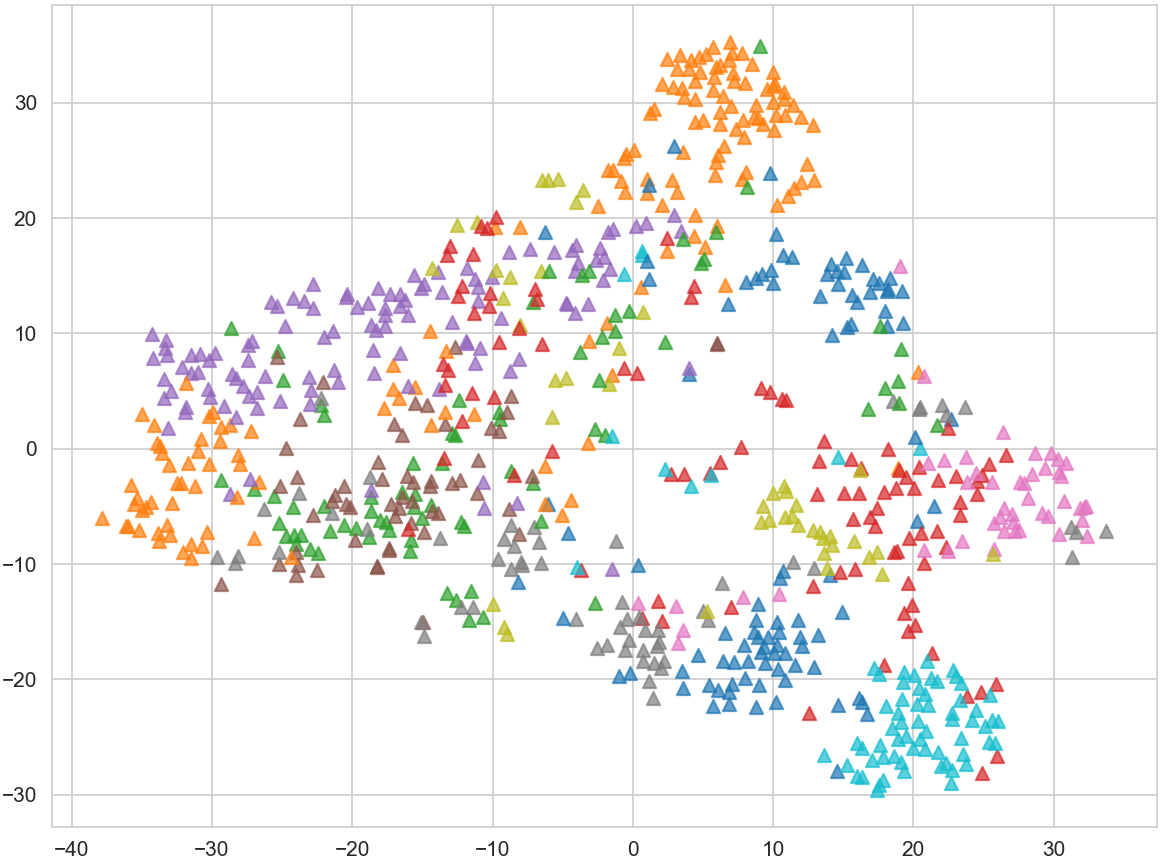}
    \subcaption{AVE audio.}\label{fig:d}
  \end{subfigure}
  \begin{subfigure}[t]{0.25\textwidth}
    \centering\includegraphics[width=\linewidth]{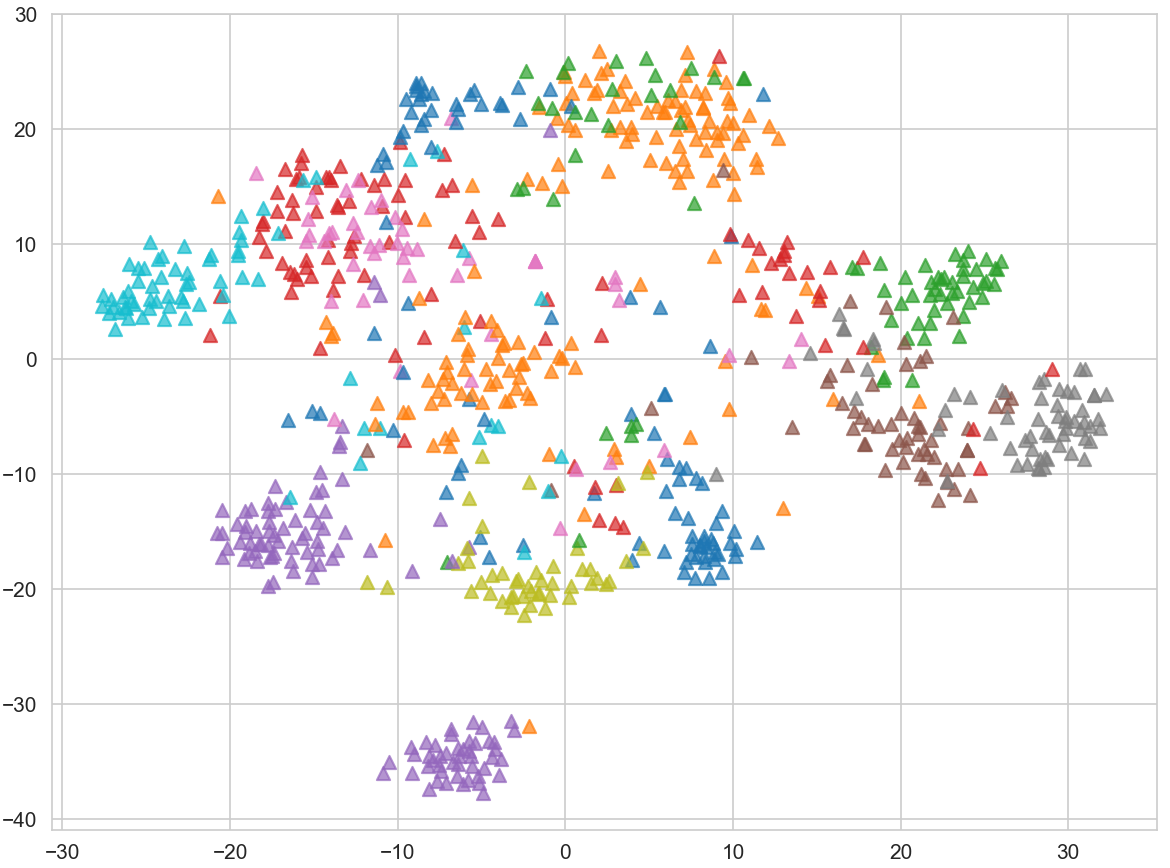}
    \subcaption{AVE video.}\label{fig:e}
  \end{subfigure}
  \begin{subfigure}[t]{0.25\textwidth}
    \centering\includegraphics[width=\linewidth]{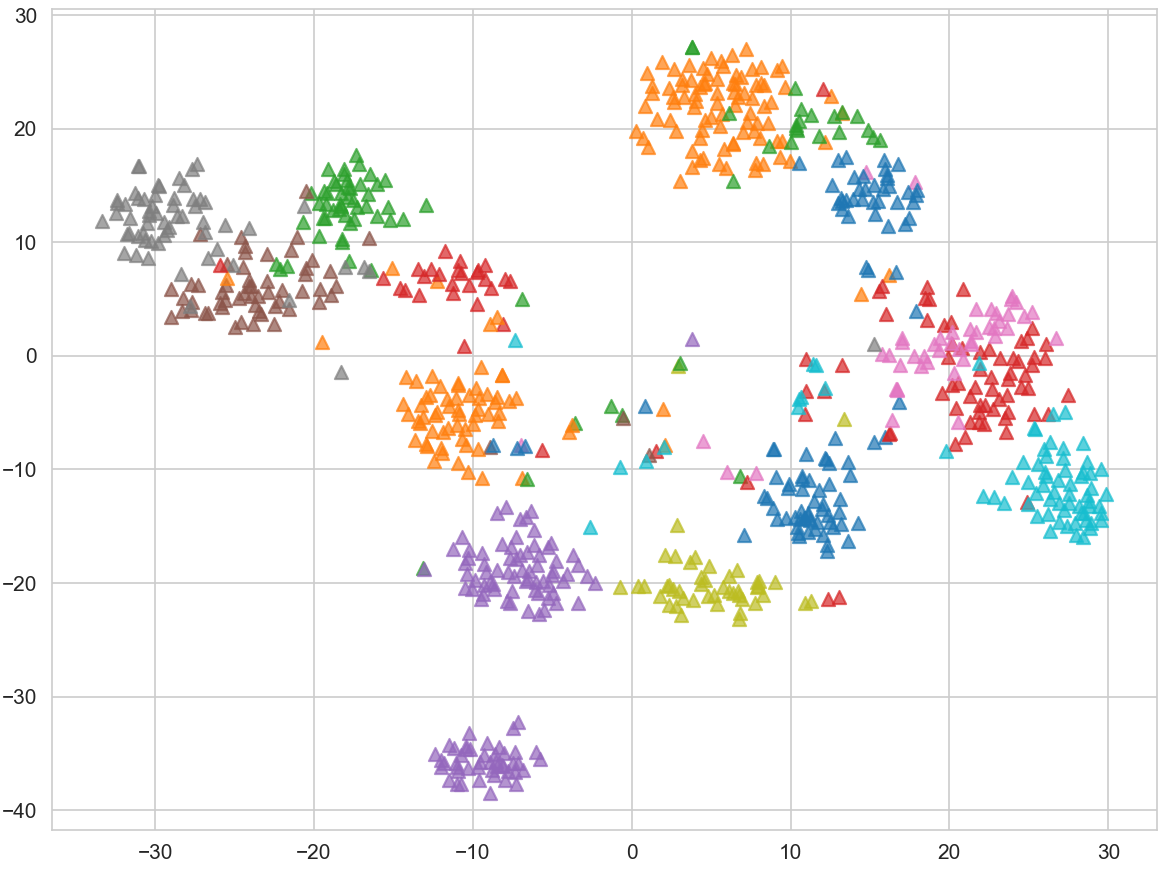}
    \subcaption{AVE fused features.}\label{fig:f}
  \end{subfigure}

  \medskip

  \begin{subfigure}[t]{0.25\textwidth}
    \centering\includegraphics[width=\linewidth]{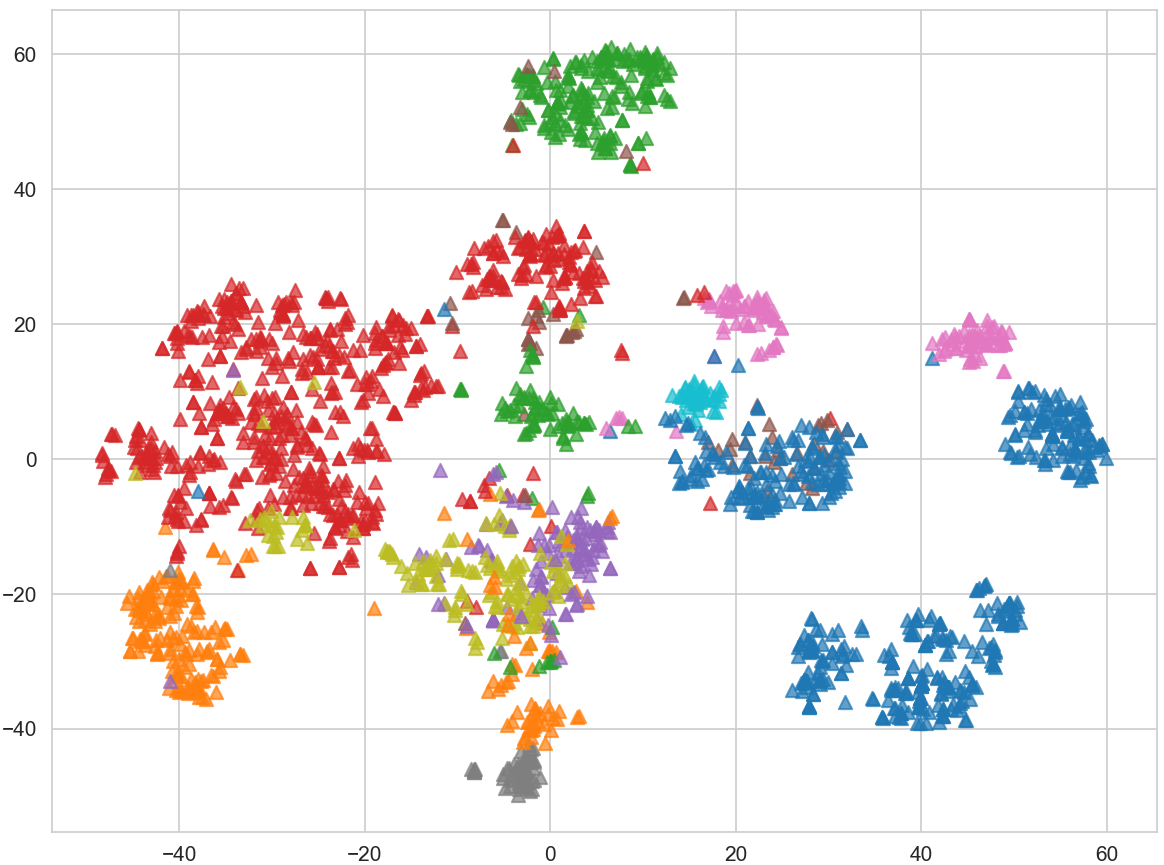}
    \subcaption{ModelNet-40 V1.}\label{fig:g}
  \end{subfigure}
  \begin{subfigure}[t]{0.25\textwidth}
    \centering\includegraphics[width=\linewidth]{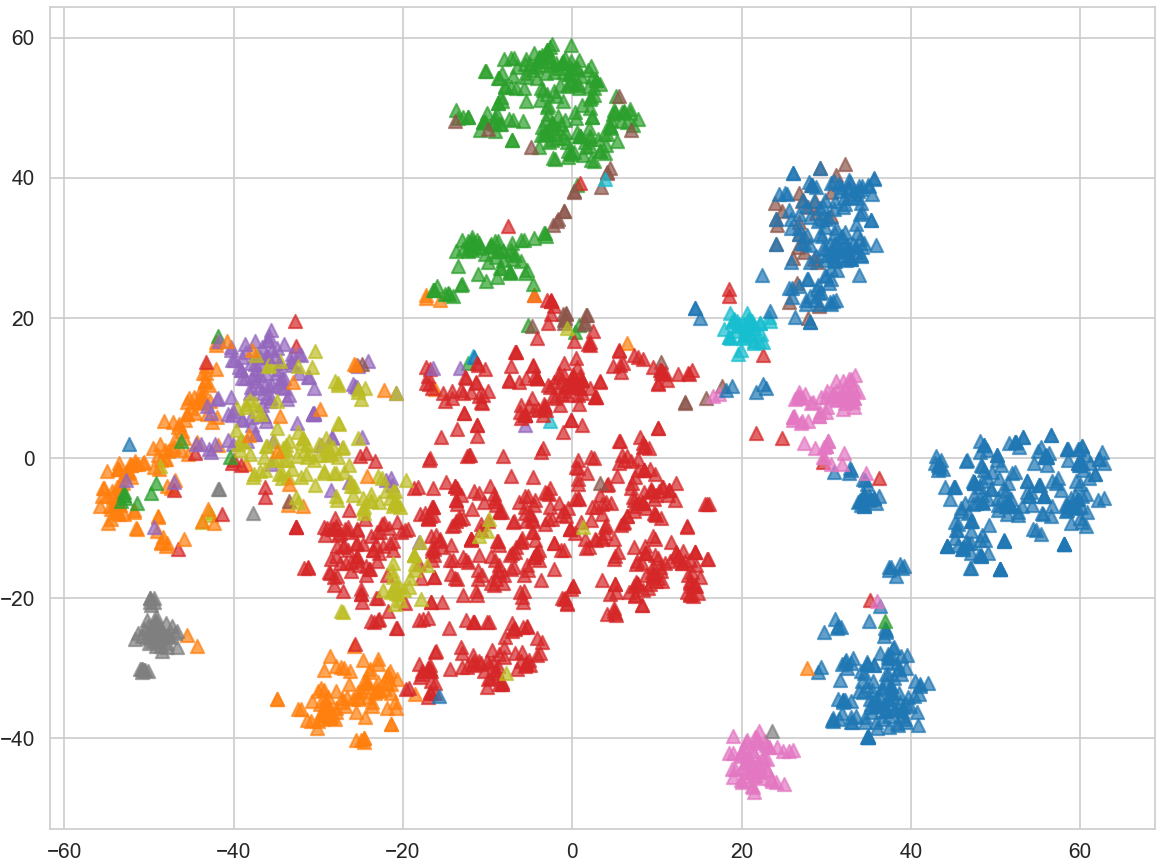}
    \subcaption{ModelNet-40 V2.}\label{fig:h}
  \end{subfigure}
  \begin{subfigure}[t]{0.25\textwidth}
    \centering\includegraphics[width=\linewidth]{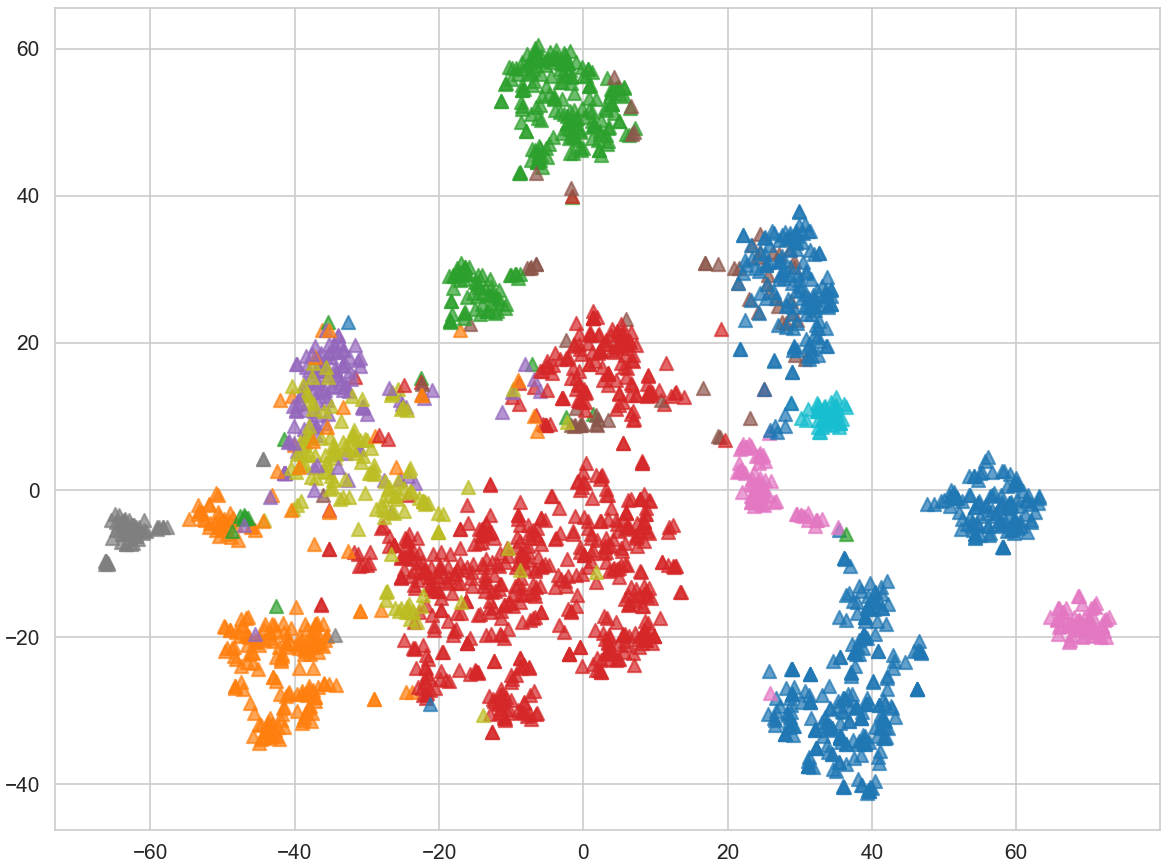}
    \subcaption{ModelNet-40 fused features.}\label{fig:i}
  \end{subfigure}

  \medskip

  \begin{subfigure}[t]{0.22\textwidth}
    \centering\includegraphics[width=\linewidth]{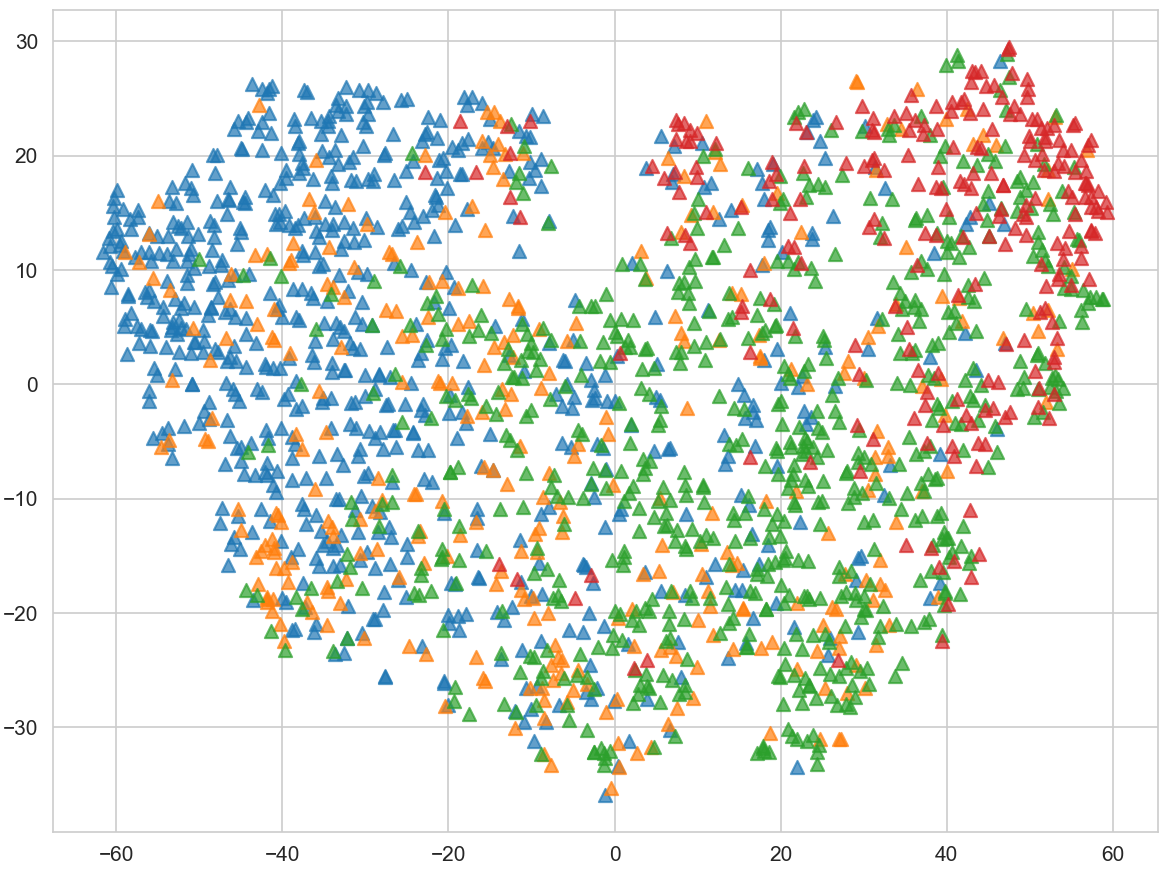}
    \subcaption{IEMOCAP audio.}\label{fig:j}
  \end{subfigure}
  \begin{subfigure}[t]{0.22\textwidth}
    \centering\includegraphics[width=\linewidth]{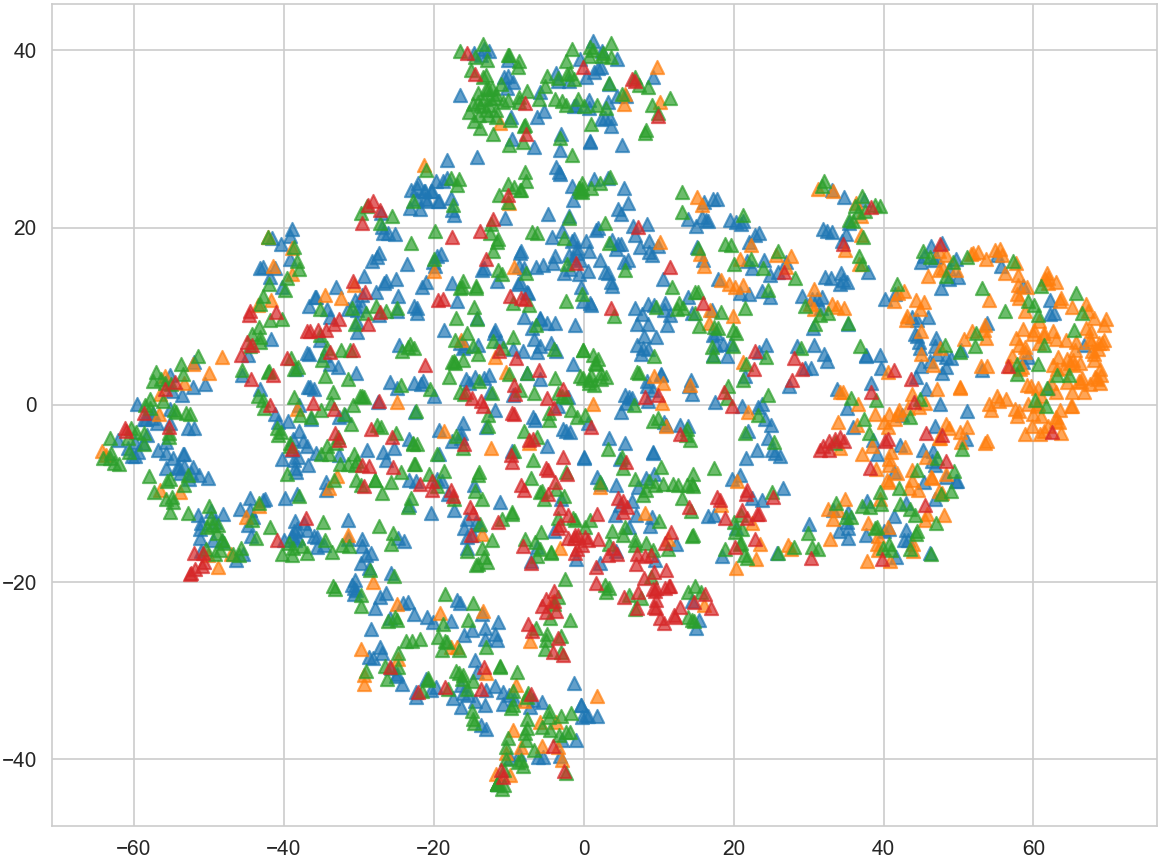}
    \subcaption{IEMOCAP video.}\label{fig:k}
  \end{subfigure}
  \begin{subfigure}[t]{0.22\textwidth}
    \centering\includegraphics[width=\linewidth]{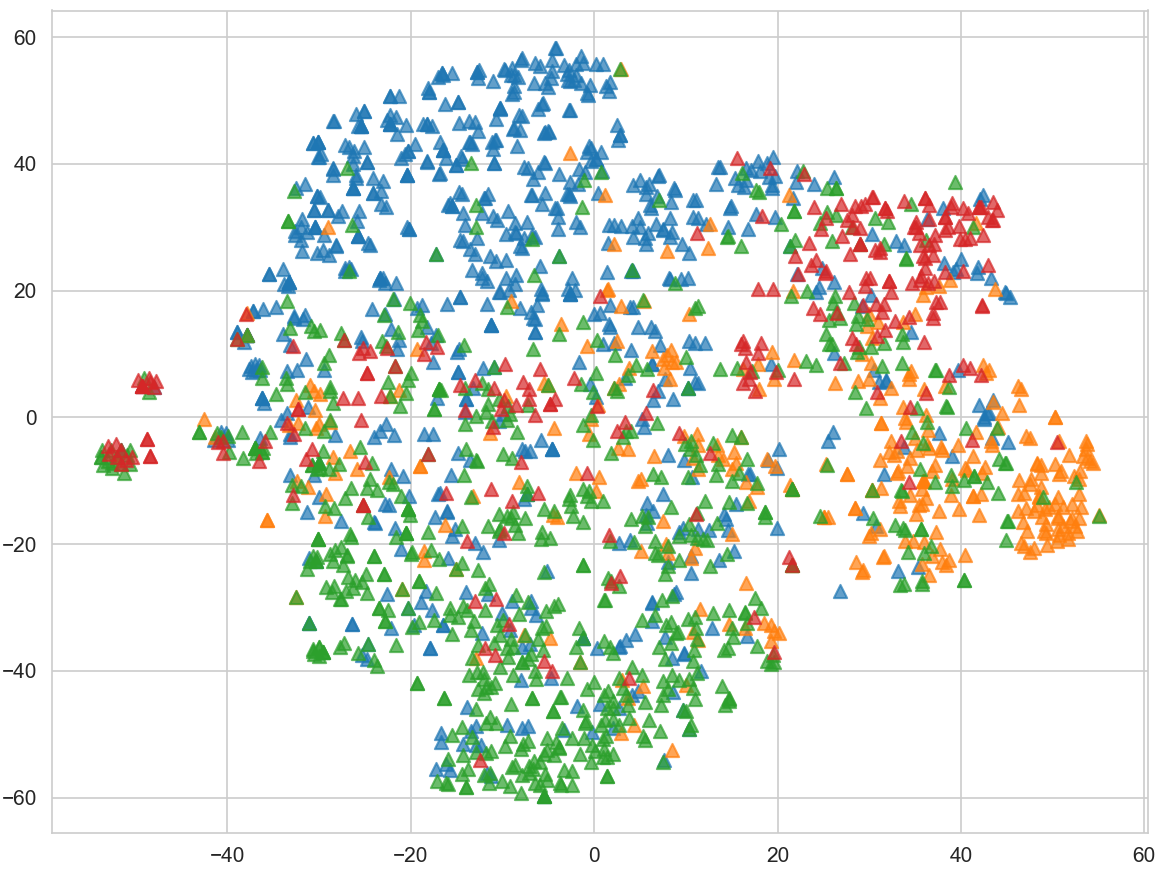}
    \subcaption{IEMOCAP text.}\label{fig:l}
  \end{subfigure}
  \begin{subfigure}[t]{0.22\textwidth}
    \centering\includegraphics[width=\linewidth]{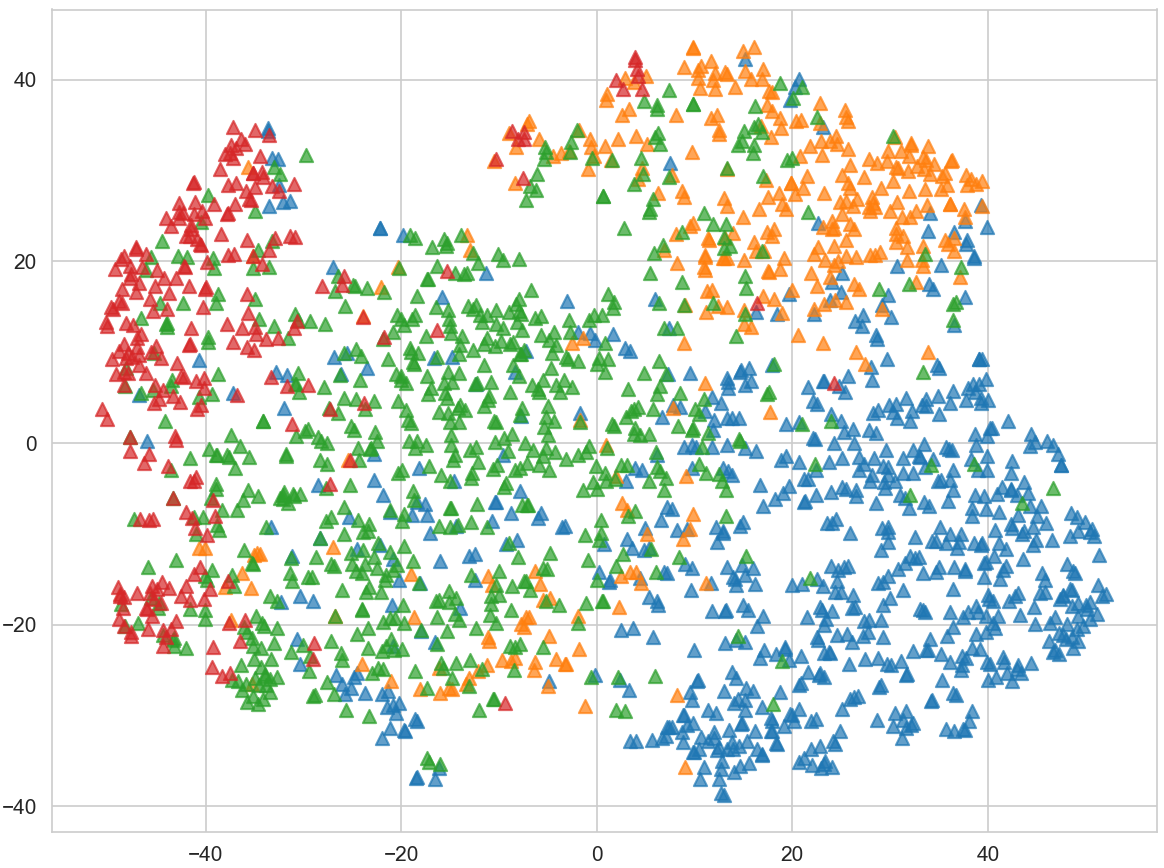}
    \subcaption{IEMOCAP fused features.}\label{fig:m}
  \end{subfigure}

  \caption{t-SNE visualization of synergistic features from each modality and after fusion, showing that cross-modal fusion yields more discriminative representations for classification.}
  \label{fig:visualization-all-datasets}
\end{figure*}
\section{Feature Visualization} \label{app:visualization}

To provide a straightforward visualization of how \PARSE captures synergistic features, we adopt the default experimental settings from Section~\ref{sec:exp}. Figure~\ref{fig:visualization-all-datasets} presents t-SNE plots of the synergistic features for each modality (before fusion) alongside the fused features. On AVE, the fused synergistic features form well-separated class clusters with larger inter-class margins and reduced intra-class scatter compared to pre-fusion features. On IEMOCAP, the fusion further separates overlapping classes, with clearer distinction between the yellow and red clusters. For ModelNet-40, where both modalities are images, the features are more aligned than complementary, reflecting their modality similarity. Overall, these results demonstrate that \PARSE effectively captures complementary information, particularly in settings with heterogeneous modalities such as audio and video.

\section{Federated Learning Comparison} \label{app:fl}

In this section, we evaluate the effectiveness of \PARSE in a centralized FL scenarion. In addition to the common baselines used in the DFL experiments, we adapt the DFL variants as follows: Fed-Modality, where each modality is trained independently and the central server aggregates encoders and classifiers within each modality group; Fed-Task, where agents perform modality-fusion learning locally and parameters are aggregated only among agents with identical modality sets; and Fed-Hybrid, which aggregates parameters by modality while still allowing local modality-fusion learning.  

Beyond existing baselines, we also compare against FL methods that explicitly rely on server-side coordination: FedMSplit \cite{chen2022fedmsplit}, which dynamically learns inter-agent relationship graphs to guide aggregation; FedMVD \cite{gao2025multimodal}, which employs global alignment to mitigate domain shifts caused by modality-based heterogeneity; and FedMVC \cite{chen2024bridging}, a multi-view approach designed to reduce modality heterogeneity across all agents. 

We adopt the same number of agents, agent ratios, and Non-IID level as in the default setting, and report the results in Table~\ref{tab:fl_comp}. As shown, even without a dedicated server-side design, \PARSE achieves leading performance across all multimodal agents (notably on IEMOCAP) and most unimodal agents (with the exception of text-only agents on IEMOCAP). These findings highlight the strong extension potential of \PARSE in standard FL settings, suggesting that further gains may be realized by combining it with widely used server-based FL designs, which we leave as future work.

\begin{table}[h]
\centering
\caption{Comparing methods on four benchmarks in a federated learning setting. We report accuracy (ACC, higher is better) on different modalities (Non-IID Alpha=0.5).}
\label{tab:fl_comp}
\footnotesize
\setlength{\tabcolsep}{4pt}
\renewcommand{\arraystretch}{0.95}

\begin{subtable}{0.6\linewidth}
\caption{KU-HAR}
\centering
\begin{tabular}{lccc}
\toprule
\textbf{Method} & \textbf{A} & \textbf{G} & \textbf{AG} \\
\midrule
Fed-Modality & 85.2$\pm$0.8 & 79.1$\pm$1.3 & 91.3$\pm$0.3 \\
Fed-Task     & 83.2$\pm$1.2 & 76.5$\pm$1.4 & 85.7$\pm$0.8 \\
Fed-Hybrid   & 83.0$\pm$0.5 & 72.5$\pm$0.4 & 89.5$\pm$0.7 \\
Harmony      & 83.7$\pm$1.3 & 78.6$\pm$0.3 & 90.5$\pm$0.6 \\
FedHGB       & 82.2$\pm$1.2 & 75.4$\pm$0.5 & 88.3$\pm$0.9 \\
DMML-KD      & 85.2$\pm$0.6 & 77.6$\pm$1.1 & 92.1$\pm$0.4 \\
FedMVC       & 85.6$\pm$1.5 & 81.2$\pm$1.1 & 92.7$\pm$0.9 \\
FedMSplit    & 82.1$\pm$0.9 & 80.7$\pm$1.1 & 90.2$\pm$1.3 \\
FedMVD       & 85.1$\pm$1.4 & 78.8$\pm$1.0 & 92.5$\pm$0.8 \\
\PARSE       & \textbf{85.8}$\pm$0.9 & \textbf{81.5}$\pm$0.5 & \textbf{93.3}$\pm$1.2 \\
\bottomrule
\end{tabular}
\end{subtable}

\vspace{0.5em}

\begin{subtable}{0.6\linewidth}
\caption{AVE}
\centering
\begin{tabular}{lccc}
\toprule
\textbf{Method} & \textbf{A} & \textbf{V} & \textbf{AV} \\
\midrule
Fed-Modality & 53.4$\pm$1.3 & 57.3$\pm$0.9 & 72.6$\pm$1.2 \\
Fed-Task     & 46.5$\pm$0.7 & 51.2$\pm$0.4 & 65.7$\pm$1.3 \\
Fed-Hybrid   & 48.7$\pm$1.0 & 54.9$\pm$0.8 & 71.2$\pm$1.1 \\
Harmony      & 53.1$\pm$0.5 & 56.9$\pm$0.8 & 71.6$\pm$0.7 \\
FedHGB       & 52.2$\pm$1.4 & 56.4$\pm$1.5 & 70.5$\pm$0.6 \\
DMML-KD      & 52.7$\pm$1.7 & 56.2$\pm$1.5 & 71.6$\pm$0.9 \\
FedMVC       & 52.9$\pm$1.5 & 57.5$\pm$1.7 & 73.1$\pm$1.3 \\
FedMSplit    & 47.7$\pm$1.5 & 52.9$\pm$1.3 & 70.1$\pm$1.0 \\
FedMVD       & 45.2$\pm$1.0 & 54.7$\pm$1.5 & 70.4$\pm$1.5 \\
\PARSE       & \textbf{54.2}$\pm$1.1 & \textbf{57.9}$\pm$1.3 & \textbf{73.3}$\pm$0.7 \\
\bottomrule
\end{tabular}
\end{subtable}

\vspace{0.5em}

\begin{subtable}{0.6\linewidth}
\caption{ModelNet-40}
\centering
\begin{tabular}{lccc}
\toprule
\textbf{Method} & \textbf{V1} & \textbf{V2} & \textbf{V1,V2} \\
\midrule
Fed-Modality & 85.0$\pm$1.3 & 78.3$\pm$0.8 & 86.9$\pm$0.4 \\
Fed-Task     & 81.2$\pm$0.4 & 75.4$\pm$1.8 & 77.3$\pm$1.4 \\
Fed-Hybrid   & 81.1$\pm$0.7 & 77.8$\pm$0.6 & 82.5$\pm$0.8 \\
Harmony      & 84.6$\pm$0.5 & 81.6$\pm$1.2 & 85.6$\pm$1.6 \\
FedHGB       & 81.9$\pm$1.4 & 78.1$\pm$0.5 & 82.3$\pm$1.5 \\
DMML-KD      & 84.8$\pm$0.5 & 80.2$\pm$0.9 & 85.1$\pm$0.4 \\
FedMVC       & 85.7$\pm$1.6 & 81.6$\pm$0.8 & 86.7$\pm$1.6 \\
FedMSplit    & 81.2$\pm$1.0 & 79.3$\pm$0.8 & 87.3$\pm$1.2 \\
FedMVD       & 83.6$\pm$1.2 & 78.8$\pm$1.6 & 87.4$\pm$1.3 \\
\PARSE       & \textbf{86.1}$\pm$0.8 & \textbf{83.3}$\pm$1.1 & \textbf{88.7}$\pm$0.9 \\
\bottomrule
\end{tabular}
\end{subtable}

\vspace{0.5em}

\begin{subtable}{0.7\linewidth}
\caption{IEMOCAP}
\centering
\begin{tabular}{lcccc}
\toprule
\textbf{Method} & \textbf{A} & \textbf{V} & \textbf{T} & \textbf{AVT} \\
\midrule
Fed-Modality & 48.6$\pm$0.4 & 52.6$\pm$0.3 & \textbf{63.6}$\pm$0.7 & 70.6$\pm$0.9 \\
Fed-Task     & 29.6$\pm$0.6 & 45.3$\pm$0.3 & 53.6$\pm$0.5 & 67.4$\pm$0.6 \\
Fed-Hybrid   & 41.3$\pm$0.3 & 52.8$\pm$0.3 & 61.0$\pm$0.7 & 69.2$\pm$0.5 \\
Harmony      & 35.4$\pm$0.4 & 52.5$\pm$0.8 & 61.1$\pm$0.5 & 68.3$\pm$0.7 \\
FedHGB       & 48.0$\pm$0.3 & 52.3$\pm$0.5 & 63.1$\pm$0.4 & 73.6$\pm$0.5 \\
DMML-KD      & 46.0$\pm$0.4 & 53.1$\pm$0.3 & 59.2$\pm$0.5 & 74.3$\pm$0.7 \\
FedMVC       & 47.1$\pm$0.8 & 52.3$\pm$0.2 & 63.2$\pm$0.6 & 73.9$\pm$1.0 \\
FedMSplit    & 47.2$\pm$1.7 & 52.1$\pm$1.1 & 60.3$\pm$0.9 & 71.9$\pm$0.8 \\
FedMVD       & 44.3$\pm$0.7 & 52.3$\pm$0.8 & 57.9$\pm$0.6 & 72.9$\pm$0.7 \\
\PARSE       & \textbf{49.6}$\pm$1.5 & \textbf{55.1}$\pm$1.2 & 61.7$\pm$0.5 & \textbf{77.8}$\pm$0.7 \\
\bottomrule
\end{tabular}
\end{subtable}
\end{table}

\end{document}